%% file: tangent-images.tex
\documentclass[10pt,twocolumn,letterpaper]{article}

\usepackage{cvpr}
\usepackage{times}
\usepackage{epsfig}
\usepackage{graphicx}
\usepackage{amsmath}
\usepackage{amssymb}
\usepackage{enumitem}
\usepackage{multirow}
\usepackage{subcaption}
\usepackage{chapterbib}
\usepackage[table]{xcolor}

\definecolor{noteblue}{rgb}{0.26, 0.34, 0.6}
\definecolor{posgreen}{rgb}{0.13, 0.55, 0.13}
\definecolor{negred}{rgb}{0.5, 0.1, 0.1}

\usepackage[pagebackref=true,breaklinks=true,letterpaper=true,colorlinks,bookmarks=false]{hyperref}

\cvprfinalcopy 

\def\cvprPaperID{175} 

\newcommand{\threesixty}{$360^{\circ}$ }

\begin{document}
\title{Tangent Images for Mitigating Spherical Distortion}

\author{Marc Eder,  
Mykhailo Shvets,
John Lim,
and Jan-Michael Frahm\\
University of North Carolina at Chapel Hill\\
Chapel Hill, NC\\
{\tt\small \{meder, mshvets, jlim13, jmf\}@cs.unc.edu}
}

\maketitle
\input{main-paper/sections/0-abstract.tex}
\input{main-paper/sections/1-introduction.tex}
\input{main-paper/sections/2-related-work.tex}
\input{main-paper/sections/3-mitigating-distortion.tex}
\input{main-paper/sections/4-experiments.tex}
\input{main-paper/sections/5-conclusion.tex}
\input{main-paper/sections/6-changelog.tex}

\hfill

\input{main-paper/sections/7-acknowledgements.tex}
\pagebreak

\enlargethispage{1.5\baselineskip}
{\small
\bibliographystyle{ieee_fullname}

}
\pagebreak


\part*{\twocolumn[\Large \centering Supplementary Material: Tangent Images for Mitigating Spherical Distortion \vspace{24pt}]}

\normalsize{
\input{supplementary/sections/s0-preamble.tex}
\input{supplementary/sections/s1-limitations.tex}
\input{supplementary/sections/s2-transfer-learning.tex}
\input{supplementary/sections/s3-semantic-segmentation.tex}
\input{supplementary/sections/s4-keypoint-dataset.tex}
\input{supplementary/sections/s5-network-details.tex}

\input{supplementary/tables/results-transfer-class.tex}
\input{supplementary/tables/results-semseg-class.tex}

\input{supplementary/figures/tangent-images.tex}

\input{supplementary/figures/qual-sem-seg.tex}

\input{supplementary/tables/keypoint-dataset.tex}

\input{supplementary/tables/keypoint-indiv-results.tex}

\input{supplementary/figures/sift-detections.tex}

\input{supplementary/figures/sift-matches.tex}
}
\end{document}

%% file: main-paper/sections/0-abstract.tex
\begin{abstract}
In this work, we propose ``tangent images,'' a spherical image representation that facilitates transferable and scalable $360^\circ$ computer vision. Inspired by techniques in cartography and computer graphics, we render a spherical image to a set of distortion-mitigated, locally-planar image grids tangent to a subdivided icosahedron. By varying the resolution of these grids independently of the subdivision level, we can effectively represent high resolution spherical images while still benefiting from the low-distortion icosahedral spherical approximation. We show that training standard convolutional neural networks on tangent images compares favorably to the many specialized spherical convolutional kernels that have been developed, while also scaling efficiently to handle significantly higher spherical resolutions. Furthermore, because our approach does not require specialized kernels, we show that we can transfer networks trained on perspective images to spherical data without fine-tuning and with limited performance drop-off. Finally, we demonstrate that tangent images can be used to improve the quality of sparse feature detection on spherical images, illustrating its usefulness for traditional computer vision tasks like structure-from-motion and SLAM.

\vspace{1em}
\textcolor{noteblue}{\small \textbf{Authors Note:} This version of this paper has been updated with new network transfer results not present in the version published at CVPR 2020. These important new results demonstrate that network transfer to spherical images using our representation provides \textbf{equivalent performance} to perspective image networks after only a single epoch of fine-tuning and actually improves performance after 10 epochs of fine-tuning. The accompanying code has also been updated to coincide with this version of the paper. A full log of the changes is provided in Section \ref{sec:changelog}}.
\end{abstract}

%% file: main-paper/sections/1-introduction.tex
\input{main-paper/figures/tangent-images.tex}
\section{Introduction}
A number of methods have been proposed to address convolutions on spherical images. These techniques vary in design, encompassing learnable transformations \cite{su2017learning, Su_2019_CVPR}, generalizations and modifications of the convolution operation \cite{cohen2018spherical, coors2018spherenet, eder2019mapped, tateno2018distortion}, and specialized kernels for spherical representations \cite{cohen2019gauge, jiang2019spherical, zhang2019orientation}. In general, these spherical convolutions fall into two classes: those that operate on equirectangular projections and those that operate on a subdivided icosahedral representation of the sphere. The latter has been shown to significantly mitigate spherical distortion, which leads to significant improvements for dense prediction tasks \cite{eder2019convolutions, eder2019mapped, lee2019spherephd}. It also has the useful property that icosahedron's faces and vertices scale roughly by a factor of $4$ at each subdivision, permitting a simple analogy to $2\times$ upsampling and downsampling operations in standard convolutional neural networks (CNNs). Because of the performance improvements provided by the subdivided icosahedron representation, we focus expressly on it in this paper.

Despite a growing body of work on these icosahedral convolutions, there are two significant impediments to further development: (1) the transferability of standard CNNs to spherical data on the icosahedron, and (2) the difficulty in scaling the proposed spherical convolution operations to high resolution spherical images. Prior work has implied \cite{cohen2019gauge, eder2019mapped} or demonstrated \cite{coors2018spherenet, tateno2018distortion, zhang2019orientation} the transferability of networks trained on perspective images to different spherical representations. However, those who report results see a noticeable decrease in accuracy compared to CNN performance on perspective images and specialized networks that are trained natively on spherical data, leaving this important and desired behavior an unresolved question. Additionally, the proposed specialized convolutional kernels either require subsequent network tuning \cite{cohen2019gauge, zhang2019orientation} or are incompatible with the standard convolution \cite{jiang2019spherical}.

Nearly all prior work on icosahedral convolutions has been built on the analogy between pixels and faces \cite{cohen2019gauge, lee2019spherephd} or pixels and vertices \cite{eder2019mapped, jiang2019spherical, zhang2019orientation}. While elegant on the surface, this parallel has led to difficulties in scaling to higher resolution spherical images. Figure \ref{fig:icosahedrallevels} depicts spherical image resolutions evaluated in the prior work. Notice that the highest resolution obtained so far is a level 8 subdivision, which is comparable to a $512 \times 1024$ equirectangular image. Superficially, this pixel resolution seems reasonably high, but the angular resolution per pixel is still quite low. A $512 \times 1024$ equirectangular image has an angular resolution of $0.352^\circ$. For comparison, a VGA resolution ($480 \times 640$) perspective image with $45^\circ \times 60^\circ$ field of view (FOV) has an angular resolution of $0.094^\circ$. This is most similar to a $2048 \times 4096$ equirectangular image, which has an angular resolution of $0.088^\circ$ and corresponds to a level 10 subdivided icosahedron. As this is a significantly higher resolution than prior work has been capable of demonstrating, this is the resolution on which we test our proposed approach.

In this work, we aim to address both transferability and scalability while leveraging efficient implementations of existing network architectures and operations. To this end, we propose a solution that decouples resolution from subdivision level using oriented, distortion-mitigated images that can be filtered with the standard grid convolution operation. Using these \textit{tangent images}, standard CNN performance is competitive with specialized networks, yet they efficiently scale to high resolution spherical data and open the door to performance-preserving network transfer between perspective and spherical data. Furthermore, use of the standard convolution operation allows us to leverage highly-optimized convolution implementations, such as those from the cuDNN library \cite{chetlur2014cudnn}, to train our networks. Additionally, the benefits of tangent images are not restricted to deep learning, as they address distortion through the data representation rather than the data processing tools. This means that our approach can be used for traditional vision applications like structure-from-motion and SLAM as well.
\input{main-paper/figures/icosahedral-levels.tex}

We summarize our contributions as follows:
\begin{itemize}[topsep=0pt,itemsep=-1ex,partopsep=1ex,parsep=1ex]
    \item We propose the tangent image spherical representation: a set of oriented, low-distortion images rendered tangent to faces of the icosahedron.
    \item We show that standard CNNs trained on tangent images perform competitively with specialized spherical convolutional kernels while also scaling effectively to high resolution spherical images.
    \item We demonstrate that tangent images facilitate network transfer between perspective and spherical images with no fine tuning and minimal performance drop-off.
    \item We illustrate the utility of tangent images for traditional computer vision tasks by using them to improve sparse keypoint matching on spherical images.
\end{itemize}

%% file: main-paper/figures/tangent-images.tex
\begin{figure}[ht]
    \centering
    \begin{subfigure}[b]{0.4\linewidth}
        \centering
        \captionsetup{justification=centering}
        \caption*{\textbf{Level 1 Icosahedron}}
        \vspace{3pt}
        \includegraphics[width=\textwidth]{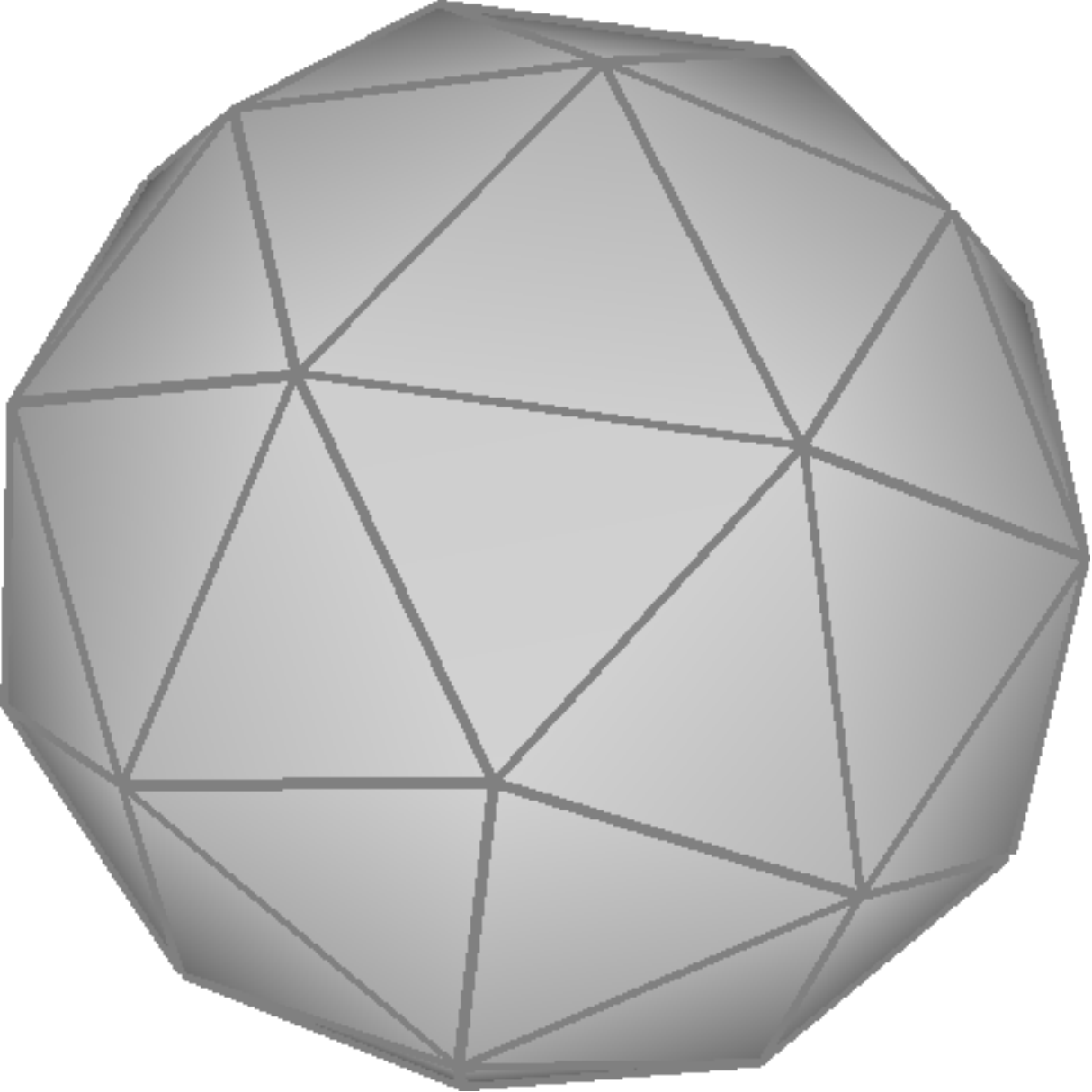}
    \end{subfigure}
    ~
    \centering
    \begin{subfigure}[b]{0.4\linewidth}
        \centering
        \captionsetup{justification=centering}
        \caption*{\textbf{Tangent Images}}
        \includegraphics[width=\textwidth]{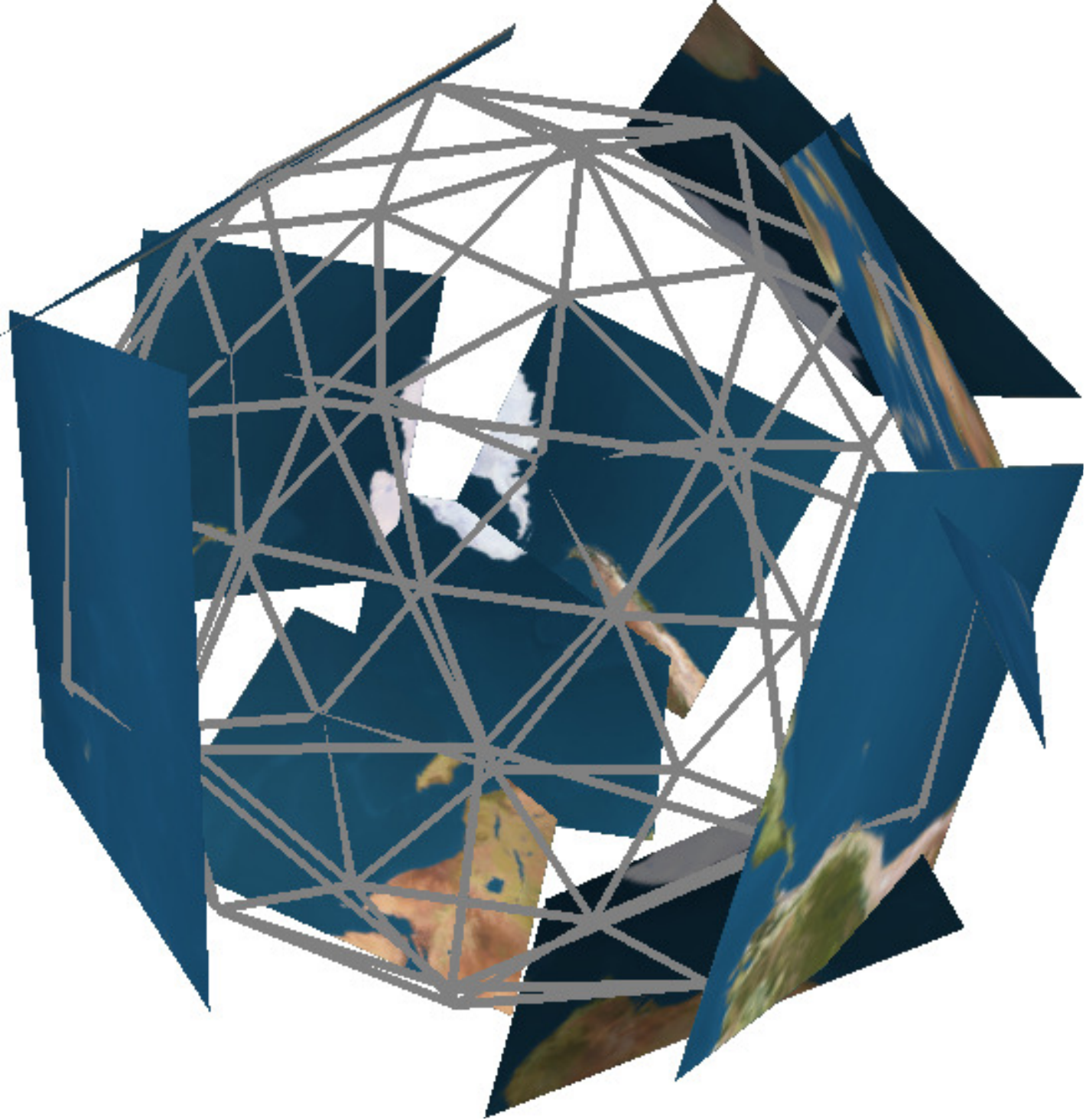}
    \end{subfigure}
    ~
    \centering
    \begin{subfigure}[b]{0.6\linewidth}
        \centering
        \captionsetup{justification=centering}
        \includegraphics[width=\textwidth]{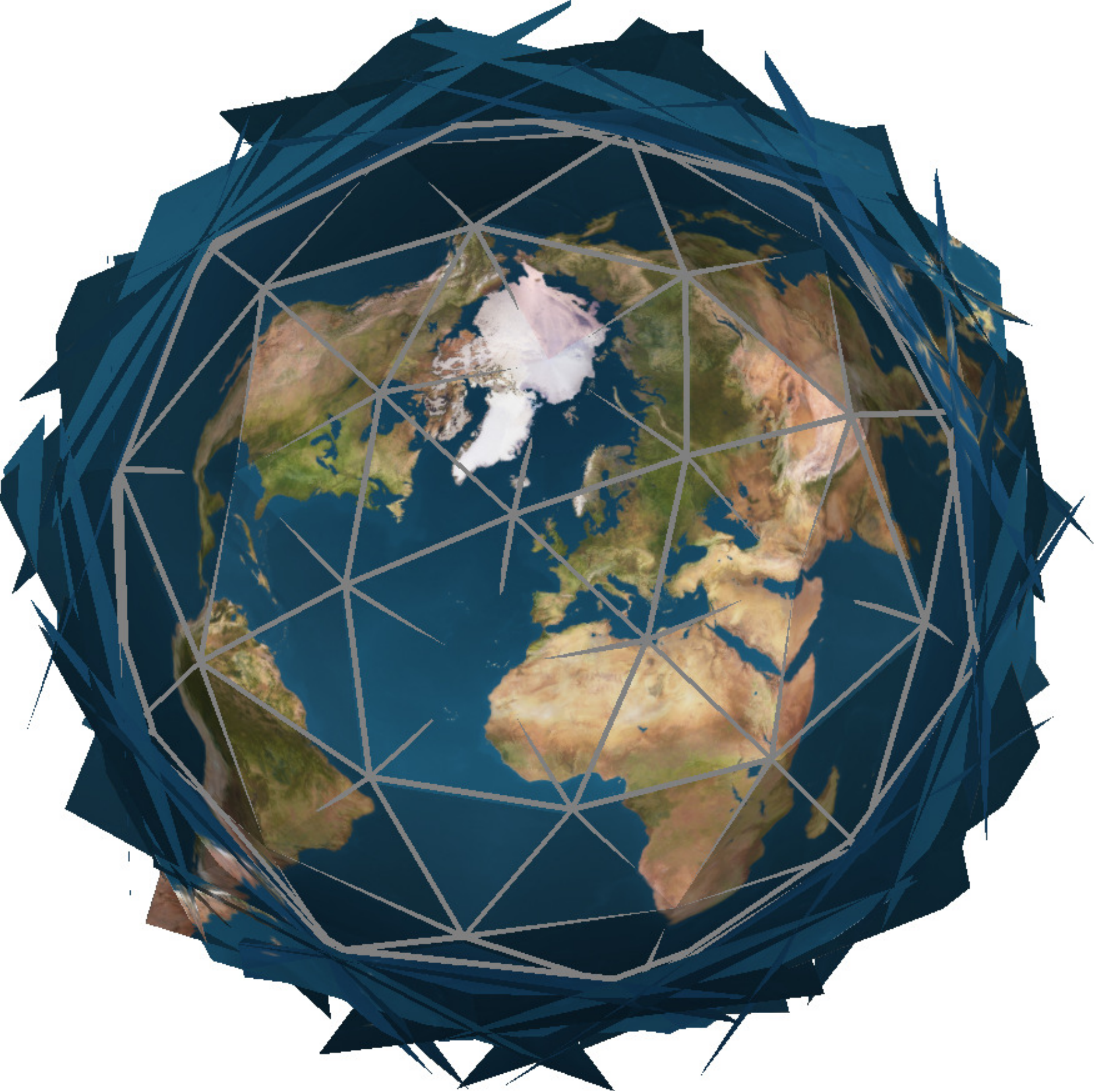}
        \caption*{\textbf{Interior View}}
    \end{subfigure}
\vspace{-2mm}
\caption{\small Using tangent images to represent a 4k Earth image \cite{earth_image}. \textbf{TL:} A base level 1 icosahedron. \textbf{TR:} Selection of tangent images rendered from the Earth image. \textbf{B:} Interior view of the tangent image spherical approximation.}
\label{fig:teaser}
\vspace{-5mm}
\end{figure}

%% file: main-paper/figures/icosahedral-levels.tex
\begin{figure*}[t]
    \centering
    \begin{subfigure}[b]{0.15\textwidth}
        \centering
        \captionsetup{justification=centering}
        \caption*{\textbf{Level 0}\\*
        {[12 vertices,\\* 20 faces]}}
        \includegraphics[height=2.5cm, keepaspectratio]{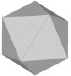}
        \caption*{\textbf{$\mathbf{90^\circ}$ / pixel}\\*
        {[$2 \times 4$ pixels]\\*
        \hfill}}
    \end{subfigure}
    ~
    \centering
    \begin{subfigure}[b]{0.15\textwidth}
        \centering
        \captionsetup{justification=centering}
        \caption*{\textbf{Level 1}\\*
        {[42 vertices,\\* 80 faces]}}
        \includegraphics[height=2.5cm, keepaspectratio]{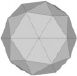}
        \caption*{\textbf{$\mathbf{45^\circ}$ / pixel}\\*
        {[$4 \times 8$ pixels]\\*
        \hfill}}
    \end{subfigure}
    ~
    \centering
    \begin{subfigure}[b]{0.15\textwidth}
        \centering
        \captionsetup{justification=centering}
        \caption*{\textbf{Level 5}\\*
        {[10,242 vertices,\\* 20,480 faces]}}
        \includegraphics[height=2.5cm, keepaspectratio]{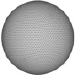}
        \caption*{\textbf{$\mathbf{2.812^\circ}$ / pixel}\\*
        {[$64 \times 128$ pixels]\\*
        \cite{cohen2019gauge, jiang2019spherical}}}
    \end{subfigure}
    ~
    \centering
    \begin{subfigure}[b]{0.15\textwidth}
        \centering
        \captionsetup{justification=centering}
        \caption*{\textbf{Level 7}\\*
        {[163,842 vertices,\\* 327,680 faces]}}
        \includegraphics[height=2.5cm, keepaspectratio]{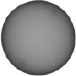}
        \caption*{\textbf{$\mathbf{0.703^\circ}$ / pixel}\\*
        {[$256 \times 512$ pixels]\\*
        \cite{eder2019convolutions, eder2019mapped, lee2019spherephd}}}
    \end{subfigure}
    ~
    \centering
    \begin{subfigure}[b]{0.15\textwidth}
        \centering
        \captionsetup{justification=centering}
        \caption*{\textbf{Level 8}\\*
        {[655,362 vertices,\\* 1,310,720 faces]}}
        \includegraphics[height=2.5cm, keepaspectratio]{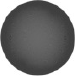}
        \caption*{\textbf{$\mathbf{0.352^\circ}$ / pixel}\\*
        {[$512 \times 1024$ pixels]\\*
        \cite{zhang2019orientation}}}
    \end{subfigure}
    ~
    \centering
    \begin{subfigure}[b]{0.18\textwidth}
        \centering
        \captionsetup{justification=centering}
        \caption*{\textbf{Level 10}\\*
        {[10,485,762 vertices,\\* 20,971,510 faces]}}
        \includegraphics[height=2.5cm, keepaspectratio]{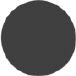}
        \caption*{\textbf{$\mathbf{0.088^\circ}$ / pixel}\\*
        {[$2048 \times 4096$ pixels]\\*
        {[\textit{Ours}]}}}
    \end{subfigure}\\
\vspace{-4mm}
\caption{\small Demonstrating the number of elements, corresponding equirectangular image dimensions, and angular pixel resolution at various icosahedral subdivision levels. The citations beneath each denote the maximum resolution examined in those respective papers. Except for ours, they have all been limited by the pixel-to-face or pixel-to-vertex analogy.}
\label{fig:icosahedrallevels}
\vspace{-5mm}
\end{figure*}

%% file: main-paper/sections/2-related-work.tex
\section{Related Work}
Recently, there have been a number of efforts to close the gap between CNN performance on perspective images and spherical images. These efforts can be naturally divided based on the spherical image representation used.
\subsection{Equirectangular images}
Equirectangular images are a popular spherical image representation thanks to their simple relation between rectangular and spherical coordinates. However, they demonstrate severe image distortion as a result. A number of methods have been proposed to address this issue. Su and Grauman \cite{su2017learning} develop a learnable, adaptive kernel to train a CNN to transfer models trained on perspective images to the equirectangular domain. Su \etal \cite{Su_2019_CVPR} extend this idea by developing a kernel that learns to transform a feature map according to local distortion properties. Cohen \etal \cite{cohen2018spherical, cohen2017convolutional} develop spherical convolutions, which provides the rotational equivariance necessary for convolutions on the sphere. This method requires a specialized kernel, however, making it difficult to transfer the insights developed from years of research into traditional CNNs. Works from Coors \etal \cite{coors2018spherenet} and Tateno \etal \cite{tateno2018distortion} address equirectangular image distortion by warping the planar convolution kernel in a location-dependent manner. Because the equirectangular representation is so highly distorted, most recent work on this topic, has looked to leverage the distorted-reducing properties of the icosahedral spherical approximation.
\subsection{Icosahedral representations}
Representing the spherical image as a subdivided icosahedron mitigates spherical distortion, thus improving CNN accuracy compared to techniques that operate on equirectangular images. Eder and Frahm \cite{eder2019convolutions} motivate this representation using analysis from the field of cartography. Further research on this representation has primarily focused on the development of novel kernel designs to handle discretization and orientation challenges on the icosahedral manifold. Lee \etal \cite{lee2019spherephd} convolve on this representation by defining new, orientation-dependent, kernels to sample from triangular faces of the icosahedron. Jiang \etal \cite{jiang2019spherical} reparameterize the convolutional kernel as a linear combination of differential operators on the surface of an icosahedral mesh. Zhang \etal \cite{zhang2019orientation} present a method that applies a special hexagonal convolution on the icosahedral net. Cohen \etal \cite{cohen2019gauge} precompute an atlas of charts at different orientations that cover the icosahedral grid and use masked kernels along with an feature-orienting transform to convolve on these planar representations. Eder \etal \cite{eder2019mapped} define the ``mapped convolution'' that allows the custom specification of convolution sampling patterns through a type of graph convolution. In this way, they specify the filters' orientation and sample from the icosahedral surface. Our tangent image representation addresses data orientation by ensuring all tangent images are consistently oriented when rendering and circumvents the discretization issue by rendering to image pixel grids.

%% file: main-paper/sections/3-mitigating-distortion.tex
\section{Mitigating Spherical Distortion}\label{sec:distortion}
\input{main-paper/figures/mnist-accuracy.tex}
Image distortion is the reason that we cannot simply apply many state-of-the-art CNNs to spherical data. Distortion changes the representation of the image, resulting in local content deformation that violates translational equivariance, the key property of a signal required for convolution functionality. The graph in Figure \ref{fig:mnistacc} shows just how little distortion is required to produce a significant drop-off in CNN performance.
Distortion in the most popular spherical image representations, equirectangular images and cube maps, is quite significant \cite{eder2019convolutions}, and hence results in even worse performance. Although we can typically remove most lens distortion in perspective images using tools like the Brown-Conrady distortion model \cite{brown1966decentering}, spherical distortion is inescapable and is actually a function of the planar representation. This follows from Gauss's Theorema Egregium, a consequence of which is that a spherical surface is not isometric to a plane. As such, any effort to represent a spherical image as a planar one will result in some degree of distortion. Thus, our objective, and one shared by cartographers for thousands of years, is limited to finding the optimal planar representation of the sphere for our use case.
\vspace{-1mm}
\subsection{The icosahedral sphere}
Consider the classical \textit{method of exhaustion} of approximating a circle with inscribed regular polygons. It follows that, in three dimensions, we can approximate a sphere in the same way. Thus, the choice of planar spherical approximation ought to be the convex Platonic solid with the most faces: the icosahedron. The icosahedron has been used by cartographers to represent Earth at least as early as Buckminster Fuller's Dymaxion map \cite{buckminster1946cartography}, which projects the globe onto the icosahedral net. Recent work in computer vision \cite{cohen2019gauge, eder2019convolutions, eder2019mapped, lee2019spherephd, jiang2019spherical,zhang2019orientation} has demonstrated the shape's utility for resolving the distortion problem for CNNs on spherical images as well.
\input{main-paper/figures/subdivision-surface-area.tex}

While an improvement over single-plane image projections and its Platonic solid cousin, the cube, the 20-face icosahedron on its own is still limited in its distortion-mitigating properties. It can be improved by repeatedly applying Loop subdivision \cite{loop1987smooth} to subdivide the faces and interpolate the vertices, producing increasingly close spherical approximations with decreasing amounts of local distortion on each face. Figure \ref{fig:surfaceareagraph} demonstrates how distortion decreases at each subdivision level. Not all prior work takes advantage of this extra distortion reduction, though. There has largely been a trade-off between efficiency and representation. The charts used by Cohen \etal \cite{cohen2019gauge} and the net used by Zhang \etal \cite{zhang2019orientation} are efficient thanks to their planar image representations, but they are limited to the distortion properties of a level 0 icosahedron. On the other hand, the mapped convolution proposed by Eder \etal \cite{eder2019mapped} operates on the mesh itself and thus can benefit from higher level subdivision, but it does not scale well to higher level meshes due to cache coherence problems when computing intermediate features on the mesh. Jiang \etal \cite{jiang2019spherical} provide efficient performance on the mesh, but do so by approximating convolution with a differential operator, which means existing networks can not be transferred. It is also interesting to note that the current top-performing method for many deep learning tasks, \cite{zhang2019orientation}, uses the net of the level 0 icosahedron. This suggests that extensive subdivisions may not be necessary for all use cases.

Practical methods for processing spherical images must address the efficient scalability problem, but also should permit the transfer of well-researched, high-performance methods designed for perspective images. They should also provide the opportunity to modulate the level of acceptable distortion depending on the application. To address these constraints, we propose to break the coupling of subdivision level and spherical image resolution by representing a spherical image as a collection of images with tunable resolution and distortion characteristics.
\subsection{Tangent images}
\input{main-paper/figures/sampling-levels.tex}
Subdividing the icosahedron provides diminishing returns rather quickly from a distortion-reduction perspective, as indicated by the red vertical line in Figure \ref{fig:surfaceareagraph}. Nonetheless, existing methods must continue to subdivide in order to match the spherical image resolution to the number of mesh elements. We untether these considerations by fixing a \textit{base level} of subdivision, $b$, to define an acceptable degree of distortion, and then rendering the spherical image to square, oriented, planar pixel grids tangent to each face at that base level. The resolution of these \textit{tangent images} is subsequently determined by the resolution of the spherical input. Given a subdivision level, $s$, corresponding to the spherical input resolution, the dimension of the tangent image, $d$, is given by the relation:
\begin{equation}\label{eq:squaredim}
    d = 2^{s - b}
\end{equation}
This design preserves the same resolution scaling that would occur through further subdivisions by instead increasing the resolution of the tangent image. This relationship is illustrated in Figure \ref{fig:samplinglevels}.

Our tangent images are motivated by existing techniques in related fields. The approximation of sections of the sphere by low-distortion planar regions is similar to the Universal Transverse Mercator (UTM) geodetic coordinate system, which divides the Earth into a number of nearly-Euclidean zones. Additionally, as tangent images can be thought of as rendering a spherical mesh to a set of quad textures, the high resolution benefits are similar to Ptex \cite{burley2008ptex}, a computer graphics technique that enables efficient high-resolution texturing by providing every quad of a 3D mesh with its own texture map. A visualization of the tangent image concept is provided in Figure \ref{fig:teaser}. 

\textbf{Computing tangent images} Tangent images are the gnomonic projection of the spherical data onto oriented, square planes centered at each face of a level $b$ subdivided icosahedron. The number of tangent images, $N$, is determined by the faces of the base level icosahedron: $N=20(4^b)$, while their spatial extent is a function of the vertex resolution, $R_v(b-1)$, of the level $b-1$ icosahedron and the resolution of the image grid, given by Equation (\ref{eq:squaredim}). Let $(\phi_f, \lambda_f)$ be the barycenter of a triangular face of the icosahedron in spherical coordinates. We then compute the bounds of the plane in spherical coordinates as the inverse gnomonic projection at central latitude and longitude $(\phi_f, \lambda_f)$ of the points:
\vspace{-1mm}
\begin{equation}\label{eq:tangentbounds}
    \left\{\phi_f \pm \frac{d-1}{2d}R_v(b-1) \right\} \times \left\{\lambda_f \pm \frac{d-1}{2d} R_v(b-1) \right\}
\end{equation}
The vertex resolution, $R_v$, of a level $b$ icosahedron, $\mathcal{S}(b)$, is computed as the mean angle between all vertices, $v$, and their neighbors, $\mathrm{adj}(v)$:
\vspace{-1mm}
\begin{equation}
    R_v(b) = \frac{1}{|\mathcal{S}(b)|}\sum_{v \in \mathcal{S}(b)} \sum_{w \in \mathrm{adj}(v)} \frac{\angle (v, w)}{|\mathrm{adj}(v)|}
\vspace{-1mm}
\end{equation}
Using $R_v(b-1)$ ensures that the tangent images completely cover their associated triangular faces. Because vertex resolution roughly halves at each subsequent subdivision level, we define $R_v(-1) = 2R_v(0)$.

\textbf{Using tangent images} Tangent images require rendering from and to the sphere only once each. First, we create the tangent image set by rendering to the planes defined by Equation (\ref{eq:tangentbounds}). Then, we apply the desired perspective image algorithm (e.g. a CNN or keypoint detector). Finally, we compute the regions on each plane visible to a spherical camera at the center of the icosahedron and render the algorithm output back to the sphere.

We have released our tangent image rendering code and associated experiments as a PyTorch extension\footnote{\url{https://github.com/meder411/Tangent-Images}}.

%% file: main-paper/figures/mnist-accuracy.tex
\begin{figure}[t]
\centering
\begin{subfigure}[b]{0.15\linewidth}
\caption*{\scriptsize \hspace{-1em}$K1 = 0.0$}\vspace{-2mm}
\includegraphics[height=1cm]{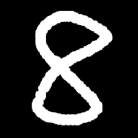}%
\end{subfigure}%
\begin{subfigure}[b]{0.15\linewidth}
\caption*{\scriptsize \hspace{-1em}$K1 = 0.1$}\vspace{-2mm}
\includegraphics[height=1cm]{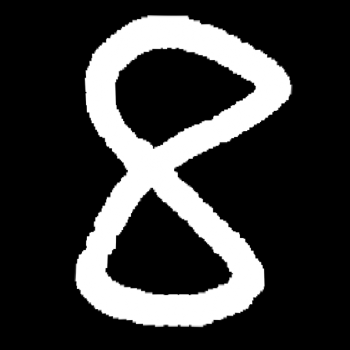}%
\end{subfigure}%
\begin{subfigure}[b]{0.15\linewidth}
\caption*{\scriptsize \hspace{-1em}$K1 = 0.2$}\vspace{-2mm}
\includegraphics[height=1cm]{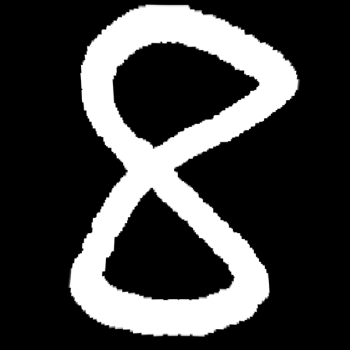}%
\end{subfigure}%
\begin{subfigure}[b]{0.15\linewidth}
\caption*{\scriptsize \hspace{-1em}$K1 = 0.3$}\vspace{-2mm}
\includegraphics[height=1cm]{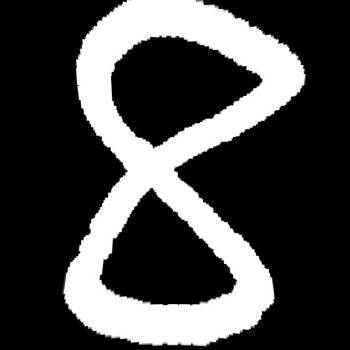}%
\end{subfigure}%
\begin{subfigure}[b]{0.15\linewidth}
\caption*{\scriptsize \hspace{-1em}$K1 = 0.4$}\vspace{-2mm}
\includegraphics[height=1cm]{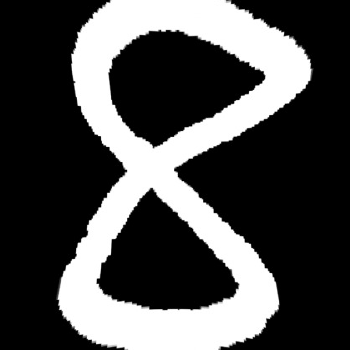}%
\end{subfigure}%
\begin{subfigure}[b]{0.15\linewidth}
\caption*{\scriptsize \hspace{-1em}$K1 = 0.5$}\vspace{-2mm}
\includegraphics[height=1cm]{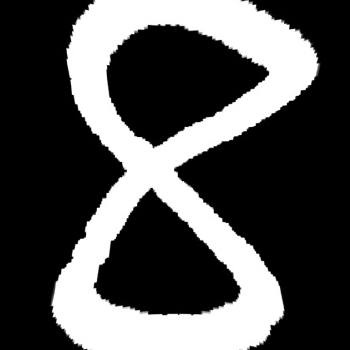}%
\end{subfigure}\\
\vspace{1mm}
\begin{subfigure}[b]{\linewidth}
\includegraphics[width=0.91\linewidth, trim={0mm 0mm 0mm 0mm}, clip]{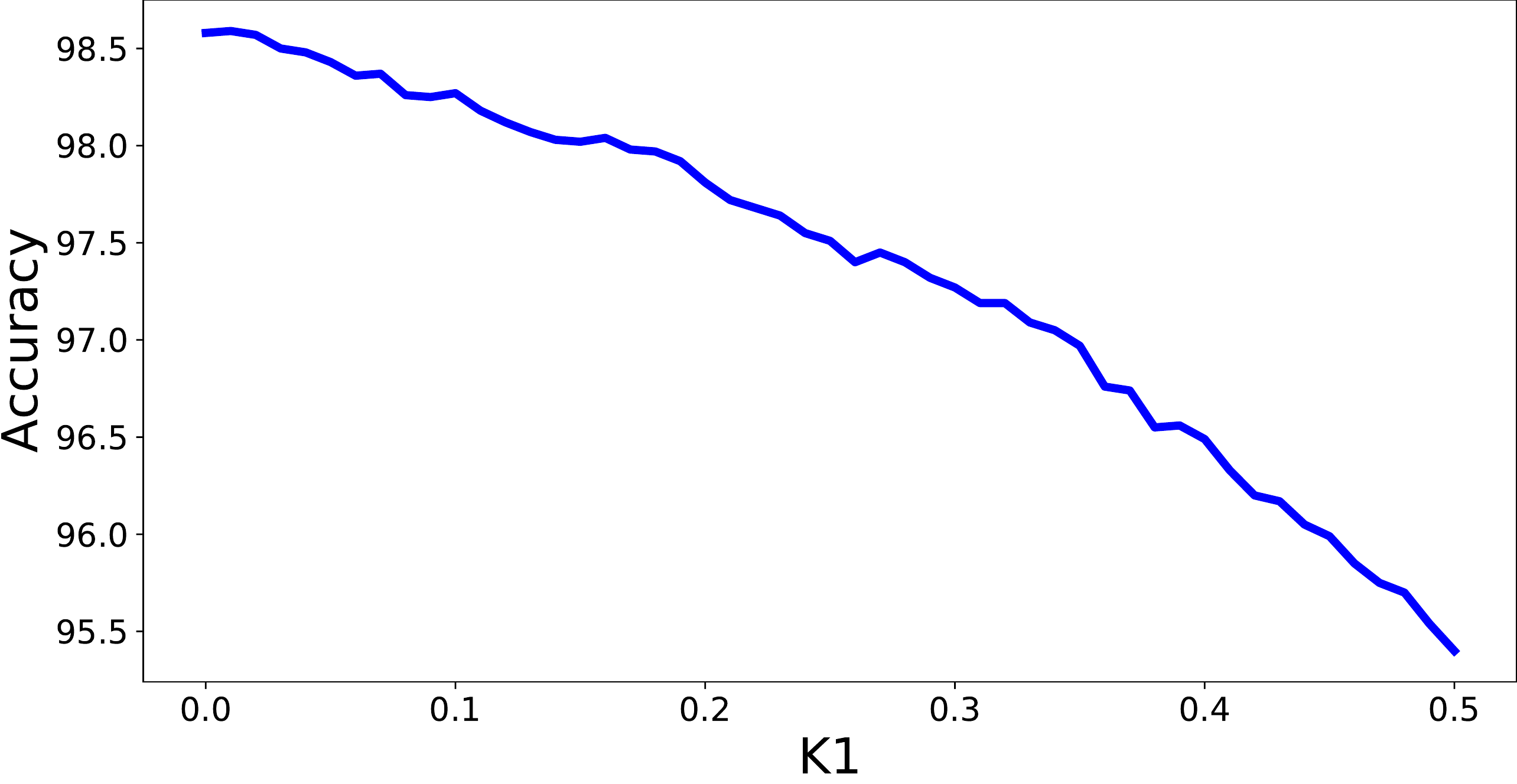}
\end{subfigure}
\vspace{-6mm}
\caption{\small MNIST classification accuracy decreases as pincushion distortion is added to test images by varying the $K1$ parameter of the Brown-Conrady radial distortion model \cite{brown1966decentering}. An example digit is shown at different distortion levels.}
\label{fig:mnistacc}
\vspace{-6mm}
\end{figure}

%% file: main-paper/figures/subdivision-surface-area.tex
\begin{figure}[tbp]
\begin{center}
\includegraphics[width=0.9\linewidth, trim={0mm 0mm 0mm 0mm},clip]{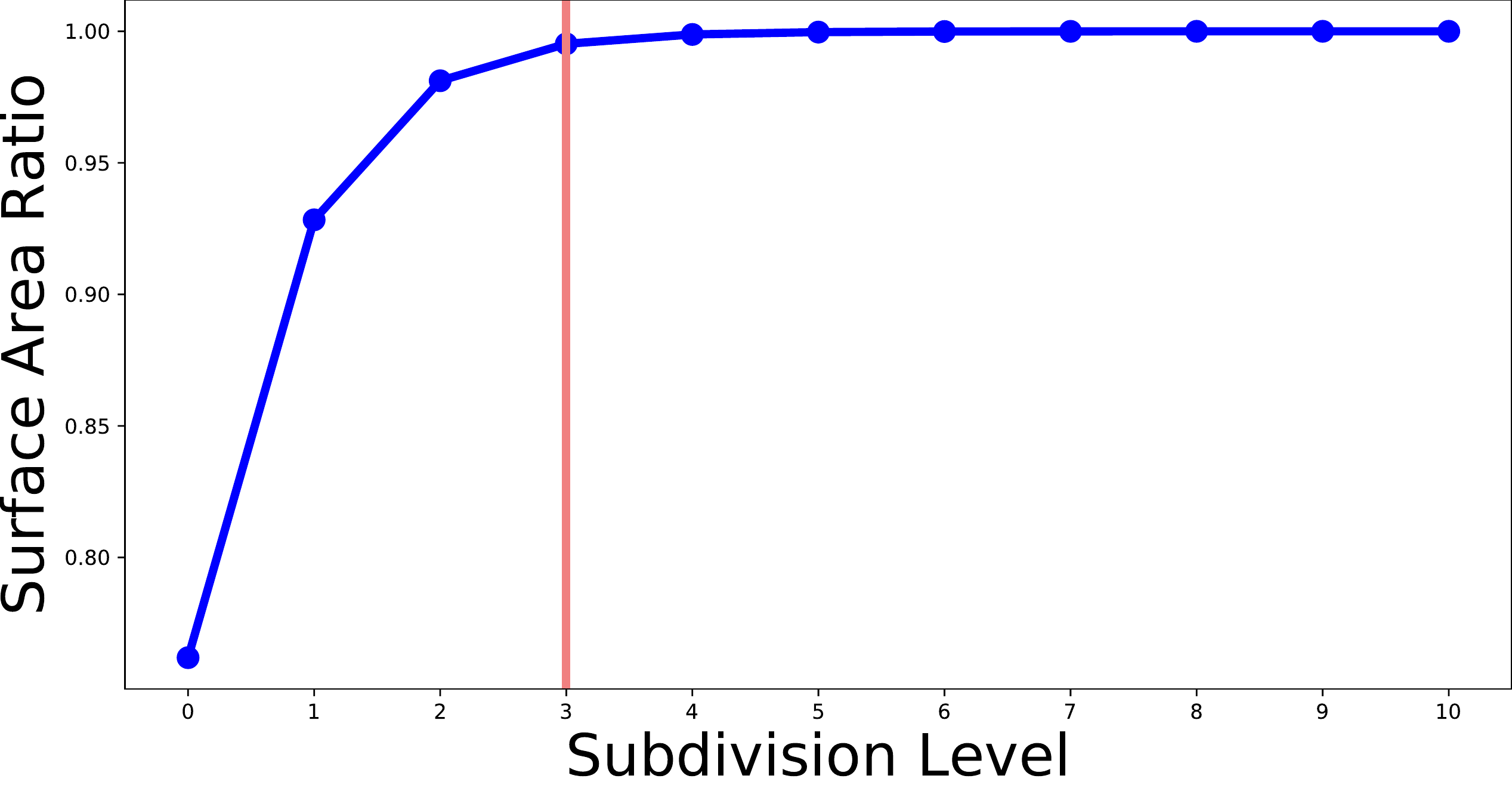}
\end{center}
\vspace{-3mm}
\caption{\small Ratio of the surface area of the subdivided icosahedron to the surface area of a sphere of the same radius at each subdivision level. This global metric demonstrates how closely the subdivision surface approximates a sphere and is drawn from established cartographic metrics \cite{kimerling1999comparing}. Note the leveling off after the third subdivision level.}
\label{fig:surfaceareagraph}
\vspace{-4mm}
\end{figure}

%% file: main-paper/figures/sampling-levels.tex
\begin{figure}[t]
    \begin{center}
    \begin{subfigure}[b]{0.4\linewidth}
        \centering
        \captionsetup{justification=centering}
        \caption*{\textbf{Base Level $\mathbf{+2}$}}
        \fbox{\includegraphics[width=\textwidth]{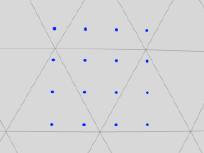}}
    \end{subfigure}
    ~
    \begin{subfigure}[b]{0.4\linewidth}
        \centering
        \captionsetup{justification=centering}
        \caption*{\textbf{Base Level $\mathbf{+3}$}}
        \fbox{\includegraphics[width=\textwidth]{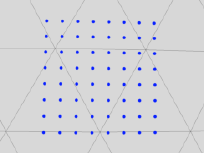}}
    \end{subfigure}\\
    \begin{subfigure}[b]{0.4\linewidth}
        \centering
        \captionsetup{justification=centering}
        \fbox{\includegraphics[width=\textwidth]{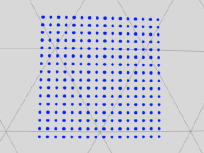}}
        \caption*{\textbf{Base Level $\mathbf{+4}$}}
    \end{subfigure}
    ~
    \begin{subfigure}[b]{0.4\linewidth}
        \centering
        \captionsetup{justification=centering}
        \fbox{\includegraphics[width=\textwidth]{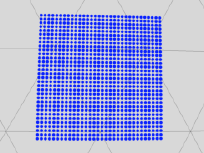}}
        \caption*{\textbf{Base Level $\mathbf{+5}$}}
    \end{subfigure}
    \end{center}
\vspace{-4mm}
\caption{\small Illustrating how the tangent image resolution increases without changing the underlying subdivision level. The field-of-view of the tangent pixel grid remains unchanged, but its resolution increases by a factor of $2$ in each dimension, demonstrated by the blue dots representing pixel samples on the sphere. This scaling maintains the angular pixel resolution of higher level icosahedrons without the need for additional subdivisions.}
\label{fig:samplinglevels}
\vspace{-4mm}
\end{figure}

%% file: main-paper/sections/4-experiments.tex
\section{Experiments}
Prior research has established a common suite of experiments that have become the test bed for new research on spherical convolutions. This set typically includes some combination of spherical MNIST classification \cite{cohen2018spherical, cohen2019gauge, jiang2019spherical, lee2019spherephd, zhang2019orientation}, shape classification \cite{cohen2018spherical, esteves2018learning, jiang2019spherical}, climate pattern segmentation \cite{cohen2019gauge, jiang2019spherical, zhang2019orientation}, and semantic segmentation \cite{cohen2019gauge, jiang2019spherical, lee2019spherephd, tateno2018distortion, zhang2019orientation}. In order to benchmark against these prior works, we evaluate our method on the shape classification and semantic segmentation tasks. Additionally, we demonstrate our method's fairly seamless transfer of CNNs trained on perspective images to spherical data. Finally, to show the versatility of the tangent image representation, we introduce a new benchmark, sparse keypoint detection on spherical images, and compare our representation to an equirectangular image baseline.

\subsection{Classification}
We first evaluate our proposed method on the shape classification task. As with prior work, we use the ModelNet40 dataset \cite{wu20153d} rendered using the method described by Cohen \etal \cite{cohen2018spherical}. Because the data densely encompasses the entire sphere, unlike spherical MNIST, which is sparse and projected only on one hemisphere, we believe this task is more indicative of general classification performance.

\textbf{Experimental setup} We use the network architecture from Jiang \etal \cite{jiang2019spherical}, but we replace the specialized kernels with simple $3 \times 3$ 2D convolutions. A forward pass involves running the convolutional blocks on each patch separately and subsequently aggregating the patch features with average pooling. We train and test on level 5 resolution data as with the prior work.

\input{main-paper/tables/results-classification.tex}

\textbf{Results and analysis}
Results of our experiments are shown in Table \ref{tab:classificationresults}. Without any specialized convolutional kernels, we outperform most of the prior work on this task. The best performing method from Jiang \etal \cite{jiang2019spherical} leverages a specialized convolution approximation on the mesh, which inhibits the ability to fine-tune existing CNN models for the task. Our method can be thought of as using a traditional CNN in a multi-view approach to spherical images. This means that, for global inference tasks like classification, we could select our favorite pre-trained network and transfer it to spherical data. In this case, it is likely that some fine-tuning may be necessary to address the final patch aggregation step in our network design.

\subsection{Semantic segmentation}\label{sec:expsemseg}
We next consider the task of semantic segmentation in order to demonstrate dense prediction capabilities. To compare to prior work, we perform a baseline evaluation of our method at low icosahedron resolutions (5 and 7), but we also evaluate the performance of our method at a level 10 input resolution in order to demonstrate the usefulness of the tangent image representation for processing high resolution spherical data. No prior work has operated at this resolution. We hope that our work can serve as a benchmark for further research on high resolution spherical images.

\textbf{Experimental setup} We train and test our method on the Stanford 2D3DS dataset \cite{armeni2017joint}, as with prior work \cite{cohen2018spherical, cohen2019gauge, jiang2019spherical, zhang2019orientation}. We evaluate RGB-D inputs at levels 5, 7, and 10, the maximum resolution provided by the dataset. At level 10 we also evaluate using only RGB inputs to demonstrate the benefit of high resolution capabilities. For the level 5 and 7 experiments, we use the residual UNet-style architecture as in \cite{jiang2019spherical, zhang2019orientation}, but we again replace the specialized kernels with $3 \times 3$ convolutions. The higher resolution of the level 10 inputs requires the larger receptive field of a deeper network, so we use a FCN-ResNet 101 \cite{he2016deep, long2015fully} model pre-trained on COCO \cite{lin2014microsoft} for those experiments. For level 5 data, we train on the entire set of tangent images, while for the higher resolution experiments, we randomly sample a subset of tangent images from each spherical input to expedite training. We found this sampling method to be useful without loss of accuracy. We liken it to training on multiple perspective views of a scene. Forward passes are run on all sampled tangent images before each backward pass. In this way, the computed gradients at every iteration come from the entire span of the spherical image's field of view.

\input{main-paper/tables/results-semseg.tex}

\textbf{Results and analysis} We report the results of our experiments in Table \ref{tab:semsegresults}. Results on the Stanford2D3DS dataset are averaged over the 3 folds. Individual class results can be found in the supplementary material (Section \ref{sec:supsemseg}). As expected, our method does not perform as well as prior work at the level 5 resolution. Recall that a level 5 resolution spherical image is equivalent to a $16 \times 16$ perspective image with $45^\circ$ FOV. Our method takes that already low angular resolution image and separates it into a set of low pixel resolution images. Although it had limited impact on classification, these dual low resolutions are problematic for dense prediction tasks. We expound on the low-resolution limitation further in the supplementary material (Section \ref{sec:suplimitations}).

Where our tangent image representation excels is when scaling to high resolution images. What we sacrifice in low-resolution performance, we make up for by efficiently scaling to high resolution inputs. By scaling to the full resolution of the dataset, we are able to report the highest performing results ever on this spherical dataset by a wide margin using only RGB inputs. Adding the extra depth channel, we are able to increase the performance further ($+4.9$ mAcc, $+6.9$ mIOU). At input level 10, we find that base level 1 delivers the best trade-off between the lower FOV at higher base levels and the increased distortion present in lower ones. We elaborate on this trade-off in the supplementary material (Section \ref{sec:suplimitations}).

\subsection{Network transfer}\label{sec:exptransfer}
Our contribution aims to address equivalent network performance regardless of the input data format. That is, for a given network, we strive to achieve \textit{equal performance on both perspective and spherical data}. This objective is motivated by the limited number of spherical image datasets and the difficulty of collecting large scale spherical training data. If we can achieve high transferability of perspective image networks, we reduce the need for large amounts of spherical training data. Because generating tangent images inherently converts a spherical image into a collection of perspective ones, this representation facilitates the desired network transferability without requiring fine-tuning on the spherical data and with limited performance drop-off.

\textbf{Experimental setup}
We evaluate the transferability of the tangent image representation in three experiments. 

In the first experiment, we evaluate semantic segmentation performance on a \textit{spherical} image test set using a network trained on the corresponding \textit{perspective image} training set. We fine-tune the pre-trained, FCN-ResNet101 model \cite{he2016deep, long2015fully} provided by the PyTorch model zoo on the Stanford2D3DS dataset's \cite{armeni2017joint} perspective image training set. We then evaluate semantic segmentation performance on the spherical image test set at a level 8 resolution. As with the other semantic segmentation experiments, this experiment uses RGB-D inputs. During the dataset fine-tuning, we make sure to consider the desired angular resolution of the spherical test images. A network trained on perspective images with an angular resolution of $1^\circ$ has learned filters accordingly. Should we apply those filters to an image captured at the identical position, at the same image resolution, but with a narrower FOV, the difference in angular resolution is effectively scale distortion. To match the angular resolution of our spherical evaluation set, we normalize the camera matrices for all perspective images during training such they have the same angular resolution as the test images. Because this is effectively a center-crop of the data, we also randomly shift our new camera center in order capture all parts of the image. Details of this pre-processing are given in the supplementary material (Section \ref{sec:suptransfer}). We evaluate performance without fine-tuning on spherical data, after 1 epoch of spherical fine-tuning, and again after 10 epochs. To control for the extra training, we also evaluate the perspective network after an additional 10 epochs of training.

The second experiment compares the transferability provided by tangent images to prior work that addresses this topic \cite{zhang2019orientation}. Using the network architecture from Zhang \etal \cite{zhang2019orientation}, we train a model on the perspective images from the SYNTHIA dataset \cite{Ros_2016_CVPR} that correspond to the OmniSYNTHIA dataset's \cite{zhang2019orientation} training set. We again utilize the camera normalization procedure mentioned above. We evaluate performance on the OmniSYNTHIA test set at base level 1.

Finally, the third experiment studies the impact of matching angular resolution between training and testing. For this, we apply our FCN-ResNet 101 semantic segmentation model from the first experiment to the spherical test set at various resolutions.

\input{main-paper/tables/results-transfer-stanford.tex}

\input{main-paper/tables/results-transfer-synthia.tex}

\textbf{Results and analysis}
Results for the first two experiments are given in Tables \ref{tab:transferresults} and \ref{tab:transferresultscomparison}, respectively.

In the first experiment, note that both results are attained using a network trained only on perspective data.  Without fine-tuning, we preserve about 97\% of perspective network accuracy and 93\% of the mean IOU. With only a single epoch of fine-tuning on spherical data, we see effective parity in performance, if not slightly improved performance on the spherical inputs. Finally, after 10 epochs of fine-tuning on the spherical format, the transferred network actually noticeably outperforms the original perspective network performance, even when compared to applying the same fine-tuning to the original network on perspective images. We surmise that this improvement comes from the greater FOV provided by the \threesixty image. These results suggest that tangent images sufficiently mitigate distortion, and, as a result, a network can begin to benefit from the ultra-wide field of view of \threesixty images. Although the tangent image representation breaks up the spherical FOV, the gradients are still computed from the full \threesixty image in our training routine, so the network still benefits from this extra information.

Additionally, we provide the results broken down by individual folds of the dataset because Fold 2 highlights the benefit of even 1 epoch of fine-tuning. Fold 2 is the hardest of the three at baseline, which might explain why tangent images provide less of a benefit than for the other two folds. However, after 10 epochs of fine-tuning on spherical data, we see Fold 2 performance increase significantly. We hypothesize that the ability to engage the wider FOV helps address particularly difficult scenes in the test set.

The results of the second experiment demonstrate that the tangent image approach significantly outperforms the prior state-of-the-art without any specialized kernels or subsequent fine-tuning. Note that Zhang \etal \cite{zhang2019orientation} only report results after 10 epochs of fine-tuning on spherical images. Using tangent images actually provides noticeably better results without any such fine-tuning, although when we add 10 epochs of fine-tuning, we see an extra performance boost. It is also worth observing that our transfer results actually outperform the Zhang \etal \cite{zhang2019orientation} results trained natively on spherical data. Our experiments have been limited by the maximum resolution of available spherical image datasets, but this outcome suggests that network transfer with tangent images may permit even higher resolution spherical image inference.

\input{main-paper/figures/transfer-plot.tex}

Finally, the results of the third experiment are plotted in Figure \ref{fig:transferplot}. Recall that this model was trained on perspective images normalized to have a per-pixel angular resolution most similar to that of a level 8 icosahedron. This chart highlights the importance of camera normalization when training on perspective images with the purpose of transferring the network. Observe how performance deteriorates as the angular resolution of the spherical input moves further from the angular resolution of the training data.
\subsection{Sparse keypoint correspondences}
Recent research on spherical images has focused on deep learning tasks, primarily because many of those works have focused on the convolution operation. As our contribution relates to the representation of spherical data, not specifically convolution, we aim to show that our approach has applications beyond deep learning. To this end, we evaluate the use of tangent images for sparse keypoint detection, a critical step of structure-from-motion, SLAM, and a variety of other traditional computer vision applications.

\textbf{Data} As there is no existing benchmark for this task, we create a dataset using a subset of the spherical images provided by the Stanford2D3DS dataset \cite{armeni2017joint}. To create this dataset, we first cluster the dataset's Area 1 images according to the provided room information. Then, for each location, we compute SIFT features \cite{lowe1999object} in the equirectangular images and identify which image pairs have FOV overlap using the spherical structure-from-motion pipeline provided by the OpenMVG library \cite{moulon2016openmvg}. Next, we compute the average volumetric FOV overlap for each overlapping image pair. Because we are dealing with \threesixty images, there are no image bounds to constrain ``visible'' regions. Instead, we use the ground truth depth maps and pose information to back-project each image pair into a canonical pose. We then compute the percentage of right image points visible to the left camera using the the left image depth map to remove occluded points, and vice versa. We average the two values to provide an FOV overlap score for the image pair. This overlap is visualized in Figure \ref{fig:fovoverlap}. We define our keypoints dataset as the top 60 image pairs according to this overlap metric. Finally, we split the resulting dataset into an ``Easy'' set and ``Hard'' set, again based on FOV overlap. The resulting dataset statistics are shown in Table \ref{tab:keypointstats}. All images are evaluated at their full, level 10 resolution. We provide the dataset details in the supplementary material (Section \ref{sec:supkeypointsdataset}) to enable further research.

\input{main-paper/figures/fov-overlap.tex}

\textbf{Experimental setup} To evaluate our proposed representation, we detect and describe keypoints on the tangent image grids and then render those keypoints back to the spherical image. This rendering step ensures only keypoints visible to a spherical camera at the center of the icosahedron are rendered, as the tangent images have overlapping content. We then use OpenMVG \cite{moulon2016openmvg} to compute putative correspondences and geometrically-consistent inlier matches.

\textbf{Results and analysis} We evaluate the quality of correspondence matching at 3 different base levels using the equirectangular image format as a baseline. We compute the \textit{putative matching ratio} (PMR), \textit{matching score} (MS), and \textit{precision} (P) metrics defined by Heinly \etal \cite{heinly2012comparative}. For an image set $\mathcal{S}$ of image pairs, $(L, R)$, with $p$ putative correspondences, $f$ inlier matches, and $n_{\{L,R\}}$ detected keypoints visible to both images, the metrics over the image pairs as defined as follows:
\begin{equation}\label{eq:revequirect}
\begin{split}
    &\mathrm{PMR} = \frac{1}{2|\mathcal{S}|} \sum_{(L,R) \in \mathcal{S}} \left(\frac{p}{n_L} + \frac{p}{n_R}\right)\\
    &\mathrm{MS} =\frac{1}{2|\mathcal{S}|} \sum_{(L,R) \in \mathcal{S}}\left(\frac{f}{n_L} + \frac{f}{n_R}\right)\\
    &\mathrm{P} = \frac{1}{|\mathcal{S}|} \sum_{(L,R) \in \mathcal{S}}\frac{f}{p}
\end{split}
\end{equation}
\noindent In the same way that we compute the FOV overlap, we use the ground truth pose and depth information provided by the dataset to determine which keypoints in the left image should be visible to the right image ($n_L$) and vice versa ($n_R$), accounting for occlusion.

Results are given in Table \ref{tab:keypointresults}. Our use of tangent images has a strong impact on the resulting correspondences, particularly on the hard split. Recall that this split has a lower FOV overlap and fewer inlier matches at the baseline equirectangular representation. Improved performance in this case is thus especially useful. We observe a significant improvement in PMR in both splits. We attribute this improvement to the computation of the SIFT feature vector on our less distorted representation. Like the convolution operation, SIFT descriptors also require translational equivariance in the detection domain. Tangent images restore this property with their low-distortion representation, which enables repeatable descriptors. The better localization of the keypoints affects the inlier matches as well, resulting in a better MS score. We attribute the leveling off in performance beyond level 1 to the reduced FOV of higher level subdivisions, which impedes the detector's ability to find keypoints at larger scales.

\input{main-paper/tables/stats-keypoints.tex}
\input{main-paper/tables/results-keypoints.tex}

%% file: main-paper/tables/results-classification.tex
\begin{table}[t]
    \small
    \centering
    \begin{tabular}{|c|c|c|}
        \hline
        \textbf{Method} & \textbf{Filter} & \textbf{Acc.}\\
        \hline
        Cohen \etal \cite{cohen2018spherical} & Spherical Correlation & 85.0\% \\
        Esteves \etal \cite{esteves2018learning} & Spectral Parameterization & 88.9\%  \\
        Jiang \etal \cite{jiang2019spherical} & MeshConv & \textbf{90.5\%} \\
        \hline
        Ours & 2D Convolution & 89.1\% \\
        \hline
    \end{tabular}
    \vspace{-2mm}
    \caption{\small Classification results on the ModelNet40 dataset \cite{wu20153d}. Without any specialized convolution operations, our approach is competitive with the state of the art spherical convolution methods.}
    \vspace{-4mm}
    \label{tab:classificationresults}
\end{table}

%% file: main-paper/tables/results-semseg.tex
\begin{table}[t]
    \small
    \centering
    \begin{tabular}{|c|c|c|c|c|c|}
        \hline
        \multicolumn{6}{|c|}{\textbf{Stanford2D3DS Dataset}}\\
        \hline
        $\mathbf{s}$ & \textbf{Method} & \textbf{Input} & $\mathbf{b}$ & \textbf{mAcc} & \textbf{mIOU}\\
        \hline
        \multirow{4}{*}{5}
        & Cohen \etal \cite{cohen2019gauge} & RGB-D &  0 & 55.9 & 39.4 \\
        & Jiang \etal \cite{jiang2019spherical} & RGB-D & 5 & 54.7 & 38.3 \\
        & Zhang \etal \cite{zhang2019orientation} & RGB-D  & 0 & \textbf{58.6} & \textbf{43.3} \\
        & Ours & RGB-D & 0 & 50.9 & 38.3 \\
        \hline
        \multirow{4}{*}{7}
        & Tateno \etal \cite{tateno2018distortion}  & RGB & ERP & - & 34.6  \\
        & Lee \etal \cite{lee2019spherephd} & RGB &  7 & 51.4 & -  \\
        \cline{2-6}
        & Ours & RGB-D & 0 & \textbf{59.1} & \textbf{44.9} \\
        \hline
        \multirow{4}{*}{10}
        & Ours & RGB & 0 & 61.0 & 44.3 \\
        & Ours & RGB & 1 & \textbf{65.2} & \textbf{45.6} \\
        & Ours & RGB & 2 & 61.5 & 42.7 \\
        \cline{2-6}
        & Ours & RGB-D & 1 & \textbf{70.1} & \textbf{52.5} \\
        \hline
    \end{tabular}
    \vspace{-2mm}
    \caption{\small Semantic segmentation results. $s$ is the input resolution in terms of equivalent icosahedron level, $b$ is the base subdivison level (ERP denotes equirectangular inputs), mIoU is the mean intersection-over-union metric, and mAcc is the weighted per-class mean prediction accuracy.}
    \label{tab:semsegresults}
    \vspace{-6mm}
\end{table}

%% file: main-paper/tables/results-transfer-stanford.tex
\begin{table}[tbp]
    \small
    \centering
    \begin{tabular}{|c|c|c|c|c|c|c|}
        \hline
        & \textbf{Format} & \textbf{Res.} & \multicolumn{2}{c|}{\textbf{mAcc}} &\multicolumn{2}{c|}{\textbf{mIOU}} \\
        \hline
        \hline
        \multirow{6}{*}{\rotatebox[origin=c]{90}{\textbf{Fold 1}}}
         & P & $d=128$ & 61.7 & - & 47.4 & - \\
         & S & L=8 & 60.2 & \textcolor{negred}{-2.5\%} & 45.3 & \textcolor{negred}{-4.5\%} \\
        \cline{2-7}
         & P-FT-1 & $d=128$ & 60.9 & - & 48.4 & - \\
         & S-FT-1 & L=8 & 62.5 & \textcolor{posgreen}{+2.6\%} & 48.4 & \textcolor{posgreen}{+0.0\%} \\
        \cline{2-7}
         & P-FT-10 & $d=128$ & 60.6 & - & 48.9 & - \\
         & S-FT-10 & L=8 & 66.0 & \textcolor{posgreen}{+8.9\%} & 50.6 & \textcolor{posgreen}{+3.5\%} \\
        \hline
        \hline
        \multirow{6}{*}{\rotatebox[origin=c]{90}{\textbf{Fold 2}}}
         & P & $d=128$ & 57.8 & - & 38.6 & - \\
         & S & L=8 & 47.8 & \textcolor{negred}{-17.2\%} & 34.4 & \textcolor{negred}{-10.6\%} \\
        \cline{2-7}
         & P-FT-1 & $d=128$ & 56.4 & - & 39.6 & - \\
         & S-FT-1 & L=8 & 52.6 & \textcolor{negred}{-6.7\%} & 40.6 & \textcolor{posgreen}{+2.5\%} \\
        \cline{2-7}
         & P-FT-10 & $d=128$ & 56.7 & - & 40.8 & - \\
         & S-FT-10 & L=8 & 55.6 & \textcolor{negred}{-1.9\%} & 43.6 & \textcolor{posgreen}{+7.1\%} \\
        \hline
        \hline
        \multirow{6}{*}{\rotatebox[origin=c]{90}{\textbf{Fold 3}}}
         & P & $d=128$ & 65.9 & - & 51.1 & - \\
         & S & L=8 & 64.2 & \textcolor{negred}{-2.6\%} & 47.3 & \textcolor{negred}{-7.5\%} \\
        \cline{2-7}
         & P-FT-1 & $d=128$ & 66.0 & - & 51.5 & - \\
         & S-FT-1 & L=8 & 68.4 & \textcolor{posgreen}{+3.6\%} & 51.9 & \textcolor{posgreen}{+0.6\%} \\
        \cline{2-7}
         & P-FT-10 & $d=128$ & 65.6 & - & 51.7 & - \\
         & S-FT-10 & L=8 & 70.3 & \textcolor{posgreen}{+7.2\%} & 54.6 & \textcolor{posgreen}{+5.6\%} \\
        \hline
        \hline
        \multirow{6}{*}{\rotatebox[origin=c]{90}{\textbf{ALL FOLDS}}}
         & P & $d=128$ & 65.9 & - & 51.1 & - \\
         & S & L=8 & 64.2 & \textcolor{negred}{-2.6\%} & 47.3 & \textcolor{negred}{-7.5\%} \\
        \cline{2-7}
         & P-FT-1 & $d=128$ & 66.0 & - & 51.5 & - \\
         & S-FT-1 & L=8 & 68.4 & \textcolor{posgreen}{+3.6\%} & 51.9 & \textcolor{posgreen}{+0.6\%} \\
        \cline{2-7}
         & P-FT-10 & $d=128$ & 65.6 & - & 51.7 & - \\
         & S-FT-10 & L=8 & 70.3 & \textcolor{posgreen}{+7.2\%} & 54.6 & \textcolor{posgreen}{+5.6\%} \\
        \hline
    \end{tabular}
    \caption{\small Transfer learning using RGB-D data from the Stanford2D3DS dataset. ``P'' means the original network trained and evaluated on perspective images only, while ``S'' is that network evaluated on spherical data using tangent images, without any fine-tuning. ``FT-\#'' denotes epochs of fine-tuning on a given format. The percentage next to the spherical results denotes how much of the original perspective network performance is maintained across input formats. Notice that, with fine-tuning, tangent images can actually lead to better performance than the corresponding central-perspective network.}
    \vspace{-5mm}
    \label{tab:transferresults}
\end{table}

%% file: main-paper/tables/results-transfer-synthia.tex
\begin{table}[tbp]
    \small
    \centering
    \begin{tabular}{|c|c|c|c|}
        \hline
        $\mathbf{s}$ & \textbf{Method} & \textbf{mAcc} & \textbf{mIOU}\\
        \hline
        \multirow{3}{*}{6}
        & Ours (no FT) & 55.2 & 43.2 \\
        & Ours (FT-10) & \textbf{59.4} & \textbf{46.1} \\
        & Zhang \etal \cite{zhang2019orientation} (FT-10) & 44.8 & 36.7 \\
        \cline{2-4}
        & Zhang \etal \cite{zhang2019orientation} (\textit{native}) & \textit{52.2} & \textit{43.6} \\
        \hline
        \hline
        \multirow{3}{*}{7}
        & Ours (no FT) & 60.2 & 44.9 \\
        & Ours (FT-10) & \textbf{65.3} & \textbf{50.2} \\
        & Zhang \etal \cite{zhang2019orientation} (FT-10) & 47.2 & 38.0 \\
        \cline{2-4}
        & Zhang \etal \cite{zhang2019orientation} (\textit{native}) & \textit{57.1} & \textit{48.3} \\
        \hline
        \hline
        \multirow{3}{*}{8}
        & Ours (no FT) & 70.8 & 54.9 \\
        & Ours (FT-10) & \textbf{73.2} & \textbf{55.7} \\
        & Zhang \etal \cite{zhang2019orientation} (FT-10) & 52.8 & 45.3 \\
        \cline{2-4}
        & Zhang \etal \cite{zhang2019orientation} (\textit{native}) & \textit{55.1} & \textit{47.1} \\
        \hline
    \end{tabular}
    \caption{\small Comparing our transfer learning results to the prior work from Zhang \etal \cite{zhang2019orientation} on the OmniSYNTHIA dataset at different input resolutions, $\mathbf{s}$. ``no-FT'' denotes no fine-tuning on spherical data, ``FT-10'' mean after 10 epochs of fine-tuning, and ``native'' means both trained and evaluated on spherical data. Even without fine-tuning tangent images significantly improve over the previous state-of-the-art. Fine-tuning provides a small, but noticeable, additional further improvement.}
    \label{tab:transferresultscomparison}
\end{table}

%% file: main-paper/figures/transfer-plot.tex
\begin{figure}[t]
\begin{center}
\includegraphics[width=1.0\linewidth, trim={0mm 0mm 0mm 0mm}, clip]{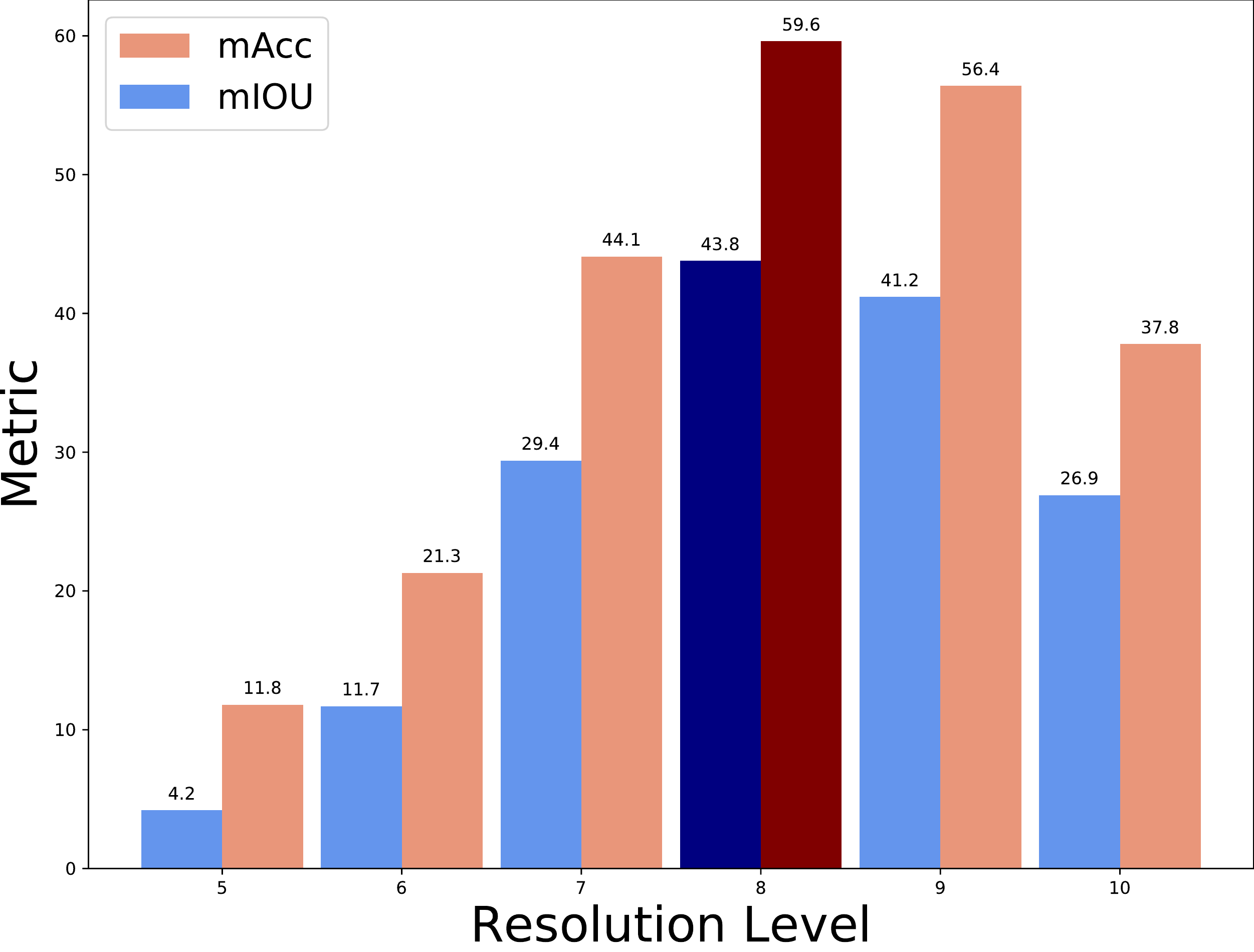}
\end{center}
\vspace{-4mm}
\caption{\small Results are shown for spherical semantic segmentation using a network trained on perspective images that are normalized to have a angular resolution equivalent to a level 8 spherical input. Performance drops off considerably as the angular resolution of the spherical inputs becomes more dissimilar to the training data. Level 8 results are darkened.}
\label{fig:transferplot}
\vspace{-4mm}
\end{figure}

%% file: main-paper/figures/fov-overlap.tex
\begin{figure}[t]
    \centering
    \begin{subfigure}[b]{0.48\linewidth}
        \centering
        \captionsetup{justification=centering}
        \includegraphics[width=\textwidth]{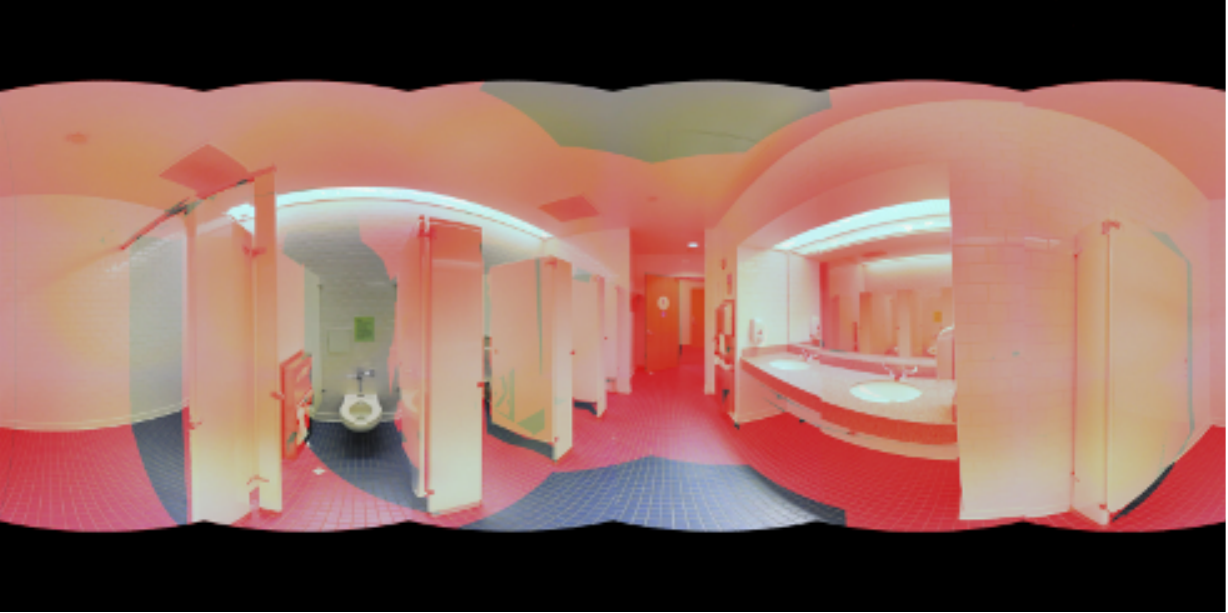}
        \caption*{\textbf{Left Image}}
    \end{subfigure}
    ~
    \centering
    \begin{subfigure}[b]{0.48\linewidth}
        \centering
        \captionsetup{justification=centering}
        \includegraphics[width=\textwidth]{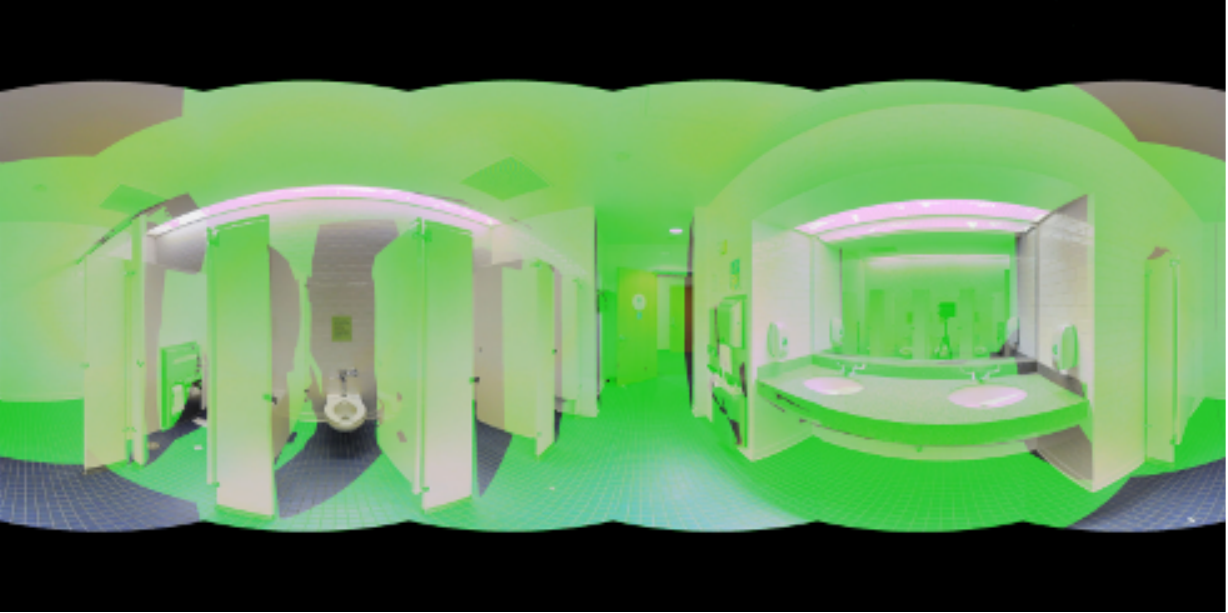}
        \caption*{\textbf{Right Image}}
    \end{subfigure}
\vspace{-2mm}
\caption{\small FOV overlap visualized between an image pair from our keypoints benchmark derived from the Stanford 2D3DS dataset \cite{armeni2017joint}. The red regions in the left image represent areas visible to the right camera, and the green regions in the right image represent areas visible to the left camera.}
\label{fig:fovoverlap}
\vspace{-6mm}
\end{figure}

%% file: main-paper/tables/stats-keypoints.tex
\begin{table}[t]
\small
\centering
\begin{tabular}{|c|c|c|c|}
    \hline
    \textbf{Split} & \textbf{\# Pairs} & \textbf{Mean FOV Overlap} & \textbf{\# Corr.}\\
    \hline
    Hard & 30 & 83.35\% & 298  \\
    Easy & 30 & 89.35\% & 515 \\
    \hline
\end{tabular}
\vspace{-2mm}
\caption{\small Statistics of our keypoints benchmark. \# Corr. is the number of inlier matches detected on the equirectangular images in that split. Statistics are averaged over the splits.}
\label{tab:keypointstats}
\vspace{-5mm}
\end{table}

%% file: main-paper/tables/results-keypoints.tex
\begin{table}[t]
\small
    \centering
    \begin{tabular}{|c|c|c|c|c|}
        \hline
        \multicolumn{5}{|c|}{\textbf{Hard}}\\
        \hline
        \textbf{Metric} & \textbf{Equirect.} & \textbf{L0} & \textbf{L1} & \textbf{L2}  \\
        \hline
        PMR & 22.2\% & 28.4\% & \textbf{30.1\%} & 27.4\% \\
        MS  & 8.2\%  & 11.1\% & \textbf{11.7\%} & 10.9\%  \\
        P   & 36.9\% & 39.5\% & 39.6\% & \textbf{40.2\%}   \\
        \hline
        \hline
        \multicolumn{5}{|c|}{\textbf{Easy}}\\
        \hline
        \textbf{Metric} & \textbf{Equirect.} & \textbf{L0} & \textbf{L1} & \textbf{L2}  \\
        \hline
        PMR & 26.3\% & 32.4\% & \textbf{34.6\%} & 31.9\%  \\
        MS  & 13.6\% & 16.6\% & \textbf{17.7\%} & 16.1\%  \\
        P   & 46.0\% & 46.4\% & \textbf{47.5\%} & 46.5\%  \\
        \hline
    \end{tabular}
    \vspace{-2mm}
    \caption{\small Keypoint evaluation metrics. We report the each metric's average over all image pairs per split. L\{0,1,2\} are the subdivision levels at which we compute the keypoints.}
    \label{tab:keypointresults}
    \vspace{-5mm}
\end{table}

%% file: main-paper/sections/5-conclusion.tex
\section{Conclusion}
We have presented tangent images, a spherical image representation that renders the image onto a oriented pixel grids tangent to a subdivided icosahedron. We have shown that these tangent images do not require specialized convolutional kernels for training CNNs and efficiently scale to represent high resolution data. We have also shown that they facilitate the transfer of networks trained on perspective images to spherical data with limited performance loss. These results further suggest that network transfer using tangent images can open the door to processing even higher resolution spherical images. Lastly, we have demonstrated the utility of tangent images for traditional computer vision tasks in addition to deep learning. Our results indicate that tangent images can be a very useful spherical representation for a wide variety of computer vision applications.

%% file: main-paper/sections/6-changelog.tex
\section{Differences from CVPR 2020 Publication}
\label{sec:changelog}
In order to present the improved performance of our experiments, ensure that our results match our publicly released code, and correct some minor errors in the CVPR 2020 version of this work, we have updated this version.

\begin{itemize}[topsep=0pt,itemsep=-1ex,partopsep=1ex,parsep=1ex]
    \item An additional note about training with tangent images is added to Section \ref{sec:expsemseg}.
    \item Table \ref{tab:semsegresults} in the CVPR version misreports results from Lee \etal \cite{lee2019spherephd}. This version corrects that mistake, but we note that it has no bearing on the analysis.
    \item We clarify that transfer learning experiments are performed on RGB-D images in Section \ref{sec:exptransfer}. The CVPR version has mixed notes on this, saying in one place that they are performed on RGB images only and in another that they are performed on RGB-D inputs.
    \item We provide further exploration of network transfer by fine-tuning the network on spherical data. This additional work was done to be able to better compare to prior work \cite{zhang2019orientation} who report results only after fine-tuning. This extra fine-tuning is done for both network transfer experiments, and demonstrates the ability of tangent images to enable improved performance on spherical images.
\end{itemize}

%% file: main-paper/sections/7-acknowledgements.tex
\noindent \small{\textbf{Acknowledgements} We would like to thank David Luebke, Pierre Moulon, Li Guan, and Jared Heinly for their consultation in support of this work. This research was funded in part by Zillow.}

%% file: supplementary/sections/s0-preamble.tex
In this supplementary material we provide the following additional information:
\begin{itemize}[topsep=0pt,itemsep=-1ex,partopsep=1ex,parsep=1ex]
    \item Expanded discussion of some of the current limitations of tangent images (Section \ref{sec:suplimitations})
    \item The details of the camera normalization process and class-level results of our transfer learning experiments (Section \ref{sec:suptransfer})
    \item Class-level and qualitative results for the semantic segmentation experiments at different input resolutions (Section \ref{sec:supsemseg})
    \item Details of our 2D3DS Keypoints dataset along with individual image pair results and a qualitative comparison of select image pairs (Section \ref{sec:supkeypointsdataset})
    \item Training and architecture details for all CNN experiments detailed in this paper (Section \ref{sec:supnetworkdetails})
    \item An example of a spherical image represented as tangent images (Figure \ref{fig:tangentimages})
\end{itemize}

%% file: supplementary/sections/s1-limitations.tex
\section{Limitations}\label{sec:suplimitations}
We have demonstrated the usefulness of our proposed tangent images, but we have also exposed some limitations of the formulation and opportunities for further research.

\input{main-paper/tables/fov_levels.tex}

\textbf{Resolution} When using tangent images, low angular resolution spherical data is processed on potentially low pixel resolution images. This can severely limit the receptive field of the networks, to which we attribute our poor performance on the level 5 semantic segmentation task, for example. However, this limitation is only notable in the context of the existing literature, because prior work has been restricted to low resolution spherical data, as shown in Figure 2 in the main paper. One viable solution is to incorporate the rendering into the convolution operation. In this way, we could regenerate the tangent images at every operation and effectively resolve the receptive field issue. However, as this is an issue for low resolution images, and our work is focused on addressing high resolution spherical data, we leave this modification for future study.

\textbf{FOV} The base subdivision level provides a constraint on the FOV of the tangent images. Table \ref{tab:angfov} shows the FOV of the tangent images at different base subdivision levels. As the FOV decreases, algorithms that rely on some sense of context or neighborhood break down. We observe this effect for both the CNN and keypoint experiments. While this is certainly a trade-off with tangent images, we have demonstrated that base levels and resolutions can be modulated to fit the required application. Another important point to observe regarding tangent image FOV is that the relationship between FOV and subdivision level does not hold perfectly at lower subdivision levels due the outsize influence of faces near the 12 singular points on the icosahedron. This effect largely disappears after base level 2, but when normalizing camera matrices to match spherical angular resolution at base levels 0 and 1, it is necessary to choose the right base level for the data. We use a base level of 1 in our transfer learning experiments on the OmniSYNTHIA dataset for this reason.

%% file: main-paper/tables/fov_levels.tex
\begin{table}[ht]
\small
\centering
\begin{tabular}{|c|c|c|c|c|c|}
    \hline
    $\mathbf{b}$ & $\mathbf{0}$ & $\mathbf{1}$ & $\mathbf{2}$ & $\mathbf{3}$ & $\mathbf{4}$\\
    \hline
    \textbf{FOV} & $73.1^\circ$ & $51.6^\circ$ & $31.5^\circ$ & $16.7^\circ$ &  $8.4^\circ$\\
    \hline
\end{tabular}
\caption{\small Tangent image field of view at different base levels ($\mathbf{b}$) for a level 10 input. There is a slight variation depending on the input resolution due to faces near the 12 icosahedral singular points, but the values stay mostly consistent.}
\label{tab:angfov}
\end{table}

%% file: supplementary/sections/s2-transfer-learning.tex
\section{Network Transfer}\label{sec:suptransfer}
In this section, we detail the camera normalization process used when training the network for transferring to spherical data. We also provide class-level results for our experiments.

\subsection{Camera normalization}
In order to ensure angular resolution conformity between perspective training data and spherical test data, we normalize the intrinsic camera matrices of our training images to match the angular resolution of the spherical inputs, $\alpha_s$. To do this, we resample all perspective image inputs to a common camera with the desired angular resolution. The angular resolution in radians-per-pixel of an image, $\alpha_x$ and $\alpha_y$, is defined as:
\begin{equation}
    \alpha_x = \frac{ \Omega_x }{W} \quad\quad
    \alpha_y = \frac{ \Omega_y }{H}
\end{equation}
where $\Omega_x$ and $\Omega_y$ are the fields of view of the image as a function of the image dimensions, $W$ and $H$, and the focal lengths, $f_x$ and $f_y$:
\begin{equation}
\begin{split}
    &\Omega_x = 2 \arctan\left( \frac{W}{2f_x} \right) \\
    &\Omega_y = 2 \arctan\left( \frac{H}{2f_y} \right)
\end{split}
\end{equation}
Because spherical inputs have uniform angular resolution in every direction, we resample our perspective inputs likewise: $\alpha_x = \alpha_y = \alpha_s$.

\textbf{Choosing camera properties}
For our camera-normalized perspective images, we want to choose fields of view, $\Omega'_{x}$ and $\Omega'_{y}$, and image dimensions, $W'$ and $H'$ that satisfy:
\begin{equation}
   \frac{\Omega'_{x}}{W'} = \frac{\Omega'_{y}}{H'} = \alpha_s
\end{equation}
While there are a variety of options that we could use, we choose to set $\Omega'_{x} = \Omega'_{y} = \frac{\pi}{4}$ because $\frac{\pi}{4}$ radians ($45^\circ$) is a reasonable field of view for a perspective image. We select $W'$ and $H'$ accordingly. For a level 8 input, this results in $W' = H' = 128$.

\textbf{Normalizing intrinsics}
Recall the definition of the intrinsic matrix, $K$, given focal lengths $f_x$ and $f_y$ and principal point $(c_x, c_y)$:
\begin{equation}
    K = \left[\begin{matrix}
         f_x & 0 & c_x \\
         0 & f_y & c_y \\
         0 & 0 & 1
        \end{matrix}\right]
\end{equation}
Given our choices for fields of view and image dimensions explained above, we compute a new, common intrinsic matrix. The new focal lengths, $f_x'$ and $f_y'$, are computed as:
\begin{equation}
   f_x' = \frac{W'}{2\tan\left(\frac{\Omega'_{x}}{2}\right)} \quad\quad
   f_y' = \frac{H'}{2\tan\left(\frac{\Omega'_{y}}{2}\right)}
\end{equation}
and, for simplicity, the new principal point is chosen to be:
\begin{equation}
   c_x' = \frac{W'}{2} \quad\quad
   c_y' = \frac{H'}{2}
\end{equation}
Defining:
\begin{equation}
    K' = \left[\begin{matrix}
         f_x' & 0 & c_x' \\
         0 & f_y' & c_y' \\
         0 & 0 & 1
        \end{matrix}\right]
\end{equation}
the camera intrinsics can be normalized using the relation:
\begin{equation}\label{eq:camnorm}
    x' = K'K^{-1}x
\end{equation}
where $x$ and $x'$ are homogeneous pixel coordinates in the original and resampled images, respectively, and $K^{-1}$ is the inverse of the intrinsic matrix associated with the original image.

\textbf{Random shifts} If we were to simply resample the image using Equation (\ref{eq:camnorm}), we would end up with center crops of the original perspective images. In order to ensure that we do not discard useful information, we randomly shift the principle point of the original camera by some $(\delta_x, \delta_y)$ before normalizing. This produces variations in where we crop the original image. Including this shift, we arrive at the formula we use for resampling the perspective training data:
\begin{equation}\label{eq:shiftedcamnorm}
    x' = K'\left(K + \Delta \right)^{-1}x
\end{equation}
where:
\begin{equation}
    \Delta = \left[\begin{matrix}
         0 & 0 & \delta_x \\
         0 & 0 & \delta_y \\
         0 & 0 & 0
        \end{matrix}\right]
\end{equation}

To ensure our crops stay within the bounds of the original image, we want:
\begin{equation}
    \begin{split}
    &\delta_x + P(0, 0)_x \geq 0 \\
    &\delta_y + P(0, 0)_y \geq 0 \\
    &\delta_x + P(W', H')_x \leq W \\
    &\delta_y + P(W', H')_y \leq H
    \end{split}
\end{equation}
where $P(x', y')_{\{x, y\}}$ denotes the $x$- and $y$-dimensions of the new camera's coordinates projected into the original camera's coordinate system:
\begin{equation}
P(x', y') = K {K'}^{-1} \left[ \begin{matrix} x' \\ y' \\ 1 \end{matrix} \right]
\end{equation}
Using this constraint, we sample crops over the entire image by randomly choosing $\delta_x$ and $\delta_y$ from the ranges:
\begin{equation}
\begin{split}
&\dfrac{f_x}{f'_x} c'_x - c_x \leq \delta_x \leq W - c_x - \dfrac{f_x}{f'_x} (W' - c'_x) \\
&\dfrac{f_y}{f'_y} c'_y - c_y \leq \delta_y \leq H - c_y - \dfrac{f_y}{f'_y} (H' - c'_y)
\end{split}
\end{equation}
\subsection{Per-class results}
Table \ref{tab:transferclassresults} gives the per-class results for our semantic segmentation transfer experiment. While perspective image performance should be considered an upper bound on spherical performance, note that in some classes, we appear to actually perform better on the spherical image. This is because the spherical evaluation is done on equirectangular images in order to be commensurate across representations. This means that certain labels are duplicated and others are reduced due to distortion, which can skew the per-class results.

%% file: supplementary/sections/s3-semantic-segmentation.tex
\section{Semantic Segmentation}\label{sec:supsemseg}
We provide the per-class quantitative results for our semantic segmentation experiments from Section 4.2 in the main paper. Additionally, we qualitatively analyze the benefits of training higher resolution networks, made possible by the tangent image representation.

\subsection{Class results}
Per-class results are given for semantic segmentation in Table \ref{tab:semsegclassresults}. Nearly every class benefits from the high-resolution inputs facilitated by tangent images. This is especially noticeable for classes with fine-grained detail and smaller objects, like \textit{chair}, \textit{window}, and \textit{bookcase}. The \textit{table} class is an interesting example of the benefit of our method. While prior work has higher accuracy, our high resolution classification has significantly better IOU. In other words, our high resolution inputs may not result in correct classifications of every \textit{table} pixel, but the classifications that are correct are much more precise. This increased precision is reflected almost across the board by mean IOU performance.

\subsection{Qualitative results}
Figure \ref{fig:semsegqual} gives 3 examples of semantic segmentation results at each resolution. The most obvious benefits of the higher resolution results are visible in the granularity of the output segmentation. Notice the fine detail preserved in the chairs in the level 10 output in the bottom row and even the doorway and whiteboard in the middle row. However, recall that our level 10 network uses a base level of 1. The effects of the smaller FOV of the tangent images are visible in the misclassifications of wall on the right of the level 10 output in the middle row. The level 5 network has no such problems classifying that surface, having been trained at a lower input resolution and using base level 0. Nevertheless, it is worth noting that large, homogeneous regions are going to be problematic for high resolution images, regardless of method, due to receptive field limitations of the network. If the region in question is larger than the receptive field of the network, there is no way for the network to infer context. As such, we are less concerned by errors on these large regions.

%% file: supplementary/sections/s4-keypoint-dataset.tex
\section{Stanford 2D3DS Keypoints Dataset}\label{sec:supkeypointsdataset}
\subsection{Details}
Tables \ref{tab:keypointdataseteasy} and \ref{tab:keypointdatasethard} give the details of the image pairs in our keypoints dataset. Tables \ref{tab:keypointindivresultseasy} and \ref{tab:keypointindivresultshard} provide the individual metrics computed for each image pair.
\subsection{Qualitative examples}
We provide a qualitative comparison of keypoint detections in Figure \ref{fig:siftdetections}. These images illustrate two interesting effects, in particular. First, in highly distorted regions of the image that have repeatable texture, like the floor in both images, detecting on the equirectangular format produces a number of redundant keypoints distributed over the distorted region. With tangent images, we see fewer redundant points as the base level increases and the ones that are detected are more accurate and robust, as indicated by the higher MS score. Additionally, the equirectangular representation results in more keypoint detections at larger scales. These outsize scales are an effect of distortion. Rotating the camera so that the corresponding keypoints are detected at different pixel locations with different distortion characteristics will produce a different scale, and consequently a difference descriptor. This demonstrates the need for translational equivariance in keypoint detection, which requires the lower distortion provided by our tangent images. This is reflected quantitatively by the higher PMR scores.

Figure \ref{fig:siftcorr} shows an example of inlier correspondences computed on the equirectangular images and at different base levels for an image pair from the hard split. Even though we detect fewer keypoints using tangent planes, we still have the same quality or better inlier correspondences. Distortion in the equirectangular format results in keypoint over-detection, which can potentially strain the subsequent inlier model fitting. Using tangent images, we detect fewer, but higher quality, samples. This results in more efficient and reliable RANSAC \cite{fischler1981random} model fitting. This is why tangent images perform noticeably better on the hard set, where there are fewer correspondences to be found.

%% file: supplementary/sections/s5-network-details.tex
\section{Network Training Details}\label{sec:supnetworkdetails}
We detail the training parameters and network architectures used in our experiments to encourage reproducible research. All deep learning experiments were performed using the PyTorch library.
\subsection{Shape classification}
For the shape classification experiment, we use the network architecture from \cite{jiang2019spherical}, replacing their MeshConv layers with $3 \times 3$ 2D convolutions with unit padding. For downsampling operations, we bilinearly interpolate the tangent images. We first render the ModelNet40 \cite{wu20153d} shapes to equirectangular images to be compatible with our tangent image generation implementation. The equirectangular image dimensions are $64 \times 128$, which is equivalent to the level 5 icosahedron. We train the network with a batch size of $16$ and learning rate of $5*10^{-3}$ using the Adam optimizer \cite{kingma2014adam}.
\subsection{Semantic segmentation}
We use the residual U-Net-style architecture from \cite{jiang2019spherical, zhang2019orientation} for our semantic segmentation results at levels 5 and 7. Instead of MeshConv \cite{jiang2019spherical} and HexConv \cite{zhang2019orientation}, we swap in $3 \times 3$ 2D convolutions with unit padding. For the level 10 results, we use the fully-convolutional ResNet101 \cite{he2016deep} pre-trained on COCO \cite{lin2014microsoft} provided in the PyTorch model zoo. We train and test on each of the standard folds of the Stanford 2D3DS dataset \cite{armeni2017joint}. For all spherical data, evaluation metrics are computed \textit{on the re-rendered spherical image}, not on the tangent images. As mentioned in the main paper, when training with tangent images, forward passes are run on all tangent images in a batch before a backward pass is run. This ensures that the whole \threesixty image is incorporated into the gradient computation.

\textbf{Level 5, 7 parameters} For the level 5 and 7 experiments, our tangent images were base level 0 RGB-D images, and we use all 20 tangent images from each image in a batch of 8 spherical images, resulting in an effective batch size of 160 tangent images. We use Adam optimization \cite{kingma2014adam} with an initial learning rate of $10^{-2}$ and decay by $0.9$ every 20 epochs, as in \cite{jiang2019spherical}.

\textbf{Level 10 parameters} For the level 10 experiments, we use RGB-D images at base level 1 and randomly sample 4 tangent images from each image in a batch of 4 spherical inputs, resulting in an effective batch size of 16 tangent images. Because the pre-trained network does not have a depth channel, we initialize the depth channel filter with zero weights. This has the effect of slowly adding the depth information to the model. Similarly, the last layer is randomly initialized, as Stanford 2D3DS has a different number of classes than COCO. We use Adam optimization \cite{kingma2014adam} with a learning rate of $10^{-4}$.
\subsection{Transfer learning}
We again use the fully-convolutional ResNet101 \cite{he2016deep} architecture pre-trained on COCO \cite{lin2014microsoft}. We fine tune for 10 epochs on the perspective images of the Standford2D3DS dataset \cite{armeni2017joint}. We use a batch size of $16$ and a learning rate of $10^{-4}$. When fine-tuning, we again use a batch size of $16$, but we reduce the learning rate to $10^{-5}$.

%% file: supplementary/tables/results-transfer-class.tex
\begin{table*}[tbp]
    \small
    \centering
    \begin{tabular}{|c||c|c|c||c|c|c|}
    \hline
    \multicolumn{7}{|c|}{\textbf{mAcc}}\\
    \hline
    \textbf{Class} & \textbf{P} & \textbf{S} & \textbf{Perf. \%} & \textbf{P-FT-10} & \textbf{S-FT-10} & \textbf{Perf. \%} \\
    \hline
    \textit{beam} & 28.0 & 22.9 & \textcolor{negred}{-18.1\%} & 15.3 & 28.1 & \textcolor{posgreen}{+83.0\%} \\
    \textit{board} & 71.6 & 57.1 & \textcolor{negred}{-20.2\%} & 68.9 & 65.3 & \textcolor{negred}{-5.2\%} \\
    \textit{bookcase} & 66.5 & 53.5 & \textcolor{negred}{-19.6\%} & 65.8 & 63.0 & \textcolor{negred}{-4.3\%} \\
    \textit{ceiling} & 88.0 & 88.8 & \textcolor{posgreen}{+0.9\%} & 89.3 & 87.8 & \textcolor{negred}{-1.7\%} \\
    \textit{chair} & 66.7 & 66.6 & \textcolor{negred}{-0.3\%} & 67.7 & 73.8 & \textcolor{posgreen}{+9.1\%} \\
    \textit{clutter} & 57.4 & 45.9 & \textcolor{negred}{-20.2\%} & 61.1 & 56.2 & \textcolor{negred}{-8.1\%} \\
    \textit{column} & 21.1 & 24.5 & \textcolor{posgreen}{+16.4\%} & 14.7 & 25.3 & \textcolor{posgreen}{+72.7\%} \\
    \textit{door} & 60.2 & 67.1 & \textcolor{posgreen}{+11.4\%} & 58.3 & 75.9 & \textcolor{posgreen}{+30.2\%} \\
    \textit{floor} & 90.0 & 94.8 & \textcolor{posgreen}{+5.4\%} & 91.3 & 94.1 & \textcolor{posgreen}{+3.1\%} \\
    \textit{sofa} & 36.3 & 36.2 & \textcolor{negred}{-0.1\%} & 41.1 & 50.1 & \textcolor{posgreen}{+21.9\%} \\
    \textit{table} & 75.7 & 54.9 & \textcolor{negred}{-27.4\%} & 73.7 & 63.3 & \textcolor{negred}{-14.1\%} \\
    \textit{wall} & 77.9 & 72.4 & \textcolor{negred}{-7.1\%} & 82.7 & 81.9 & \textcolor{negred}{-0.9\%} \\
    \textit{window} & 64.1 & 61.7 & \textcolor{negred}{-3.7\%} & 62.7 & 67.0 & \textcolor{posgreen}{+6.9\%} \\
    \hline
    \end{tabular}\\
    \vspace{2em}
    \begin{tabular}{|c||c|c|c||c|c|c|}
    \hline
    \multicolumn{7}{|c|}{\textbf{mIOU}}\\
    \hline
    \textbf{Class} & \textbf{P} & \textbf{S} & \textbf{Perf. \%} & \textbf{P-FT-10} & \textbf{S-FT-10} & \textbf{Perf. \%} \\
    \hline
    \textit{beam} & 8.6 & 8.4 & \textcolor{negred}{-2.3\%} & 6.7 & 10.8 & \textcolor{posgreen}{+62.2\%} \\
    \textit{board} & 50.5 & 48.1 & \textcolor{negred}{-4.7\%} & 56.2 & 54.3 & \textcolor{negred}{-3.4\%} \\
    \textit{bookcase} & 45.5 & 42.5 & \textcolor{negred}{-6.6\%} & 47.0 & 49.2 & \textcolor{posgreen}{+4.6\%} \\
    \textit{ceiling} & 72.6 & 73.1 & \textcolor{posgreen}{+0.6\%} & 73.6 & 83.9 & \textcolor{posgreen}{+13.9\%} \\
    \textit{chair} & 50.1 & 50.3 & \textcolor{posgreen}{+0.4\%} & 51.2 & 57.9 & \textcolor{posgreen}{+13.2\%} \\
    \textit{clutter} & 36.8 & 34.3 & \textcolor{negred}{-6.9\%} & 38.4 & 39.4 & \textcolor{posgreen}{+2.7\%} \\
    \textit{column} & 11.5 & 9.9 & \textcolor{negred}{-14.2\%} & 10.0 & 10.2 & \textcolor{posgreen}{+2.8\%} \\
    \textit{door} & 49.0 & 43.2 & \textcolor{negred}{-11.7\%} & 48.9 & 51.3 & \textcolor{posgreen}{+4.9\%} \\
    \textit{floor} & 82.0 & 71.3 & \textcolor{negred}{-13.0\%} & 84.0 & 90.5 & \textcolor{posgreen}{+7.7\%} \\
    \textit{sofa} & 22.2 & 23.3 & \textcolor{posgreen}{+4.8\%} & 24.9 & 30.5 & \textcolor{posgreen}{+22.5\%} \\
    \textit{table} & 50.5 & 46.2 & \textcolor{negred}{-8.5\%} & 53.7 & 53.8 & \textcolor{posgreen}{+0.3\%} \\
    \textit{wall} & 64.6 & 60.0 & \textcolor{negred}{-7.0\%} & 66.5 & 68.1 & \textcolor{posgreen}{+2.5\%} \\
    \textit{window} & 50.1 & 39.7 & \textcolor{negred}{-20.8\%} & 51.6 & 45.0 & \textcolor{negred}{-12.7\%} \\
    \hline
    \end{tabular}
    \caption{\small Per-class results for the semantic segmentation transfer learning experiment on the Stanford 2D3DS dataset \cite{armeni2017joint}. ``Perf. \%'' denotes how much better (+) or worse (-) the transferred network performance is as a percentage of the corresponding perspective image network. ``P'' and ``S'' refer to evaluation on perspective images and spherical images, respectively. ``FT-10'' refers to the network fine-tuned on the associated format for 10 epochs. These metrics are averaged over all folds.}
    \label{tab:transferclassresults}
\end{table*}

%% file: supplementary/tables/results-semseg-class.tex
\begin{table*}[ht]
    \small
    \centering
    \begin{tabular}{|c|c|c|c|c|c|c|c|c|c|c|c|c|c|}
    \hline
    \multicolumn{14}{|c|}{\textbf{mAcc}}\\
    \hline
    \textit{Method} & \textit{beam} & \textit{board} & \textit{bookcase} & \textit{ceiling} & \textit{chair} & \textit{clutter} & \textit{column} & \textit{door} & \textit{floor} & \textit{sofa} & \textit{table} & \textit{wall} & \textit{window} \\
    \hline
    Jiang \etal \cite{jiang2019spherical} & 19.6 & 48.6 & 49.6 & 93.6 & 63.8 & 43.1 & 28.0 & 63.2 & 96.4 & 21.0 & 70.0 & 74.6 & 39.0 \\
    Zhang \etal \cite{zhang2019orientation} & 23.2 & 56.5 & 62.1 & \textbf{94.6} & 66.7 & 41.5 & 18.3 & 64.5 & 96.2 & 41.1 & \textbf{79.7} & 77.2 & 41.1 \\
    \hline
    Ours L5 & 28.5 & 39.3 & 48.0 & 85.8 & 51.3 & 54.2 & 11.9 & 63.5 & 92.7 & 16.6 & 51.1 & 77.5 & 41.6 \\
    Ours L7 & \textbf{30.4} & 57.2 & 56.8 & 88.6 & 72.7 & 56.6 & 18.4 & 68.9 & 95.2 & 27.8 & 57.9 & \textbf{78.6} & 58.6 \\
    Ours L10 & 26.6 & \textbf{76.6} & \textbf{63.0} & 90.2 & \textbf{85.0} & \textbf{62.5} & \textbf{37.4} & \textbf{72.0} & \textbf{97.3} & \textbf{67.5} & 73.6 & \textbf{78.6} & \textbf{80.5} \\
    \hline
    \end{tabular}
    
    \vspace{4mm}
    \begin{tabular}{|c|c|c|c|c|c|c|c|c|c|c|c|c|c|}
    \hline
    \multicolumn{14}{|c|}{\textbf{mIOU}}\\
    \hline
    \textit{Method} & \textit{beam} & \textit{board} & \textit{bookcase} & \textit{ceiling} & \textit{chair} & \textit{clutter} & \textit{column} & \textit{door} & \textit{floor} & \textit{sofa} & \textit{table} & \textit{wall} & \textit{window} \\
    \hline
    Jiang \etal \cite{jiang2019spherical} & 8.7 & 32.7 & 33.4 & 82.2 & 42.0 & 25.6 & 10.1 & 41.6 & 87.0 & 7.6 & 41.7 & 61.7 & 23.5 \\
    Zhang \etal \cite{zhang2019orientation} & \textbf{10.9} & 39.7 & 37.2 & 84.8 & 50.5 & 29.2 & 11.5 & 45.3 & 92.9 & 19.1 & 49.1 & 63.8 & 29.4 \\
    \hline
    Ours L5 & 9.0 & 27.5 & 34.7 & 81.4 & 38.4 & 30.2 & 5.2 & 42.6 & 89.2 & 10.2 & 42.0 & 58.3 & 29.5 \\
    Ours L7 & 10.3 & 41.3 & 40.5 & 84.8 & 51.0 & 40.0 & 7.3 & 47.2 & 92.6 & 16.0 & 49.7 & 66.0 & 37.2 \\
    Ours L10 & \textbf{6.0} & \textbf{49.3} & \textbf{49.7} & \textbf{85.4} & \textbf{71.7} & \textbf{44.4} & \textbf{16.0} & \textbf{52.2} & \textbf{94.5} & \textbf{33.1} & \textbf{62.1} & \textbf{70.0} & \textbf{48.5} \\
    \hline
    \end{tabular}
    \caption{\small Per-class results for RGB-D inputs on the Stanford 2D3DS dataset \cite{armeni2017joint}}
    \label{tab:semsegclassresults}
\end{table*}

%% file: supplementary/figures/tangent-images.tex
\begin{figure*}[ht]
    \setlength{\fboxsep}{0pt}
    \setlength{\fboxrule}{1pt}

    \begin{subfigure}[b]{\textwidth}
        \centering
            \fbox{\includegraphics[height=4cm, keepaspectratio, trim={0mm 0mm 0mm 0mm}, clip]{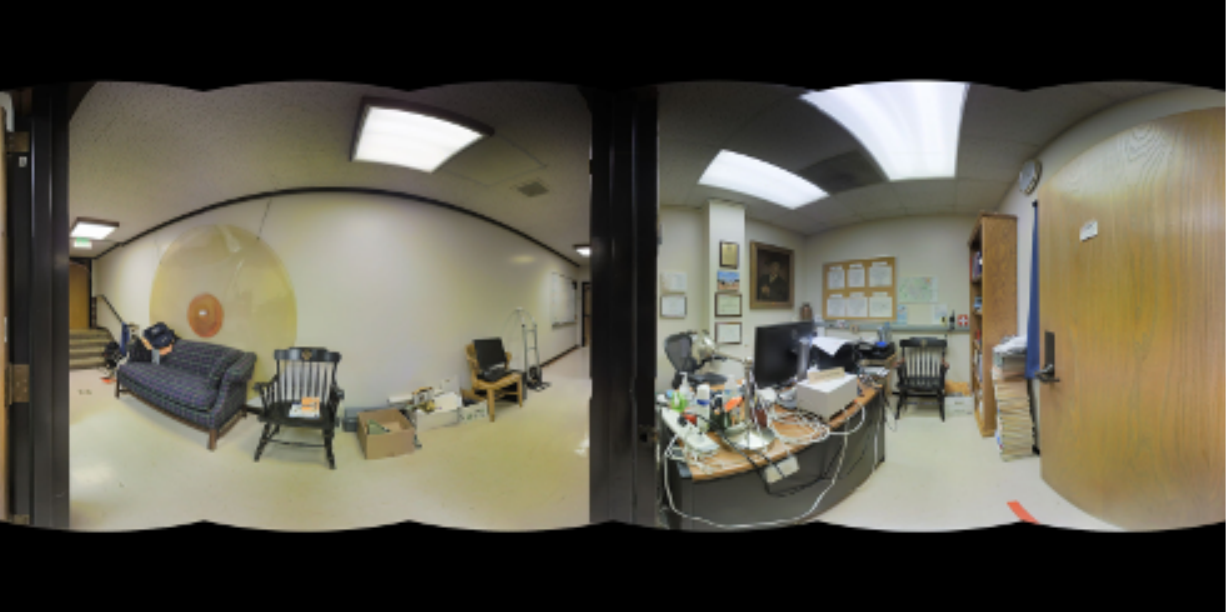}}%
            \caption{\small Spherical image (equirectangular format)}
    \end{subfigure}\\
    
    \vspace{4mm}
    \begin{subfigure}[b]{\textwidth}
        \centering
        \begin{subfigure}[b]{\textwidth}
            \centering
            \fbox{\includegraphics[height=2cm, keepaspectratio, trim={0mm 0mm 0mm 0mm}, clip]{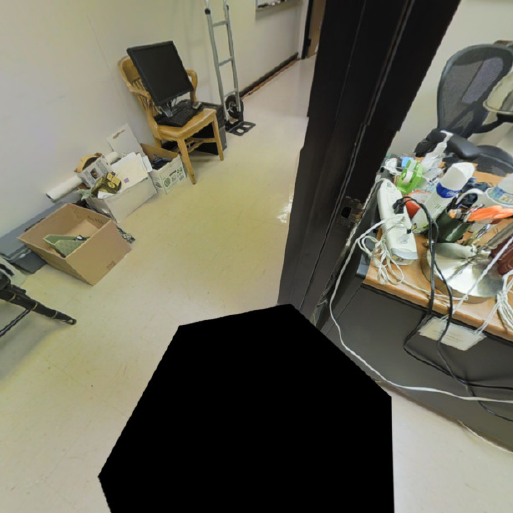}}%
            \fbox{\includegraphics[height=2cm, keepaspectratio, trim={0mm 0mm 0mm 0mm}, clip]{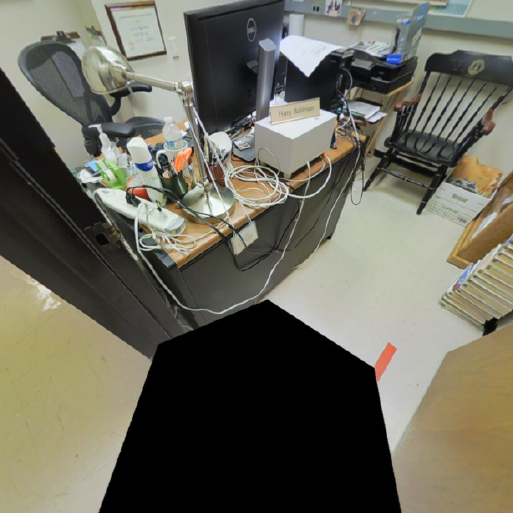}}%
            \fbox{\includegraphics[height=2cm, keepaspectratio, trim={0mm 0mm 0mm 0mm}, clip]{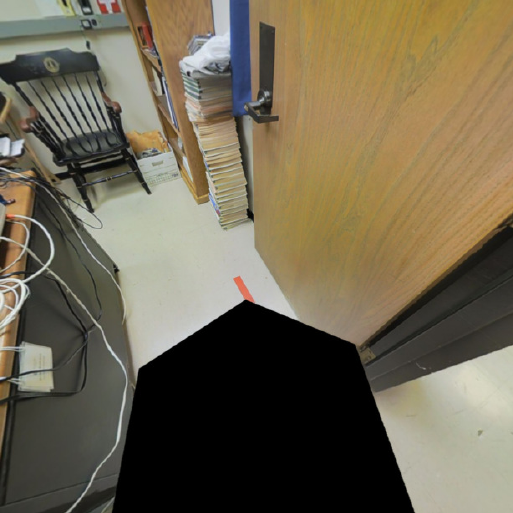}}%
            \fbox{\includegraphics[height=2cm, keepaspectratio, trim={0mm 0mm 0mm 0mm}, clip]{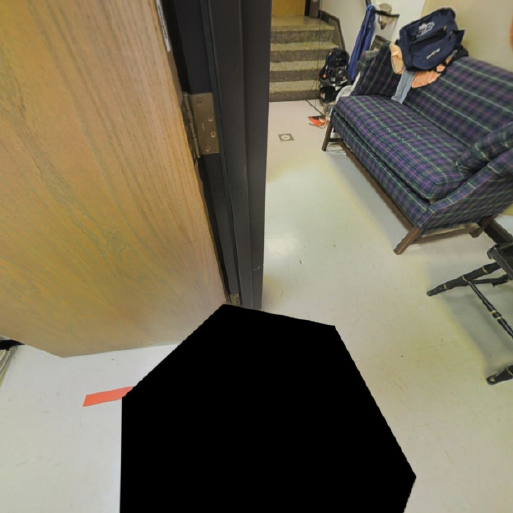}}%
            \fbox{\includegraphics[height=2cm, keepaspectratio, trim={0mm 0mm 0mm 0mm}, clip]{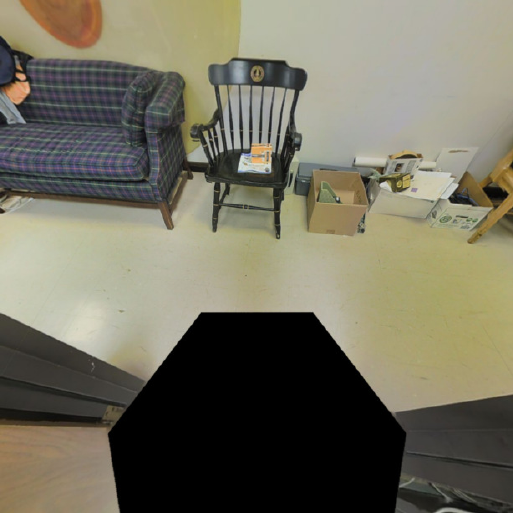}}%
        \end{subfigure}\\
        \vspace{-1pt}
        \begin{subfigure}[b]{\textwidth}
            \centering
            \fbox{\includegraphics[height=2cm, keepaspectratio, trim={0mm 0mm 0mm 0mm}, clip]{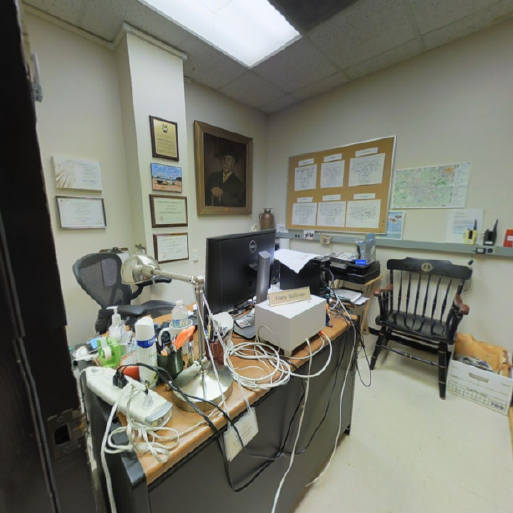}}%
            \fbox{\includegraphics[height=2cm, keepaspectratio, trim={0mm 0mm 0mm 0mm}, clip]{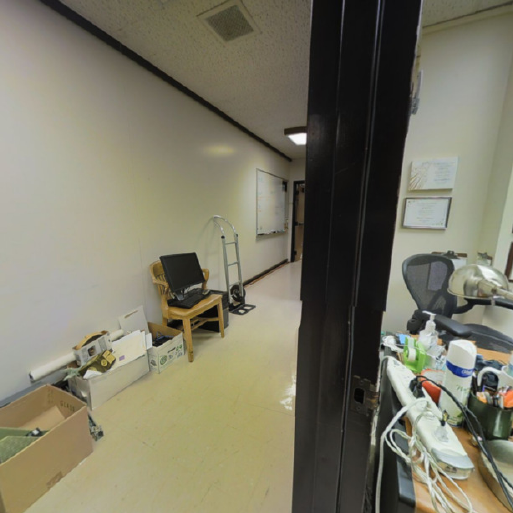}}%
            \fbox{\includegraphics[height=2cm, keepaspectratio, trim={0mm 0mm 0mm 0mm}, clip]{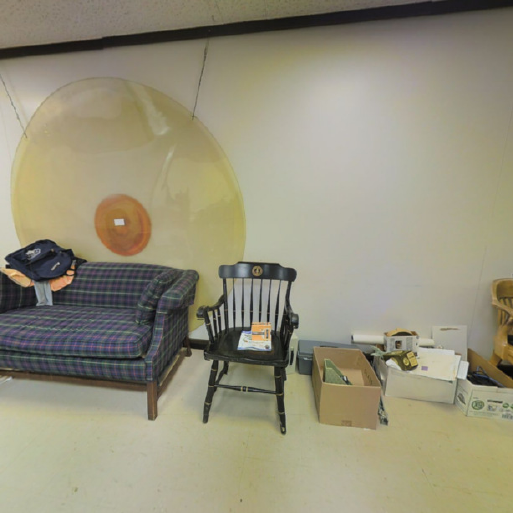}}%
            \fbox{\includegraphics[height=2cm, keepaspectratio, trim={0mm 0mm 0mm 0mm}, clip]{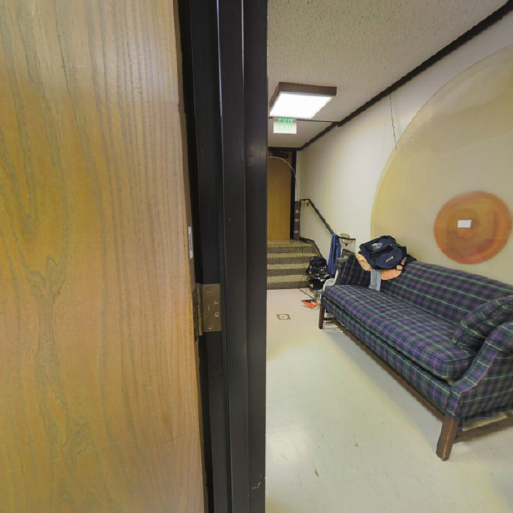}}%
            \fbox{\includegraphics[height=2cm, keepaspectratio, trim={0mm 0mm 0mm 0mm}, clip]{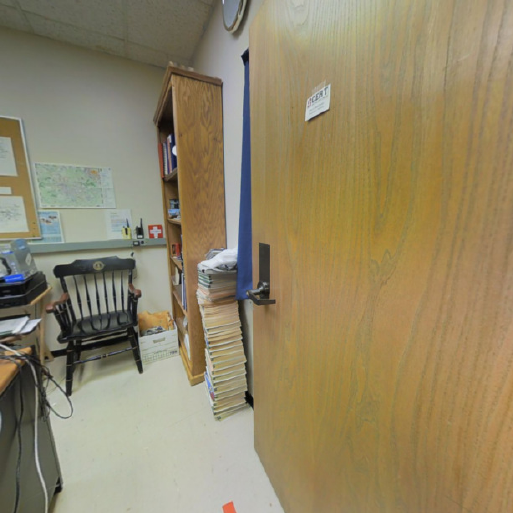}}%
        \end{subfigure}\\
        \vspace{-1pt}
        \begin{subfigure}[b]{\textwidth}
            \centering
            \fbox{\includegraphics[height=2cm, keepaspectratio, trim={0mm 0mm 0mm 0mm}, clip]{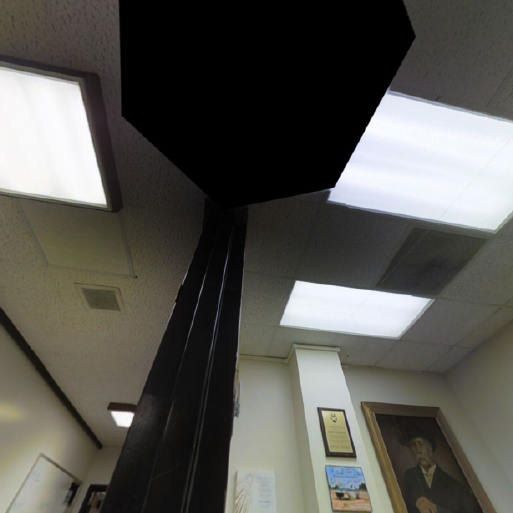}}%
            \fbox{\includegraphics[height=2cm, keepaspectratio, trim={0mm 0mm 0mm 0mm}, clip]{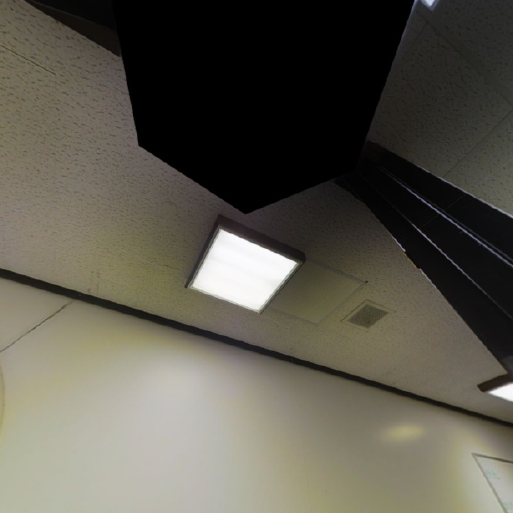}}%
            \fbox{\includegraphics[height=2cm, keepaspectratio, trim={0mm 0mm 0mm 0mm}, clip]{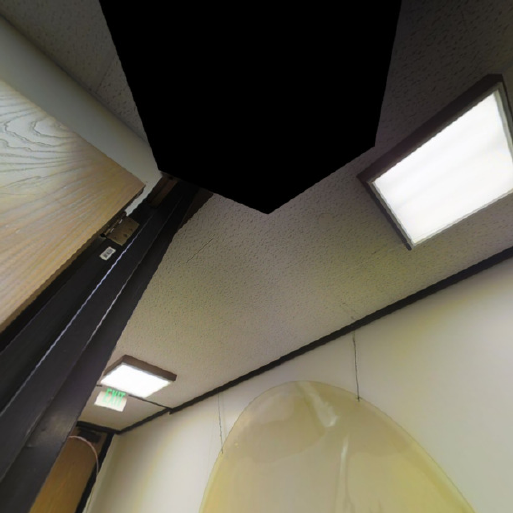}}%
            \fbox{\includegraphics[height=2cm, keepaspectratio, trim={0mm 0mm 0mm 0mm}, clip]{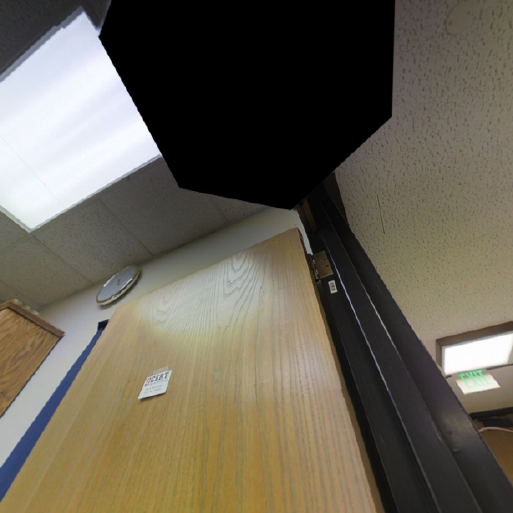}}%
            \fbox{\includegraphics[height=2cm, keepaspectratio, trim={0mm 0mm 0mm 0mm}, clip]{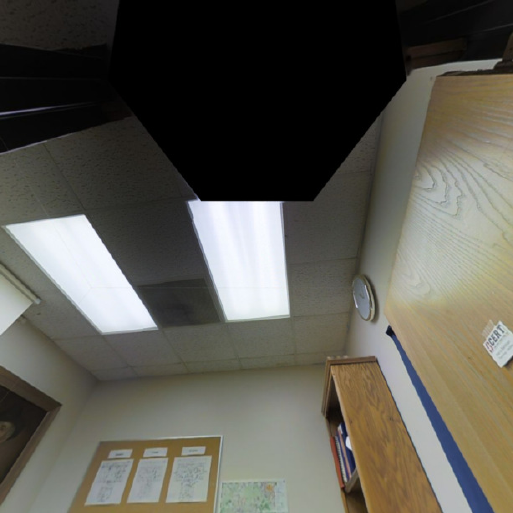}}%
        \end{subfigure}\\
        \vspace{-1pt}
        \begin{subfigure}[b]{\textwidth}
            \centering
            \fbox{\includegraphics[height=2cm, keepaspectratio, trim={0mm 0mm 0mm 0mm}, clip]{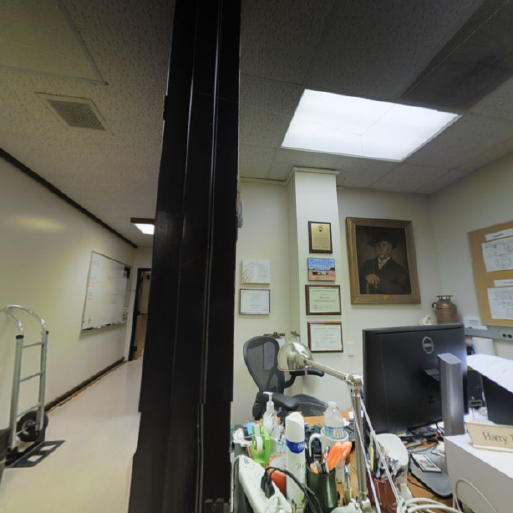}}%
            \fbox{\includegraphics[height=2cm, keepaspectratio, trim={0mm 0mm 0mm 0mm}, clip]{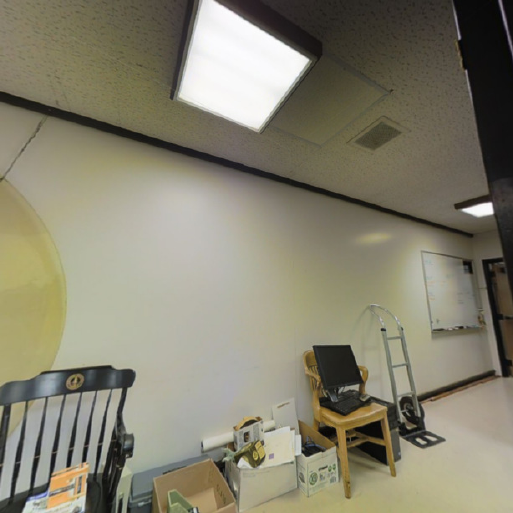}}%
            \fbox{\includegraphics[height=2cm, keepaspectratio, trim={0mm 0mm 0mm 0mm}, clip]{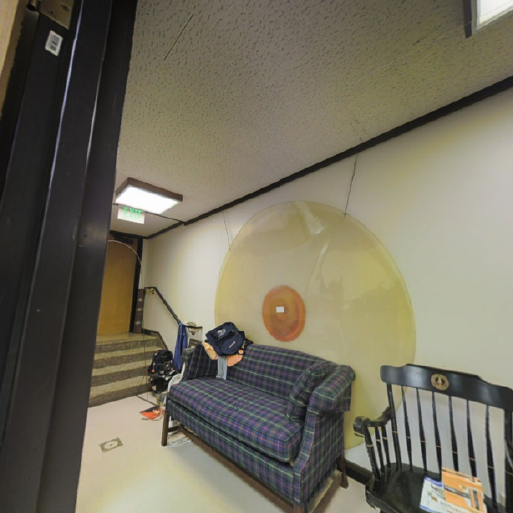}}%
            \fbox{\includegraphics[height=2cm, keepaspectratio, trim={0mm 0mm 0mm 0mm}, clip]{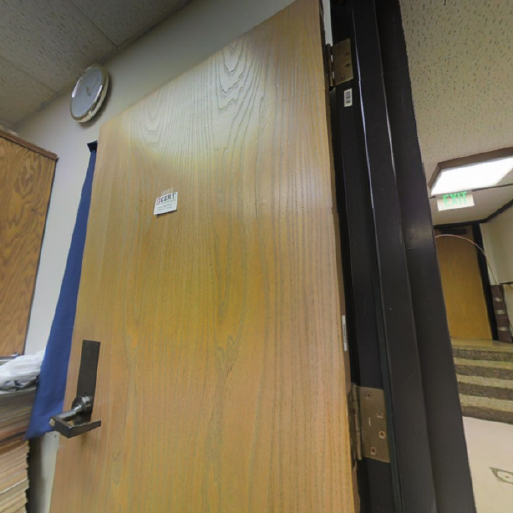}}%
            \fbox{\includegraphics[height=2cm, keepaspectratio, trim={0mm 0mm 0mm 0mm}, clip]{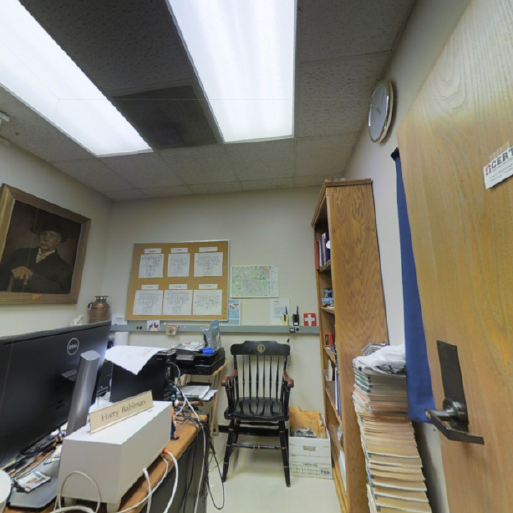}}%
        \end{subfigure}
        \caption{\small Base level 0}
    \end{subfigure}\\
    
    \vspace{4mm}
    \begin{subfigure}[b]{\textwidth}
        \centering
        \begin{subfigure}[b]{\textwidth}
            \centering
            \fbox{\includegraphics[height=1cm, keepaspectratio, trim={0mm 0mm 0mm 0mm}, clip]{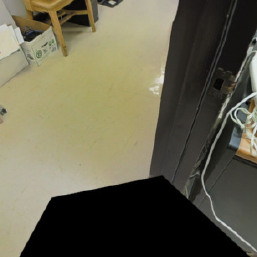}}%
            \fbox{\includegraphics[height=1cm, keepaspectratio, trim={0mm 0mm 0mm 0mm}, clip]{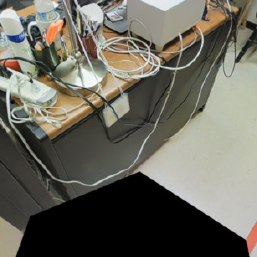}}%
            \fbox{\includegraphics[height=1cm, keepaspectratio, trim={0mm 0mm 0mm 0mm}, clip]{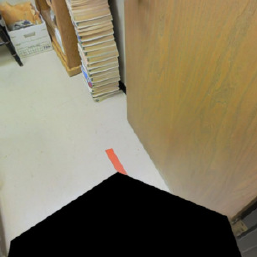}}%
            \fbox{\includegraphics[height=1cm, keepaspectratio, trim={0mm 0mm 0mm 0mm}, clip]{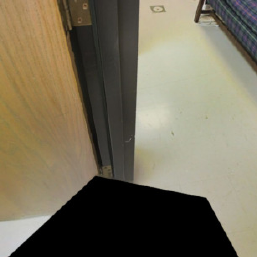}}%
            \fbox{\includegraphics[height=1cm, keepaspectratio, trim={0mm 0mm 0mm 0mm}, clip]{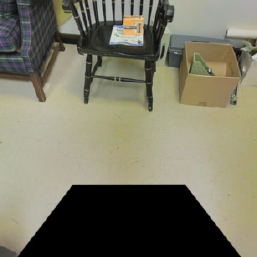}}%
            \fbox{\includegraphics[height=1cm, keepaspectratio, trim={0mm 0mm 0mm 0mm}, clip]{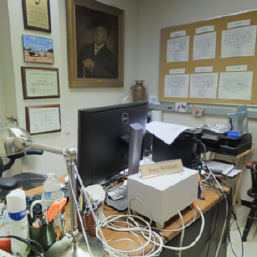}}%
            \fbox{\includegraphics[height=1cm, keepaspectratio, trim={0mm 0mm 0mm 0mm}, clip]{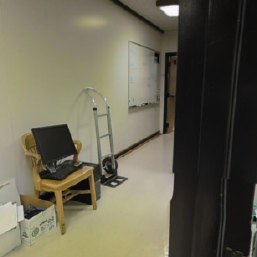}}%
            \fbox{\includegraphics[height=1cm, keepaspectratio, trim={0mm 0mm 0mm 0mm}, clip]{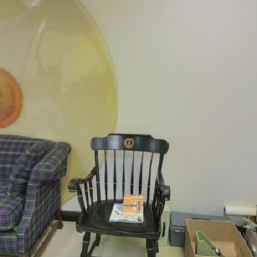}}%
            \fbox{\includegraphics[height=1cm, keepaspectratio, trim={0mm 0mm 0mm 0mm}, clip]{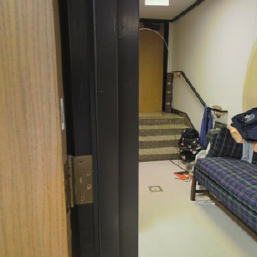}}%
            \fbox{\includegraphics[height=1cm, keepaspectratio, trim={0mm 0mm 0mm 0mm}, clip]{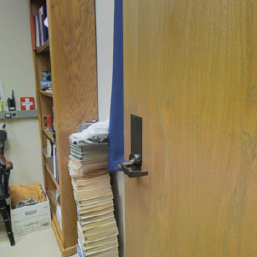}}%
            \fbox{\includegraphics[height=1cm, keepaspectratio, trim={0mm 0mm 0mm 0mm}, clip]{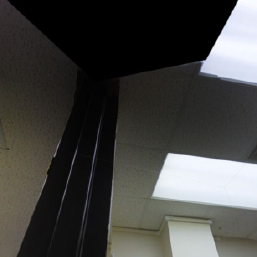}}%
            \fbox{\includegraphics[height=1cm, keepaspectratio, trim={0mm 0mm 0mm 0mm}, clip]{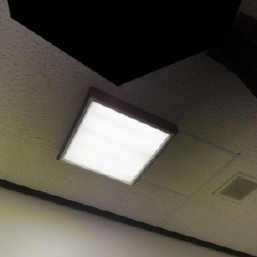}}%
            \fbox{\includegraphics[height=1cm, keepaspectratio, trim={0mm 0mm 0mm 0mm}, clip]{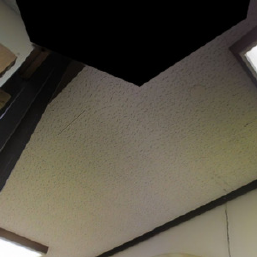}}%
            \fbox{\includegraphics[height=1cm, keepaspectratio, trim={0mm 0mm 0mm 0mm}, clip]{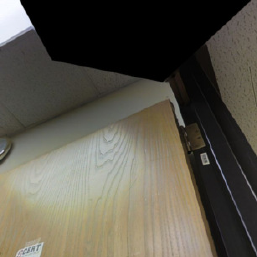}}%
            \fbox{\includegraphics[height=1cm, keepaspectratio, trim={0mm 0mm 0mm 0mm}, clip]{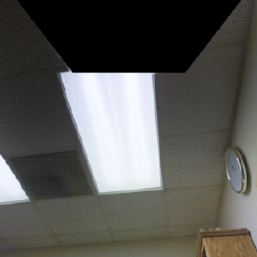}}%
            \fbox{\includegraphics[height=1cm, keepaspectratio, trim={0mm 0mm 0mm 0mm}, clip]{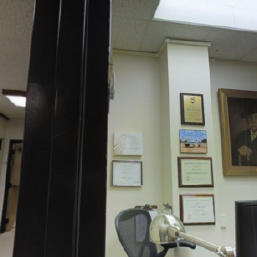}}%
        \end{subfigure}\\
        \vspace{-1pt}
        \begin{subfigure}[b]{\textwidth}
            \centering
            \fbox{\includegraphics[height=1cm, keepaspectratio, trim={0mm 0mm 0mm 0mm}, clip]{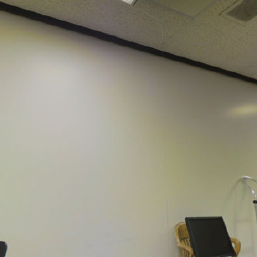}}%
            \fbox{\includegraphics[height=1cm, keepaspectratio, trim={0mm 0mm 0mm 0mm}, clip]{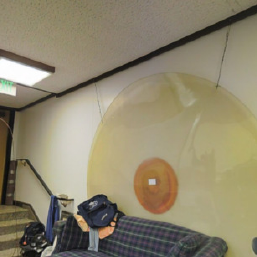}}%
            \fbox{\includegraphics[height=1cm, keepaspectratio, trim={0mm 0mm 0mm 0mm}, clip]{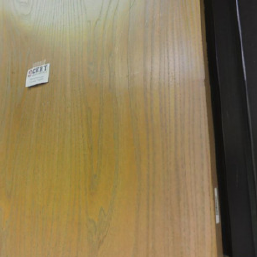}}%
            \fbox{\includegraphics[height=1cm, keepaspectratio, trim={0mm 0mm 0mm 0mm}, clip]{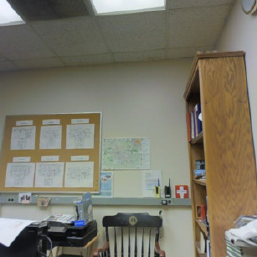}}%
            \fbox{\includegraphics[height=1cm, keepaspectratio, trim={0mm 0mm 0mm 0mm}, clip]{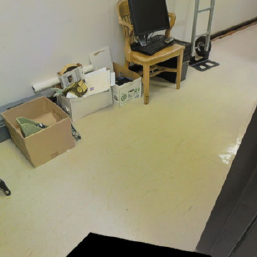}}%
            \fbox{\includegraphics[height=1cm, keepaspectratio, trim={0mm 0mm 0mm 0mm}, clip]{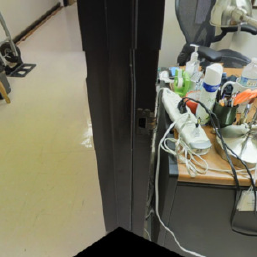}}%
            \fbox{\includegraphics[height=1cm, keepaspectratio, trim={0mm 0mm 0mm 0mm}, clip]{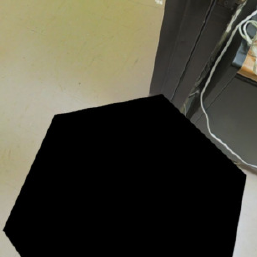}}%
            \fbox{\includegraphics[height=1cm, keepaspectratio, trim={0mm 0mm 0mm 0mm}, clip]{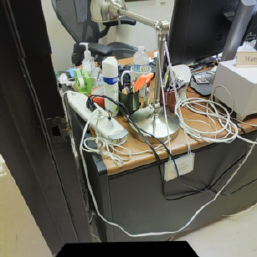}}%
            \fbox{\includegraphics[height=1cm, keepaspectratio, trim={0mm 0mm 0mm 0mm}, clip]{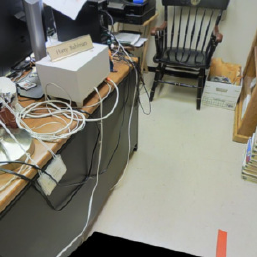}}%
            \fbox{\includegraphics[height=1cm, keepaspectratio, trim={0mm 0mm 0mm 0mm}, clip]{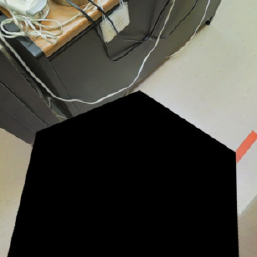}}%
            \fbox{\includegraphics[height=1cm, keepaspectratio, trim={0mm 0mm 0mm 0mm}, clip]{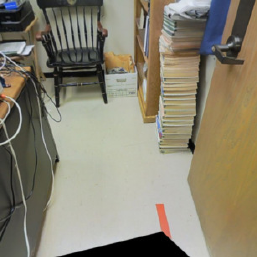}}%
            \fbox{\includegraphics[height=1cm, keepaspectratio, trim={0mm 0mm 0mm 0mm}, clip]{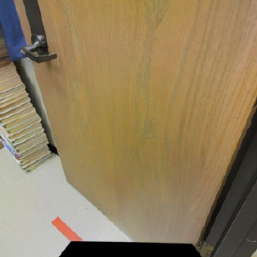}}%
            \fbox{\includegraphics[height=1cm, keepaspectratio, trim={0mm 0mm 0mm 0mm}, clip]{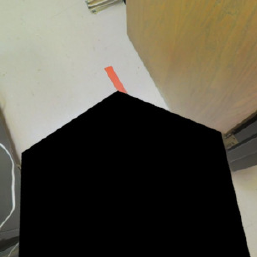}}%
            \fbox{\includegraphics[height=1cm, keepaspectratio, trim={0mm 0mm 0mm 0mm}, clip]{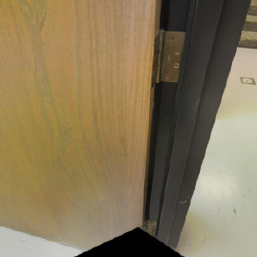}}%
            \fbox{\includegraphics[height=1cm, keepaspectratio, trim={0mm 0mm 0mm 0mm}, clip]{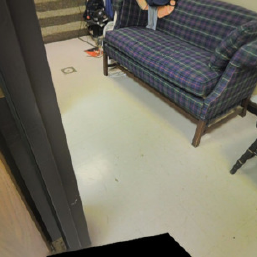}}%
            \fbox{\includegraphics[height=1cm, keepaspectratio, trim={0mm 0mm 0mm 0mm}, clip]{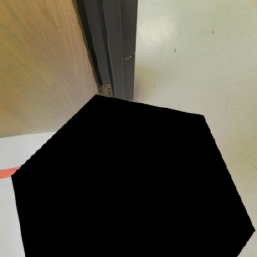}}%
        \end{subfigure}\\
        \vspace{-1pt}
        \begin{subfigure}[b]{\textwidth}
            \centering
            \fbox{\includegraphics[height=1cm, keepaspectratio, trim={0mm 0mm 0mm 0mm}, clip]{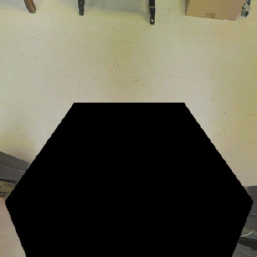}}%
            \fbox{\includegraphics[height=1cm, keepaspectratio, trim={0mm 0mm 0mm 0mm}, clip]{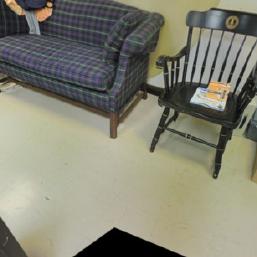}}%
            \fbox{\includegraphics[height=1cm, keepaspectratio, trim={0mm 0mm 0mm 0mm}, clip]{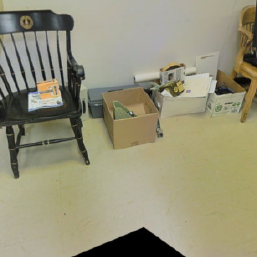}}%
            \fbox{\includegraphics[height=1cm, keepaspectratio, trim={0mm 0mm 0mm 0mm}, clip]{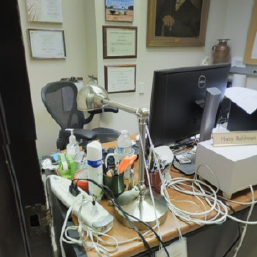}}%
            \fbox{\includegraphics[height=1cm, keepaspectratio, trim={0mm 0mm 0mm 0mm}, clip]{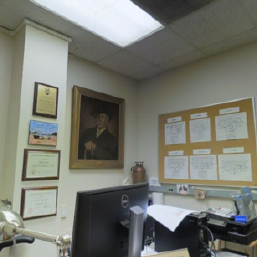}}%
            \fbox{\includegraphics[height=1cm, keepaspectratio, trim={0mm 0mm 0mm 0mm}, clip]{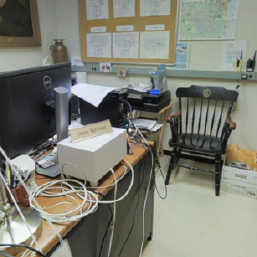}}%
            \fbox{\includegraphics[height=1cm, keepaspectratio, trim={0mm 0mm 0mm 0mm}, clip]{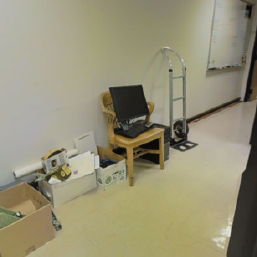}}%
            \fbox{\includegraphics[height=1cm, keepaspectratio, trim={0mm 0mm 0mm 0mm}, clip]{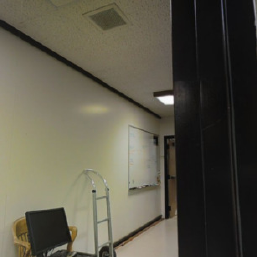}}%
            \fbox{\includegraphics[height=1cm, keepaspectratio, trim={0mm 0mm 0mm 0mm}, clip]{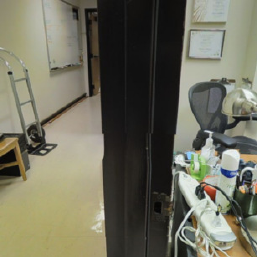}}%
            \fbox{\includegraphics[height=1cm, keepaspectratio, trim={0mm 0mm 0mm 0mm}, clip]{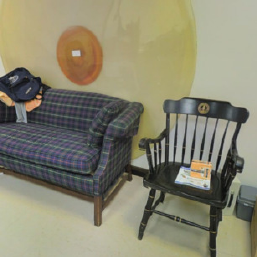}}%
            \fbox{\includegraphics[height=1cm, keepaspectratio, trim={0mm 0mm 0mm 0mm}, clip]{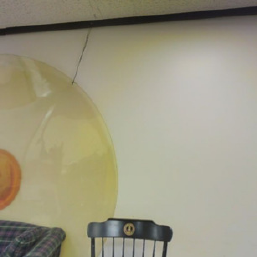}}%
            \fbox{\includegraphics[height=1cm, keepaspectratio, trim={0mm 0mm 0mm 0mm}, clip]{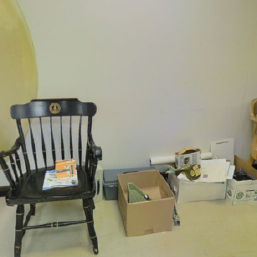}}%
            \fbox{\includegraphics[height=1cm, keepaspectratio, trim={0mm 0mm 0mm 0mm}, clip]{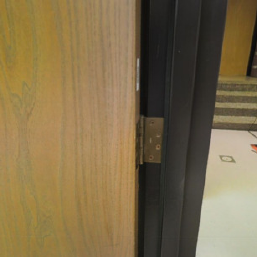}}%
            \fbox{\includegraphics[height=1cm, keepaspectratio, trim={0mm 0mm 0mm 0mm}, clip]{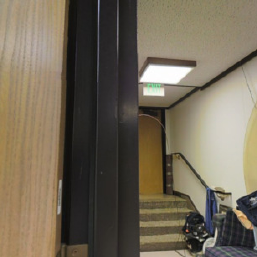}}%
            \fbox{\includegraphics[height=1cm, keepaspectratio, trim={0mm 0mm 0mm 0mm}, clip]{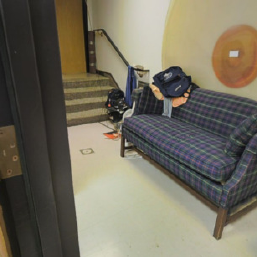}}%
            \fbox{\includegraphics[height=1cm, keepaspectratio, trim={0mm 0mm 0mm 0mm}, clip]{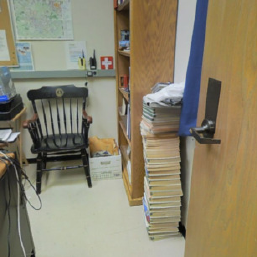}}%
        \end{subfigure}\\
        \vspace{-1pt}
        \begin{subfigure}[b]{\textwidth}
            \centering
            \fbox{\includegraphics[height=1cm, keepaspectratio, trim={0mm 0mm 0mm 0mm}, clip]{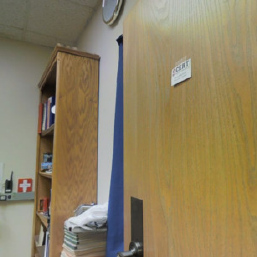}}%
            \fbox{\includegraphics[height=1cm, keepaspectratio, trim={0mm 0mm 0mm 0mm}, clip]{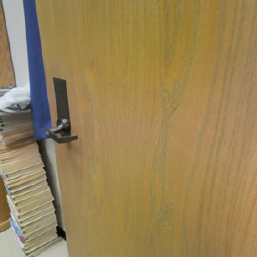}}%
            \fbox{\includegraphics[height=1cm, keepaspectratio, trim={0mm 0mm 0mm 0mm}, clip]{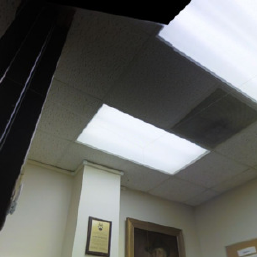}}%
            \fbox{\includegraphics[height=1cm, keepaspectratio, trim={0mm 0mm 0mm 0mm}, clip]{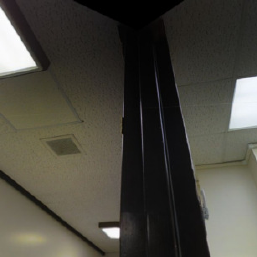}}%
            \fbox{\includegraphics[height=1cm, keepaspectratio, trim={0mm 0mm 0mm 0mm}, clip]{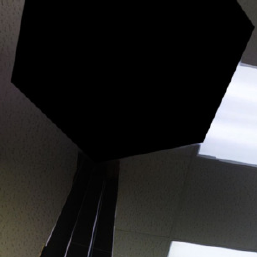}}%
            \fbox{\includegraphics[height=1cm, keepaspectratio, trim={0mm 0mm 0mm 0mm}, clip]{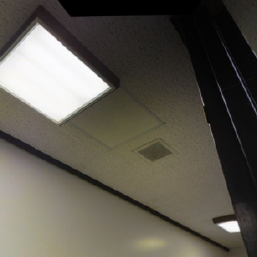}}%
            \fbox{\includegraphics[height=1cm, keepaspectratio, trim={0mm 0mm 0mm 0mm}, clip]{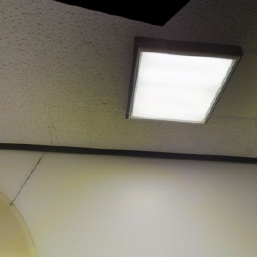}}%
            \fbox{\includegraphics[height=1cm, keepaspectratio, trim={0mm 0mm 0mm 0mm}, clip]{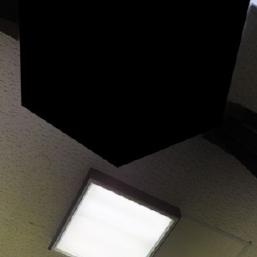}}%
            \fbox{\includegraphics[height=1cm, keepaspectratio, trim={0mm 0mm 0mm 0mm}, clip]{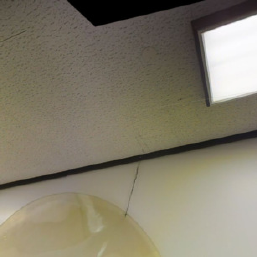}}%
            \fbox{\includegraphics[height=1cm, keepaspectratio, trim={0mm 0mm 0mm 0mm}, clip]{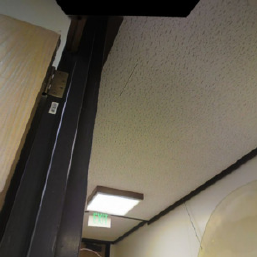}}%
            \fbox{\includegraphics[height=1cm, keepaspectratio, trim={0mm 0mm 0mm 0mm}, clip]{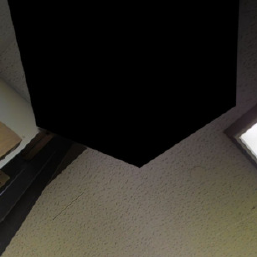}}%
            \fbox{\includegraphics[height=1cm, keepaspectratio, trim={0mm 0mm 0mm 0mm}, clip]{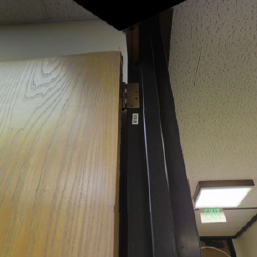}}%
            \fbox{\includegraphics[height=1cm, keepaspectratio, trim={0mm 0mm 0mm 0mm}, clip]{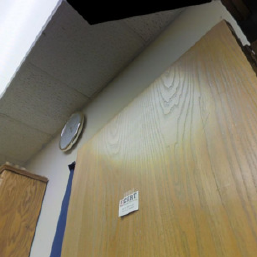}}%
            \fbox{\includegraphics[height=1cm, keepaspectratio, trim={0mm 0mm 0mm 0mm}, clip]{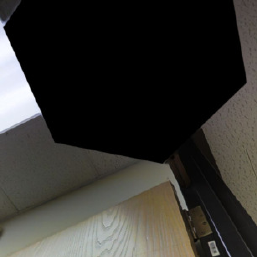}}%
            \fbox{\includegraphics[height=1cm, keepaspectratio, trim={0mm 0mm 0mm 0mm}, clip]{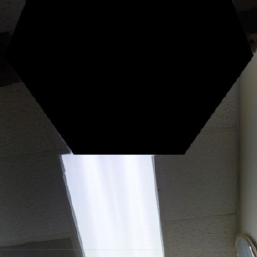}}%
            \fbox{\includegraphics[height=1cm, keepaspectratio, trim={0mm 0mm 0mm 0mm}, clip]{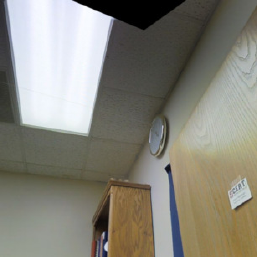}}%
        \end{subfigure}\\
        \vspace{-1pt}
        \begin{subfigure}[b]{\textwidth}
            \centering
            \fbox{\includegraphics[height=1cm, keepaspectratio, trim={0mm 0mm 0mm 0mm}, clip]{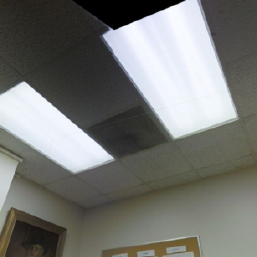}}%
            \fbox{\includegraphics[height=1cm, keepaspectratio, trim={0mm 0mm 0mm 0mm}, clip]{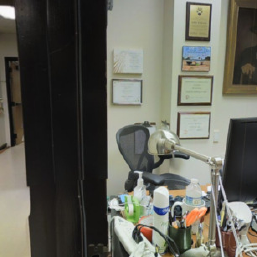}}%
            \fbox{\includegraphics[height=1cm, keepaspectratio, trim={0mm 0mm 0mm 0mm}, clip]{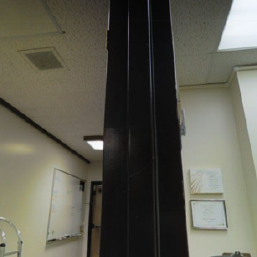}}%
            \fbox{\includegraphics[height=1cm, keepaspectratio, trim={0mm 0mm 0mm 0mm}, clip]{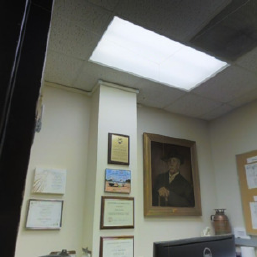}}%
            \fbox{\includegraphics[height=1cm, keepaspectratio, trim={0mm 0mm 0mm 0mm}, clip]{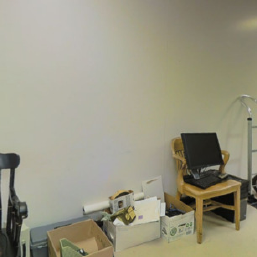}}%
            \fbox{\includegraphics[height=1cm, keepaspectratio, trim={0mm 0mm 0mm 0mm}, clip]{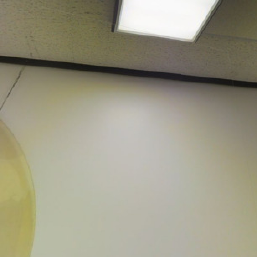}}%
            \fbox{\includegraphics[height=1cm, keepaspectratio, trim={0mm 0mm 0mm 0mm}, clip]{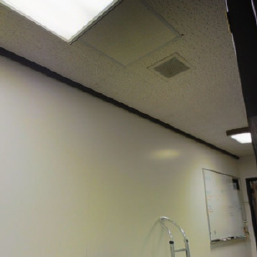}}%
            \fbox{\includegraphics[height=1cm, keepaspectratio, trim={0mm 0mm 0mm 0mm}, clip]{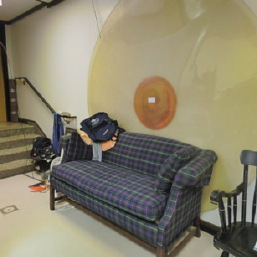}}%
            \fbox{\includegraphics[height=1cm, keepaspectratio, trim={0mm 0mm 0mm 0mm}, clip]{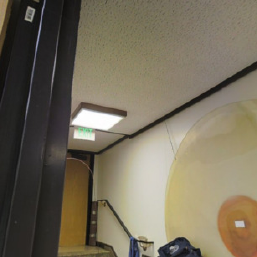}}%
            \fbox{\includegraphics[height=1cm, keepaspectratio, trim={0mm 0mm 0mm 0mm}, clip]{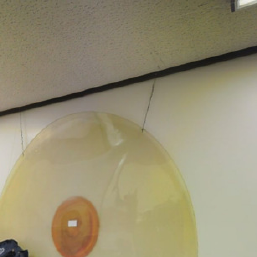}}%
            \fbox{\includegraphics[height=1cm, keepaspectratio, trim={0mm 0mm 0mm 0mm}, clip]{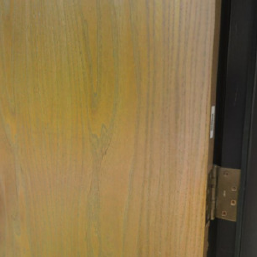}}%
            \fbox{\includegraphics[height=1cm, keepaspectratio, trim={0mm 0mm 0mm 0mm}, clip]{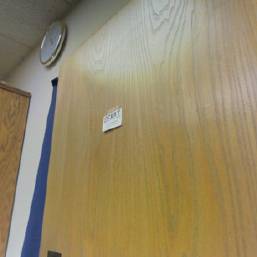}}%
            \fbox{\includegraphics[height=1cm, keepaspectratio, trim={0mm 0mm 0mm 0mm}, clip]{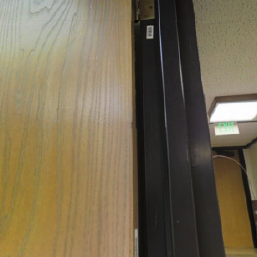}}%
            \fbox{\includegraphics[height=1cm, keepaspectratio, trim={0mm 0mm 0mm 0mm}, clip]{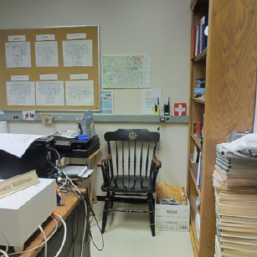}}%
            \fbox{\includegraphics[height=1cm, keepaspectratio, trim={0mm 0mm 0mm 0mm}, clip]{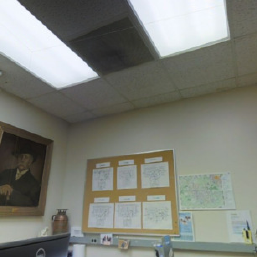}}%
            \fbox{\includegraphics[height=1cm, keepaspectratio, trim={0mm 0mm 0mm 0mm}, clip]{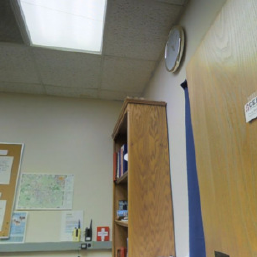}}%
        \end{subfigure}\\
        \caption{\small Base level 1}
    \end{subfigure}
    \caption{\small Example of tangent images at base levels 0 and 1.}
\label{fig:tangentimages}
\end{figure*}

%% file: supplementary/figures/qual-sem-seg.tex
\begin{figure*}
    \centering

    \begin{subfigure}[b]{0.3\textwidth}
        \centering
        \includegraphics[width=1.0\linewidth, trim={0mm 0mm 0mm 0mm}, clip]{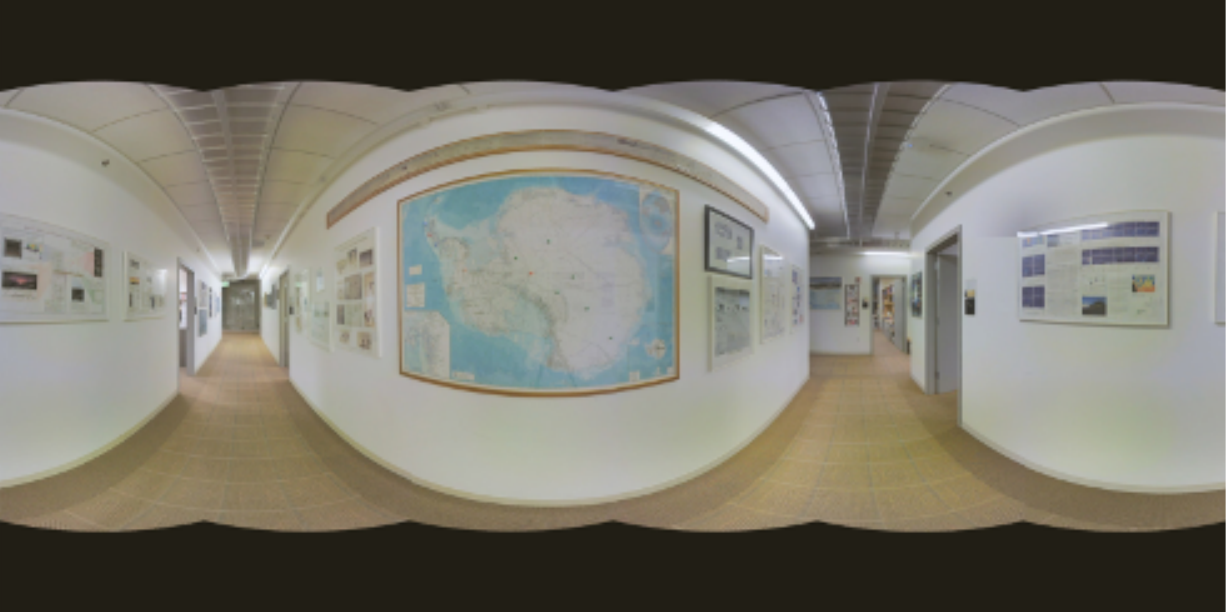}
        \caption*{\small Input}
    \end{subfigure}
    ~
    \begin{subfigure}[b]{0.3\textwidth}
        \centering
        \includegraphics[width=1.0\linewidth, trim={0mm 0mm 0mm 0mm}, clip]{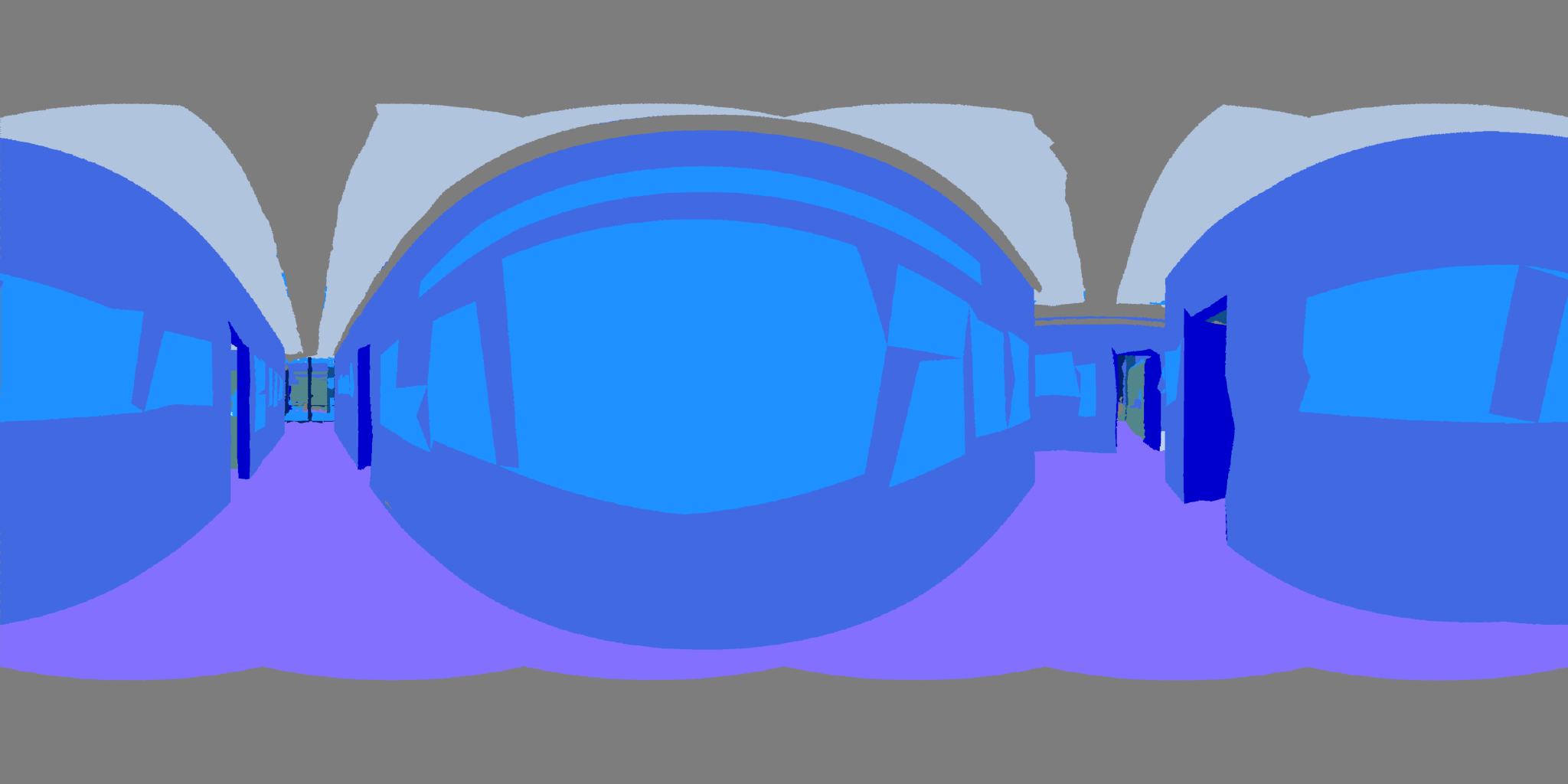}
        \caption*{\small GT}
    \end{subfigure}\\
    \vspace{1mm}
    \begin{subfigure}[b]{0.3\textwidth}
        \centering
        \includegraphics[width=1.0\linewidth, trim={0mm 0mm 0mm 0mm}, clip]{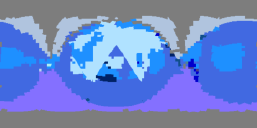}
        \caption*{\small L5}
    \end{subfigure}
    ~
    \begin{subfigure}[b]{0.3\textwidth}
        \centering
        \includegraphics[width=1.0\linewidth, trim={0mm 0mm 0mm 0mm}, clip]{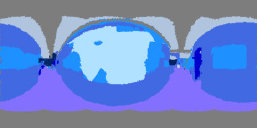}
        \caption*{\small L7}
    \end{subfigure}
    ~
    \begin{subfigure}[b]{0.3\textwidth}
        \centering
        \includegraphics[width=1.0\linewidth, trim={0mm 0mm 0mm 0mm}, clip]{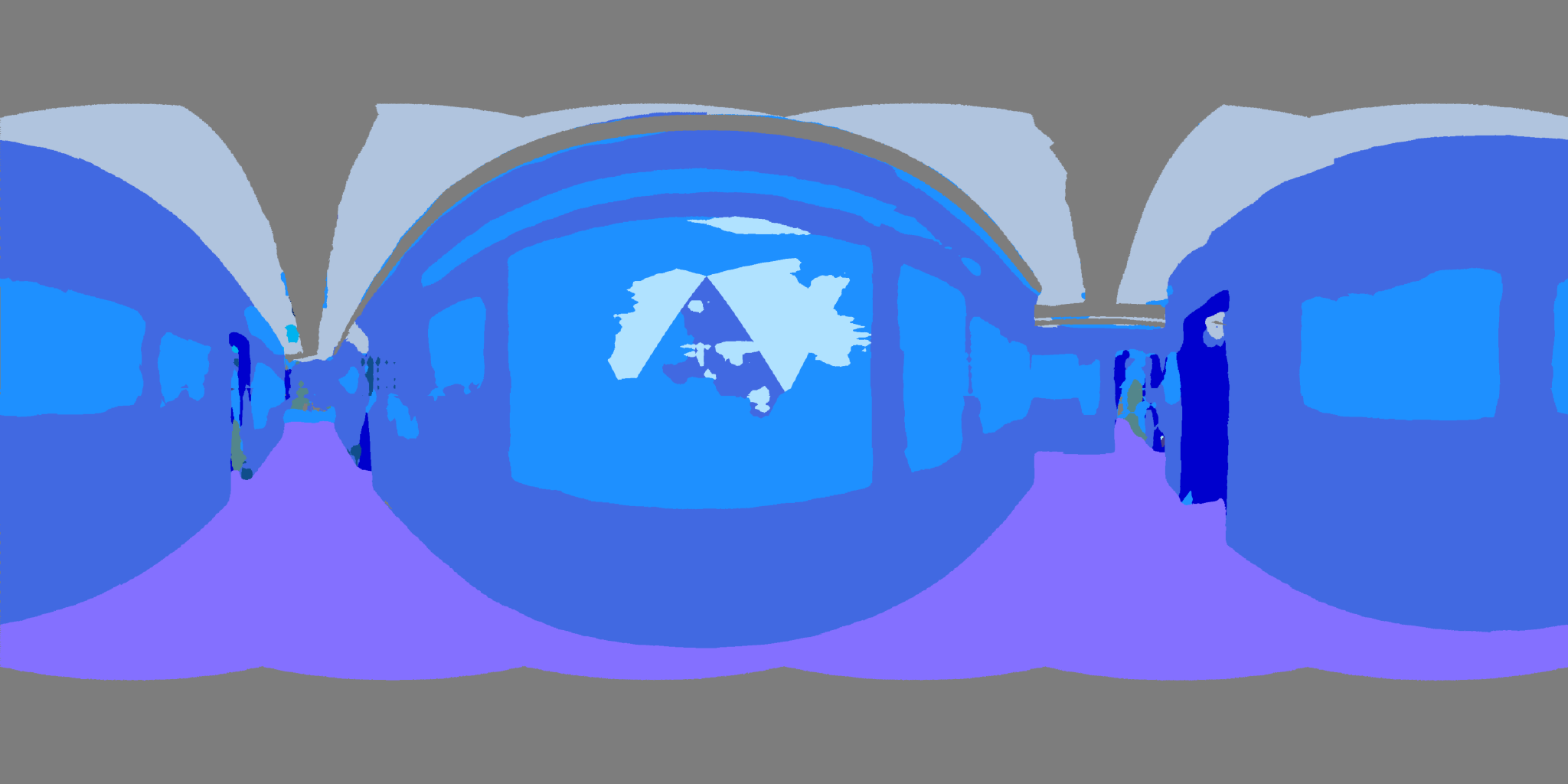}
        \caption*{\small L10}
    \end{subfigure}\\
    \vspace{4mm}
    \begin{subfigure}[b]{0.3\textwidth}
        \centering
        \includegraphics[width=1.0\linewidth, trim={0mm 0mm 0mm 0mm}, clip]{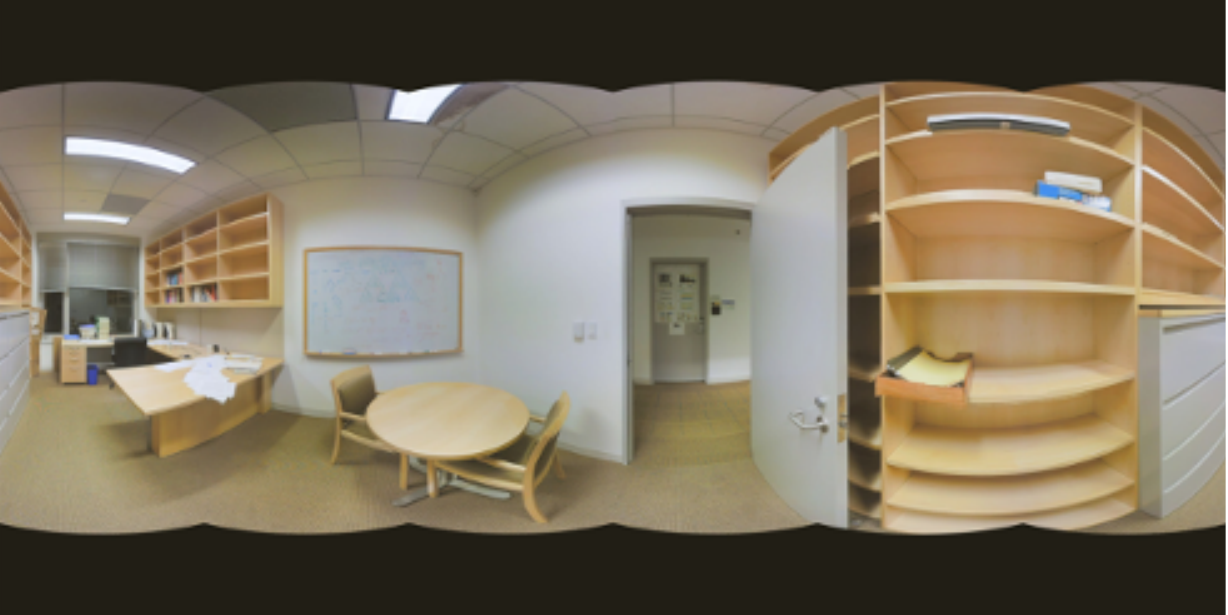}
        \caption*{\small Input}
    \end{subfigure}
    ~
    \begin{subfigure}[b]{0.3\textwidth}
        \centering
        \includegraphics[width=1.0\linewidth, trim={0mm 0mm 0mm 0mm}, clip]{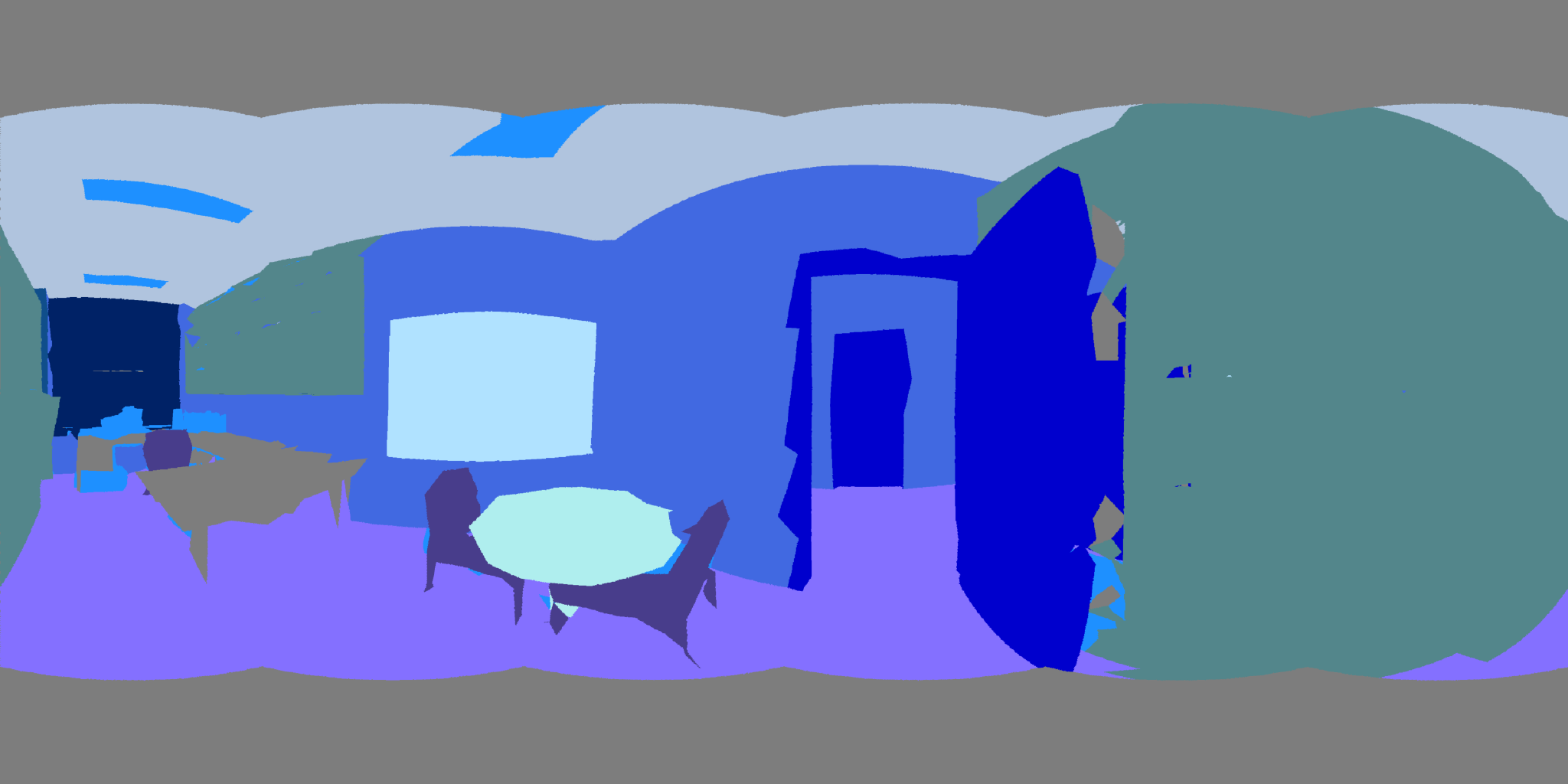}
        \caption*{\small GT}
    \end{subfigure}\\
    \vspace{1mm}
    \begin{subfigure}[b]{0.3\textwidth}
        \centering
        \includegraphics[width=1.0\linewidth, trim={0mm 0mm 0mm 0mm}, clip]{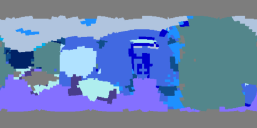}
        \caption*{\small L5}
    \end{subfigure}
    ~
    \begin{subfigure}[b]{0.3\textwidth}
        \centering
        \includegraphics[width=1.0\linewidth, trim={0mm 0mm 0mm 0mm}, clip]{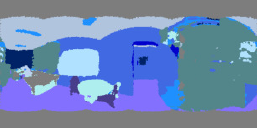}
        \caption*{\small L7}
    \end{subfigure}
    ~
    \begin{subfigure}[b]{0.3\textwidth}
        \centering
        \includegraphics[width=1.0\linewidth, trim={0mm 0mm 0mm 0mm}, clip]{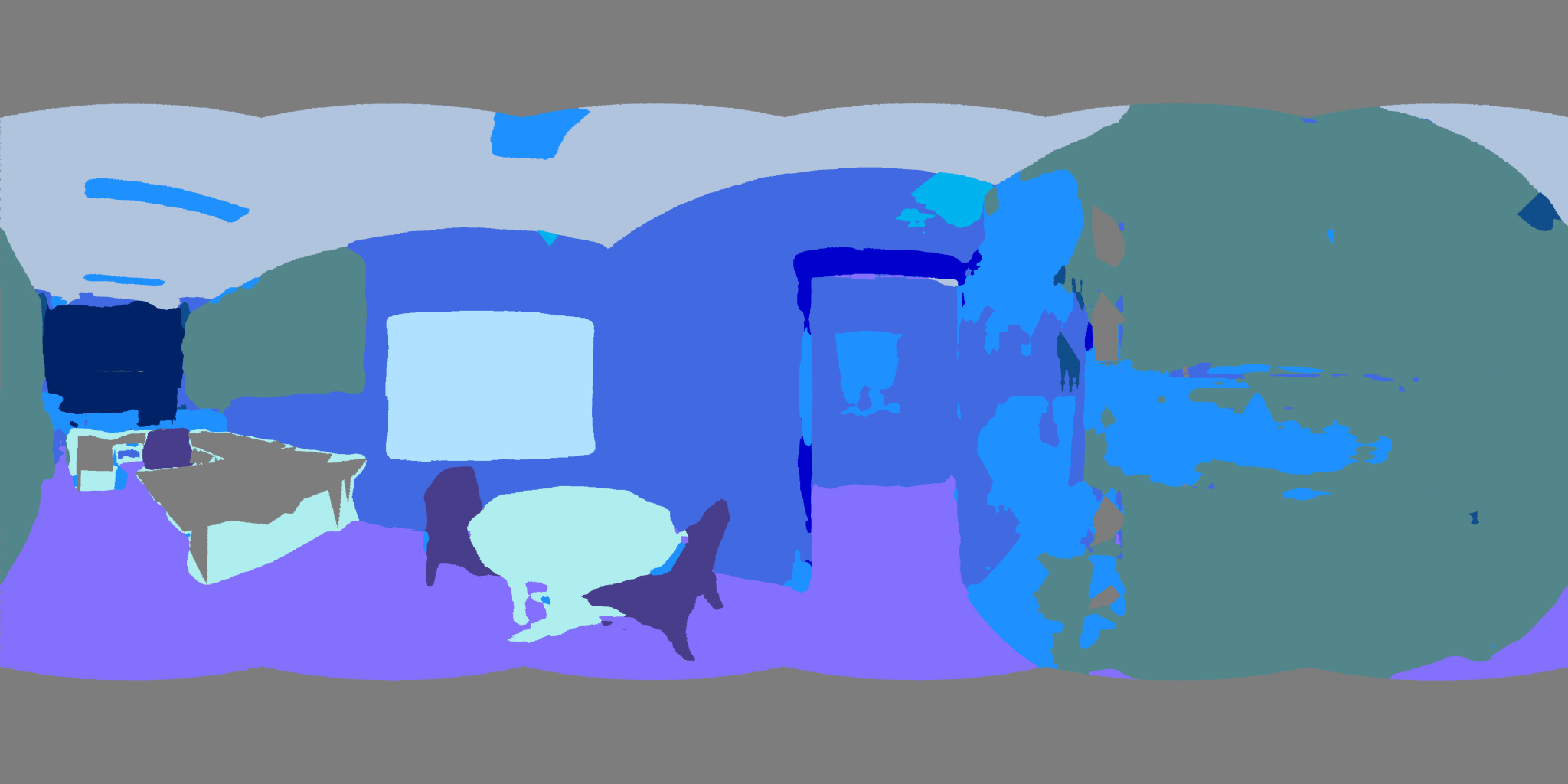}
        \caption*{\small L10}
    \end{subfigure}\\
    \vspace{4mm}
    \begin{subfigure}[b]{0.3\textwidth}
        \centering
        \includegraphics[width=1.0\linewidth, trim={0mm 0mm 0mm 0mm}, clip]{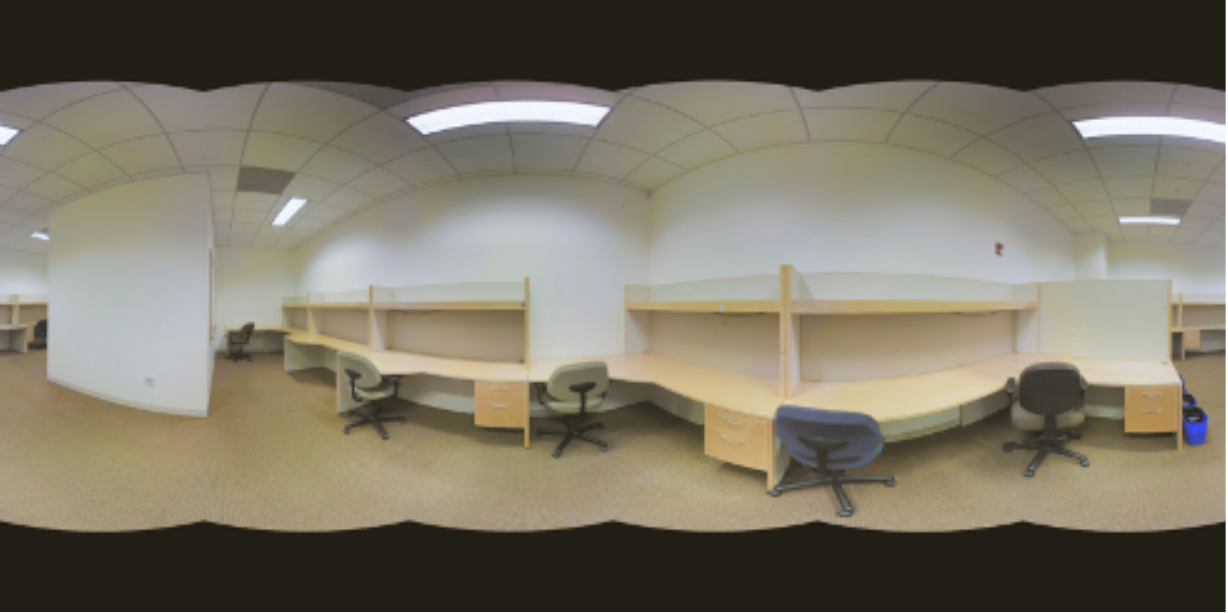}
        \caption*{\small Input}
    \end{subfigure}
    ~
    \begin{subfigure}[b]{0.3\textwidth}
        \centering
        \includegraphics[width=1.0\linewidth, trim={0mm 0mm 0mm 0mm}, clip]{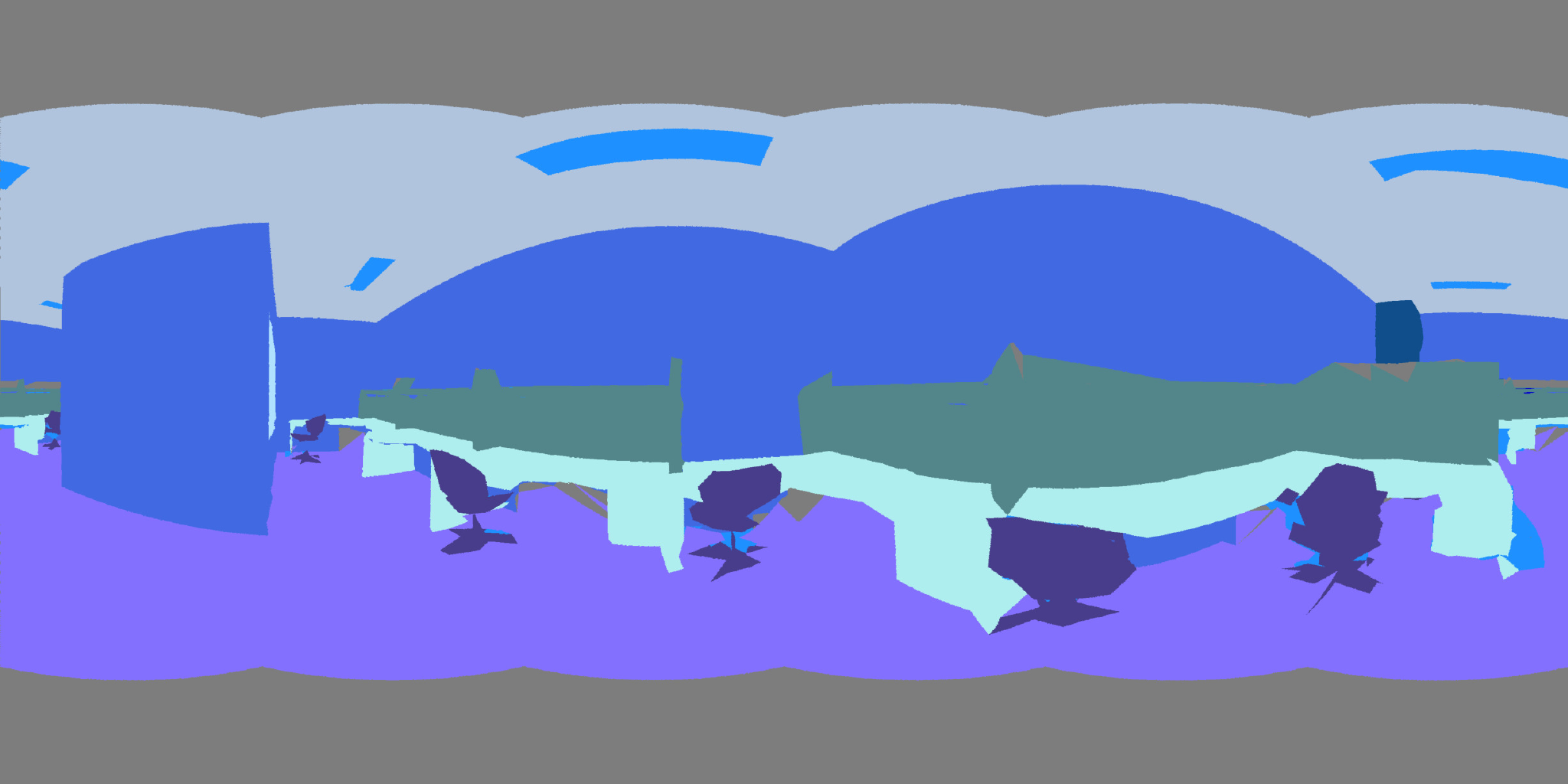}
        \caption*{\small GT}
    \end{subfigure}\\
    \vspace{1mm}
    \begin{subfigure}[b]{0.3\textwidth}
        \centering
        \includegraphics[width=1.0\linewidth, trim={0mm 0mm 0mm 0mm}, clip]{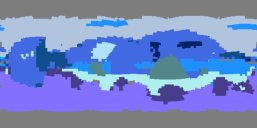}
        \caption*{\small L5}
    \end{subfigure}
    ~
    \begin{subfigure}[b]{0.3\textwidth}
        \centering
        \includegraphics[width=1.0\linewidth, trim={0mm 0mm 0mm 0mm}, clip]{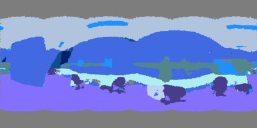}
        \caption*{\small L7}
    \end{subfigure}
    ~
    \begin{subfigure}[b]{0.3\textwidth}
        \centering
        \includegraphics[width=1.0\linewidth, trim={0mm 0mm 0mm 0mm}, clip]{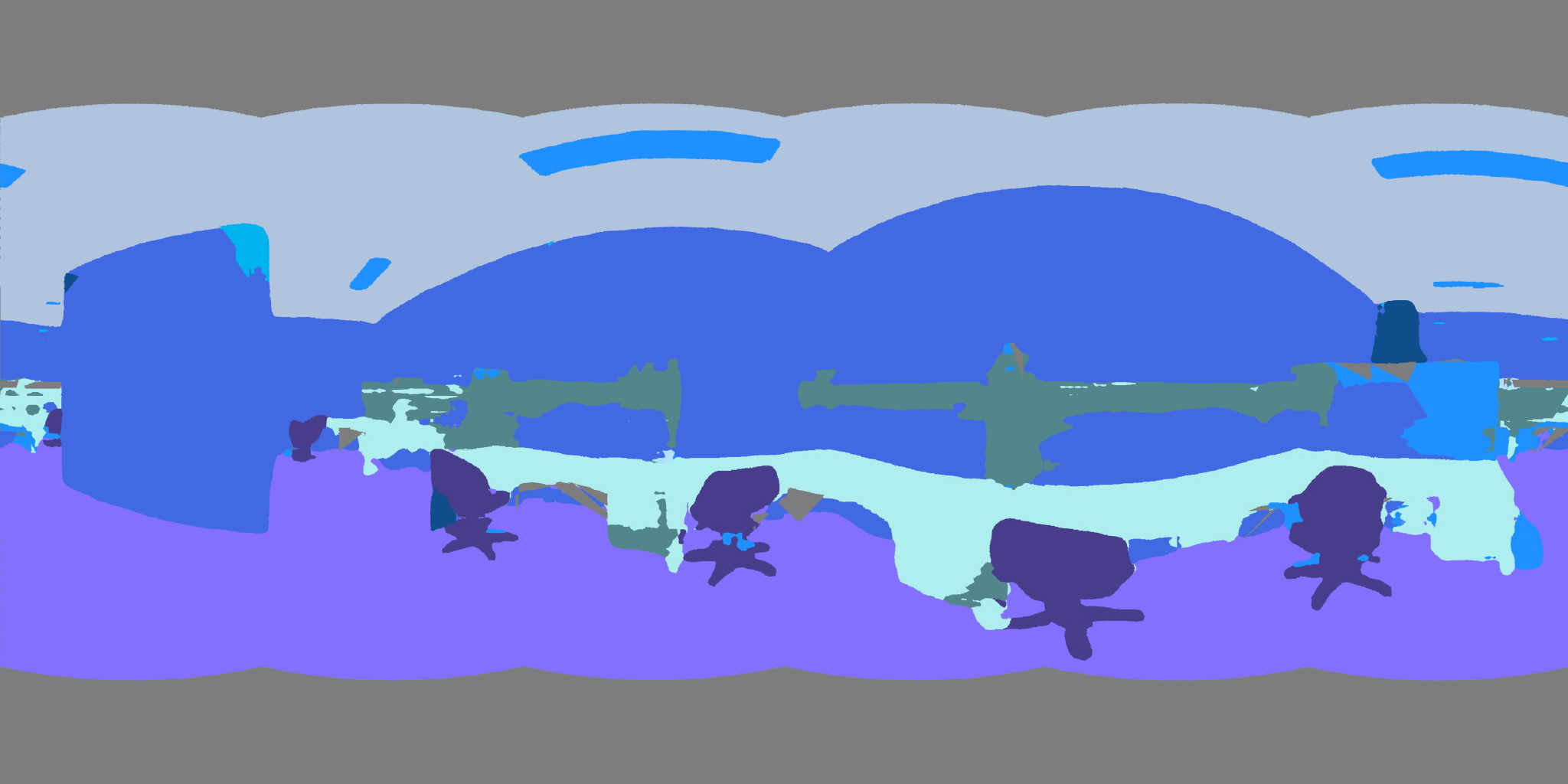}
        \caption*{\small L10}
    \end{subfigure}\\
    \vspace{4mm}
\caption{\small Qualitative results of semantic segmentation on the Stanford 2D3DS dataset \cite{armeni2017joint} at different input resolutions. These results illustrate the performance gains we access by being able to scale to high resolution spherical inputs.}
\label{fig:semsegqual}
\end{figure*}

%% file: supplementary/tables/keypoint-dataset.tex
\begin{table*}[t]
    \small
    \centering
    \begin{tabular}{|c|l|l|c|}
    \hline
    \multicolumn{4}{|c|}{\textbf{Easy}} \\
    \hline
    \textit{Pair ID} & \textit{Left Image} & \textit{Right Image} & \textit{FOV Overlap} \\
    \hline
    1 & 0f65e09\_hallway\_7 & 1ebdfef\_hallway\_7 & 0.87 \\ 
    2 & 08a99a5\_hallway\_7 & 251a331\_hallway\_7 & 0.97 \\ 
    3 & 08a99a5\_hallway\_7 & f7c6c2a\_hallway\_7 & 0.89 \\ 
    4 & 251a331\_hallway\_7 & f7c6c2a\_hallway\_7 & 0.87 \\ 
    5 & 251a331\_hallway\_7 & b261c3b\_hallway\_7 & 0.97 \\ 
    10 & f7c6c2a\_hallway\_7 & b261c3b\_hallway\_7 & 0.89 \\ 
    20 & 9178f6a\_hallway\_6 & 29abbc1\_hallway\_6 & 0.89 \\ 
    23 & bc12865\_hallway\_6 & 7d58331\_hallway\_6 & 0.88 \\ 
    24 & ee20957\_hallway\_6 & bed890d\_hallway\_6 & 0.86 \\ 
    25 & ee20957\_hallway\_6 & eaba8c8\_hallway\_6 & 1.00 \\ 
    28 & bed890d\_hallway\_6 & eaba8c8\_hallway\_6 & 0.86 \\ 
    29 & 077f181\_hallway\_6 & 83baa70\_hallway\_6 & 0.86 \\ 
    30 & 97ab30c\_hallway\_6 & eaba8c8\_hallway\_6 & 0.86 \\ 
    31 & fc19236\_office\_18 & e7d9e58\_office\_18 & 0.88 \\ 
    34 & 09ad38a\_office\_26 & 04a59ce\_office\_26 & 0.96 \\ 
    35 & 04a59ce\_office\_26 & c16a90f\_office\_26 & 0.96 \\ 
    37 & c40ca55\_office\_31 & 7b74e08\_office\_31 & 0.87 \\ 
    39 & 4a41b27\_office\_31 & 7b74e08\_office\_31 & 0.87 \\ 
    43 & 5512025\_office\_23 & 7f04c9b\_office\_23 & 0.92 \\ 
    44 & 5512025\_office\_23 & 5a18aa0\_office\_23 & 0.86 \\ 
    45 & 7f04c9b\_office\_23 & 5a18aa0\_office\_23 & 0.87 \\ 
    47 & 433548f\_hallway\_3 & dcab252\_hallway\_3 & 0.91 \\ 
    49 & d31d981\_office\_8 & 54e6de3\_office\_8 & 0.89 \\ 
    50 & f85a909\_office\_3 & c9feabc\_office\_3 & 0.88 \\ 
    51 & f85a909\_office\_3 & 97be01e\_office\_3 & 0.89 \\ 
    52 & c9feabc\_office\_3 & 97be01e\_office\_3 & 0.90 \\ 
    54 & 8fd8146\_office\_10 & ab03f88\_office\_10 & 0.88 \\ 
    55 & 7c870c2\_hallway\_8 & 4de69cf\_hallway\_8 & 0.87 \\ 
    56 & 33e598f\_office\_15 & 8910cb1\_office\_15 & 0.88 \\ 
    58 & 46b4538\_office\_1 & db2e53f\_office\_1 & 0.92 \\ 
    \hline
    \end{tabular}
    \caption{\small Easy split of Stanford2D3DS keypoints dataset image pairs.}
    \label{tab:keypointdataseteasy}
\end{table*}
    
\vspace{1cm}

\begin{table*}[t]
    \small
    \centering
    \begin{tabular}{|c|l|l|c|}
    \hline
    \multicolumn{4}{|c|}{\textbf{Hard}} \\
    \hline
    \textit{Pair ID} & \textit{Left Image} & \textit{Right Image} & \textit{FOV Overlap} \\
    \hline
    0 & c14611b\_hallway\_7 & 37a4f42\_hallway\_7 & 0.83 \\ 
    6 & 1b253d2\_hallway\_7 & 6e945c8\_hallway\_7 & 0.84 \\ 
    7 & 5d3a59a\_hallway\_7 & ec0b9b3\_hallway\_7 & 0.81 \\ 
    8 & ac01e35\_hallway\_7 & 649838b\_hallway\_7 & 0.85 \\ 
    9 & f6c6ce3\_hallway\_7 & 5221e31\_hallway\_7 & 0.85 \\ 
    11 & 438c5fb\_hallway\_7 & ec0b9b3\_hallway\_7 & 0.82 \\ 
    12 & ec0b9b3\_hallway\_7 & 531efee\_hallway\_7 & 0.85 \\ 
    13 & 724bbea\_hallway\_7 & c8c806b\_hallway\_7 & 0.85 \\ 
    14 & 724bbea\_hallway\_7 & 55db392\_hallway\_7 & 0.82 \\ 
    15 & 32d9e73\_hallway\_7 & 55db392\_hallway\_7 & 0.85 \\ 
    16 & fcd2380\_office\_22 & 2d842ce\_office\_22 & 0.85 \\ 
    17 & 2d842ce\_office\_22 & ffd2cca\_office\_22 & 0.86 \\ 
    18 & 89d9828\_hallway\_6 & 87e6e35\_hallway\_6 & 0.81 \\ 
    19 & 89d9828\_hallway\_6 & 7d58331\_hallway\_6 & 0.84 \\ 
    21 & 75acaa8\_hallway\_6 & 87e6e35\_hallway\_6 & 0.84 \\ 
    22 & 92b146f\_hallway\_6 & 8c78856\_hallway\_6 & 0.86 \\ 
    26 & b640b47\_hallway\_6 & 87e6e35\_hallway\_6 & 0.80 \\ 
    27 & bed890d\_hallway\_6 & 97ab30c\_hallway\_6 & 0.85 \\ 
    32 & af50002\_WC\_1 & 36dd48f\_WC\_1 & 0.84 \\ 
    33 & 1edba7e\_WC\_1 & e0c041d\_WC\_1 & 0.84 \\ 
    36 & c40ca55\_office\_31 & a77fba5\_office\_31 & 0.85 \\ 
    38 & 4a41b27\_office\_31 & da4629d\_office\_31 & 0.82 \\ 
    40 & da4629d\_office\_31 & 9084f21\_office\_31 & 0.84 \\ 
    41 & 75361af\_office\_31 & ecf7fb4\_office\_31 & 0.82 \\ 
    42 & 2100dd9\_office\_4 & 26c24c7\_office\_4 & 0.83 \\ 
    46 & 84cdc9a\_conferenceRoom\_1 & 0d600f9\_conferenceRoom\_1 & 0.83 \\ 
    48 & dcab252\_hallway\_3 & a9cda4d\_hallway\_3 & 0.82 \\ 
    53 & 6549526\_office\_21 & 08aa476\_office\_21 & 0.83 \\ 
    57 & dbcdb33\_office\_20 & f02c98c\_office\_20 & 0.83 \\ 
    59 & 24f42d6\_hallway\_5 & 684b940\_hallway\_5 & 0.84 \\
    \hline
    \end{tabular}
    \caption{\small Hard split of Stanford2D3DS keypoints dataset image pairs.}
    \label{tab:keypointdatasethard}
\end{table*}

%% file: supplementary/tables/keypoint-indiv-results.tex
\begin{table*}[ht]
    \small
    \centering
    \begin{tabular}{|c|||c|c|c|||c|c|c||c|c|c||c|c|c|}
    \hline
    \multicolumn{13}{|c|}{\textbf{Easy}} \\
    \hline
    \textbf{Pair ID} & \multicolumn{3}{|c|||}{\textbf{Equirect.}} & \multicolumn{3}{|c||}{\textbf{L0}} & \multicolumn{3}{|c||}{\textbf{L1}} & \multicolumn{3}{|c|}{\textbf{L2}} \\
    \hline
    & \textit{PMR} & \textit{MS} & \textit{P} & \textit{PMR} & \textit{MS} & \textit{P} & \textit{PMR} & \textit{MS} & \textit{P} & \textit{PMR} & \textit{MS} & \textit{P} \\
    \hline
    1 & 0.27 & 0.07 & 0.27 & 0.35 & 0.09 & 0.26 & \textbf{0.40} & \textbf{0.13} & \textbf{0.33} & 0.34 & 0.10 & 0.28 \\ 
2 & 0.41 & 0.17 & 0.42 & 0.45 & 0.27 & \textbf{0.60} & \textbf{0.50} & \textbf{0.28} & 0.57 & 0.48 & \textbf{0.28} & 0.58 \\ 
3 & 0.34 & 0.19 & 0.55 & 0.40 & 0.23 & 0.56 & \textbf{0.46} & \textbf{0.27} & \textbf{0.60} & 0.44 & 0.25 & 0.58 \\ 
4 & 0.21 & 0.08 & 0.36 & 0.27 & 0.11 & 0.41 & \textbf{0.30} & \textbf{0.14} & \textbf{0.48} & 0.25 & 0.12 & \textbf{0.48} \\ 
5 & 0.59 & \textbf{0.48} & \textbf{0.80} & 0.67 & 0.47 & 0.70 & \textbf{0.70} & 0.45 & 0.64 & 0.66 & 0.35 & 0.54 \\ 
10 & 0.29 & 0.10 & 0.35 & 0.35 & 0.12 & 0.35 & \textbf{0.36} & \textbf{0.14} & \textbf{0.38} & 0.34 & 0.11 & 0.33 \\ 
20 & 0.30 & 0.09 & 0.30 & 0.46 & 0.14 & 0.32 & \textbf{0.50} & \textbf{0.19} & \textbf{0.38} & 0.43 & 0.14 & 0.33 \\ 
23 & 0.23 & 0.07 & 0.32 & 0.33 & 0.13 & 0.39 & \textbf{0.34} & \textbf{0.14} & \textbf{0.42} & \textbf{0.34} & 0.13 & 0.38 \\ 
24 & 0.16 & 0.07 & 0.43 & 0.22 & 0.09 & 0.42 & \textbf{0.24} & \textbf{0.10} & 0.43 & 0.23 & \textbf{0.10} & \textbf{0.44} \\ 
25 & 0.83 & 0.48 & 0.58 & 0.99 & \textbf{0.71} & \textbf{0.72} & \textbf{1.01} & 0.58 & 0.58 & 0.95 & 0.58 & 0.61 \\ 
28 & 0.18 & 0.06 & 0.34 & \textbf{0.25} & 0.08 & 0.33 & 0.24 & \textbf{0.10} & \textbf{0.40} & 0.24 & \textbf{0.10} & \textbf{0.40} \\ 
29 & 0.29 & 0.13 & 0.46 & 0.35 & 0.16 & 0.46 & \textbf{0.38} & \textbf{0.21} & \textbf{0.55} & 0.34 & 0.17 & 0.51 \\ 
30 & 0.17 & 0.07 & \textbf{0.42} & \textbf{0.25} & \textbf{0.08} & 0.33 & 0.24 & \textbf{0.08} & 0.32 & 0.21 & \textbf{0.08} & 0.37 \\ 
31 & 0.14 & 0.06 & 0.47 & 0.15 & 0.07 & 0.48 & \textbf{0.16} & \textbf{0.08} & \textbf{0.51} & 0.14 & 0.07 & \textbf{0.51} \\ 
34 & 0.46 & 0.39 & \textbf{0.86} & 0.48 & 0.38 & 0.78 & 0.51 & 0.42 & 0.82 & \textbf{0.52} & \textbf{0.43} & 0.84 \\ 
35 & 0.42 & 0.34 & \textbf{0.82} & 0.43 & 0.33 & 0.77 & 0.46 & \textbf{0.38} & \textbf{0.82} & \textbf{0.47} & \textbf{0.38} & \textbf{0.82} \\ 
37 & 0.23 & 0.09 & 0.40 & 0.24 & 0.10 & 0.42 & \textbf{0.25} & \textbf{0.12} & \textbf{0.47} & \textbf{0.25} & 0.11 & 0.43 \\ 
39 & 0.22 & 0.08 & \textbf{0.39} & 0.23 & 0.08 & 0.34 & \textbf{0.24} & \textbf{0.09} & 0.36 & 0.23 & 0.07 & 0.32 \\ 
43 & 0.27 & 0.19 & \textbf{0.70} & 0.38 & 0.24 & 0.63 & \textbf{0.39} & \textbf{0.25} & 0.63 & 0.37 & 0.24 & 0.65 \\ 
44 & 0.13 & 0.04 & 0.30 & 0.18 & 0.07 & \textbf{0.39} & \textbf{0.23} & \textbf{0.08} & 0.36 & 0.16 & 0.06 & 0.36 \\ 
45 & 0.14 & 0.05 & 0.36 & 0.17 & 0.07 & 0.40 & \textbf{0.21} & \textbf{0.09} & \textbf{0.44} & 0.17 & 0.06 & 0.39 \\ 
47 & 0.31 & 0.21 & \textbf{0.67} & 0.40 & \textbf{0.25} & 0.64 & \textbf{0.42} & 0.22 & 0.52 & 0.36 & 0.21 & 0.57 \\ 
49 & 0.10 & 0.04 & 0.41 & \textbf{0.15} & 0.05 & 0.36 & \textbf{0.15} & \textbf{0.06} & 0.37 & \textbf{0.15} & \textbf{0.06} & \textbf{0.42} \\ 
50 & 0.18 & 0.08 & 0.46 & 0.21 & \textbf{0.11} & 0.50 & \textbf{0.23} & \textbf{0.11} & 0.47 & 0.22 & \textbf{0.11} & \textbf{0.52} \\ 
51 & 0.15 & 0.04 & 0.31 & 0.19 & 0.06 & 0.32 & \textbf{0.22} & \textbf{0.07} & 0.31 & 0.19 & \textbf{0.07} & \textbf{0.38} \\ 
52 & 0.15 & 0.05 & 0.32 & 0.18 & 0.06 & 0.35 & \textbf{0.19} & \textbf{0.08} & \textbf{0.39} & 0.18 & 0.06 & 0.34 \\ 
54 & 0.17 & 0.05 & 0.31 & 0.24 & \textbf{0.09} & 0.35 & \textbf{0.25} & 0.08 & 0.33 & 0.22 & 0.08 & \textbf{0.38} \\ 
55 & 0.18 & 0.10 & \textbf{0.53} & 0.24 & 0.11 & 0.45 & \textbf{0.26} & \textbf{0.13} & 0.49 & 0.22 & 0.07 & 0.32 \\ 
56 & 0.22 & 0.11 & \textbf{0.50} & 0.32 & \textbf{0.16} & \textbf{0.50} & \textbf{0.33} & \textbf{0.16} & 0.48 & 0.29 & 0.14 & 0.49 \\ 
58 & 0.16 & 0.06 & 0.37 & 0.19 & 0.07 & 0.37 & \textbf{0.22} & \textbf{0.09} & \textbf{0.40} & 0.20 & 0.08 & 0.38 \\ 
    \hline
    \end{tabular}
    \caption{\small Keypoint matching results on individual image pairs in the easy split.}
    \label{tab:keypointindivresultseasy}
\end{table*}

\begin{table*}[ht]
    \small
    \centering
    
    \begin{tabular}{|c|||c|c|c|||c|c|c||c|c|c||c|c|c|}
    \hline
    \multicolumn{13}{|c|}{\textbf{Hard}} \\
    \hline
    \textbf{Pair ID} & \multicolumn{3}{|c|||}{\textbf{Equirect.}} & \multicolumn{3}{|c||}{\textbf{L0}} & \multicolumn{3}{|c||}{\textbf{L1}} & \multicolumn{3}{|c|}{\textbf{L2}} \\
    \hline
    & \textit{PMR} & \textit{MS} & \textit{P} & \textit{PMR} & \textit{MS} & \textit{P} & \textit{PMR} & \textit{MS} & \textit{P} & \textit{PMR} & \textit{MS} & \textit{P} \\
    \hline
    0 & 0.25 & 0.12 & \textbf{0.51} & 0.28 & 0.14 & 0.50 & \textbf{0.30} & \textbf{0.15} & 0.49 & \textbf{0.30} & \textbf{0.15} & 0.49 \\ 
6 & 0.26 & 0.09 & 0.33 & 0.35 & 0.15 & \textbf{0.43} & \textbf{0.38} & \textbf{0.16} & \textbf{0.43} & 0.34 & 0.14 & 0.40 \\ 
7 & 0.22 & 0.07 & 0.32 & 0.28 & 0.09 & \textbf{0.34} & \textbf{0.29} & 0.09 & 0.32 & \textbf{0.29} & \textbf{0.10} & \textbf{0.34} \\ 
8 & 0.23 & 0.08 & 0.38 & \textbf{0.30} & \textbf{0.14} & \textbf{0.46} & \textbf{0.30} & 0.11 & 0.37 & 0.29 & 0.11 & 0.39 \\ 
9 & 0.31 & 0.08 & 0.26 & \textbf{0.42} & 0.12 & 0.28 & \textbf{0.42} & \textbf{0.13} & \textbf{0.32} & 0.39 & 0.11 & 0.27 \\ 
11 & 0.23 & 0.07 & 0.32 & 0.28 & \textbf{0.11} & \textbf{0.41} & \textbf{0.32} & 0.08 & 0.25 & 0.27 & 0.08 & 0.29 \\ 
12 & 0.20 & 0.05 & 0.24 & \textbf{0.34} & \textbf{0.09} & 0.26 & 0.33 & 0.07 & 0.21 & 0.29 & \textbf{0.09} & \textbf{0.32} \\ 
13 & 0.24 & 0.07 & 0.30 & 0.35 & 0.10 & 0.29 & \textbf{0.37} & \textbf{0.11} & 0.30 & 0.31 & 0.10 & \textbf{0.31} \\ 
14 & 0.26 & 0.08 & 0.32 & 0.43 & 0.10 & 0.24 & \textbf{0.50} & \textbf{0.17} & \textbf{0.33} & 0.40 & 0.12 & 0.30 \\ 
15 & 0.30 & 0.12 & 0.40 & 0.40 & 0.19 & 0.47 & \textbf{0.42} & \textbf{0.21} & \textbf{0.51} & 0.36 & 0.17 & 0.49 \\ 
16 & 0.16 & 0.05 & 0.34 & \textbf{0.19} & 0.06 & 0.35 & \textbf{0.19} & \textbf{0.07} & \textbf{0.37} & 0.17 & 0.06 & 0.36 \\ 
17 & 0.19 & 0.09 & 0.47 & 0.21 & 0.11 & \textbf{0.51} & \textbf{0.24} & \textbf{0.12} & 0.49 & 0.21 & 0.11 & \textbf{0.51} \\ 
18 & 0.24 & 0.06 & 0.26 & \textbf{0.36} & \textbf{0.12} & 0.33 & \textbf{0.36} & 0.10 & 0.28 & 0.31 & \textbf{0.12} & \textbf{0.38} \\ 
19 & 0.20 & 0.06 & 0.28 & 0.31 & 0.09 & 0.29 & \textbf{0.34} & \textbf{0.12} & 0.34 & 0.28 & \textbf{0.12} & \textbf{0.42} \\ 
21 & 0.22 & 0.08 & 0.35 & 0.30 & 0.11 & 0.37 & \textbf{0.31} & \textbf{0.12} & \textbf{0.38} & 0.29 & 0.10 & 0.34 \\ 
22 & 0.25 & 0.07 & 0.29 & 0.35 & 0.12 & 0.35 & \textbf{0.36} & \textbf{0.13} & 0.37 & 0.33 & \textbf{0.13} & \textbf{0.41} \\ 
26 & 0.21 & 0.06 & \textbf{0.31} & 0.31 & \textbf{0.10} & \textbf{0.31} & \textbf{0.33} & \textbf{0.10} & 0.30 & 0.29 & 0.08 & 0.29 \\ 
27 & 0.16 & 0.06 & 0.37 & 0.24 & 0.11 & 0.46 & \textbf{0.25} & \textbf{0.12} & \textbf{0.48} & 0.22 & 0.10 & 0.46 \\ 
32 & 0.25 & 0.09 & 0.37 & 0.30 & 0.12 & 0.39 & \textbf{0.34} & \textbf{0.15} & \textbf{0.43} & 0.30 & 0.12 & 0.39 \\ 
33 & 0.19 & 0.09 & 0.49 & 0.24 & 0.12 & 0.50 & 0.25 & 0.13 & 0.51 & \textbf{0.26} & \textbf{0.14} & \textbf{0.53} \\ 
36 & 0.23 & 0.10 & 0.42 & 0.25 & \textbf{0.11} & \textbf{0.44} & \textbf{0.26} & \textbf{0.11} & \textbf{0.44} & 0.25 & 0.10 & 0.42 \\ 
38 & 0.22 & \textbf{0.09} & 0.39 & \textbf{0.23} & 0.08 & 0.37 & \textbf{0.23} & \textbf{0.09} & 0.37 & \textbf{0.23} & \textbf{0.09} & \textbf{0.41} \\ 
40 & 0.20 & 0.10 & 0.50 & 0.22 & \textbf{0.11} & 0.51 & \textbf{0.23} & \textbf{0.11} & 0.50 & 0.22 & \textbf{0.11} & \textbf{0.52} \\ 
41 & 0.23 & 0.12 & 0.52 & 0.25 & 0.14 & 0.54 & 0.26 & 0.14 & 0.54 & \textbf{0.27} & \textbf{0.15} & \textbf{0.56} \\ 
42 & 0.17 & 0.05 & 0.30 & \textbf{0.21} & 0.08 & 0.37 & \textbf{0.21} & \textbf{0.09} & 0.41 & 0.20 & 0.08 & \textbf{0.42} \\ 
46 & 0.21 & \textbf{0.10} & \textbf{0.50} & \textbf{0.25} & \textbf{0.10} & 0.40 & \textbf{0.25} & \textbf{0.10} & 0.39 & 0.24 & 0.08 & 0.35 \\ 
48 & 0.27 & 0.08 & 0.29 & \textbf{0.32} & 0.12 & 0.38 & \textbf{0.32} & \textbf{0.13} & 0.41 & 0.29 & \textbf{0.13} & \textbf{0.44} \\ 
53 & 0.15 & 0.05 & \textbf{0.33} & 0.15 & 0.05 & 0.32 & \textbf{0.17} & \textbf{0.06} & 0.32 & 0.15 & 0.04 & 0.26 \\ 
57 & 0.17 & 0.07 & 0.43 & \textbf{0.20} & \textbf{0.10} & 0.47 & \textbf{0.20} & \textbf{0.10} & \textbf{0.51} & \textbf{0.20} & \textbf{0.10} & 0.49 \\ 
59 & 0.26 & 0.13 & 0.50 & 0.28 & 0.15 & 0.53 & 0.27 & 0.15 & 0.53 & \textbf{0.29} & \textbf{0.16} & \textbf{0.54} \\ 
    \hline
    \end{tabular}
    \caption{\small Keypoint matching results on individual image pairs in the hard split.}
    \label{tab:keypointindivresultshard}
\end{table*}

%% file: supplementary/figures/sift-detections.tex
\begin{figure*}[ht]
    \centering
    \begin{subfigure}[b]{0.45\textwidth}
        \begin{subfigure}[b]{\textwidth}
            \centering
            \includegraphics[width=1.0\linewidth, trim={0mm 0mm 0mm 0mm}, clip]{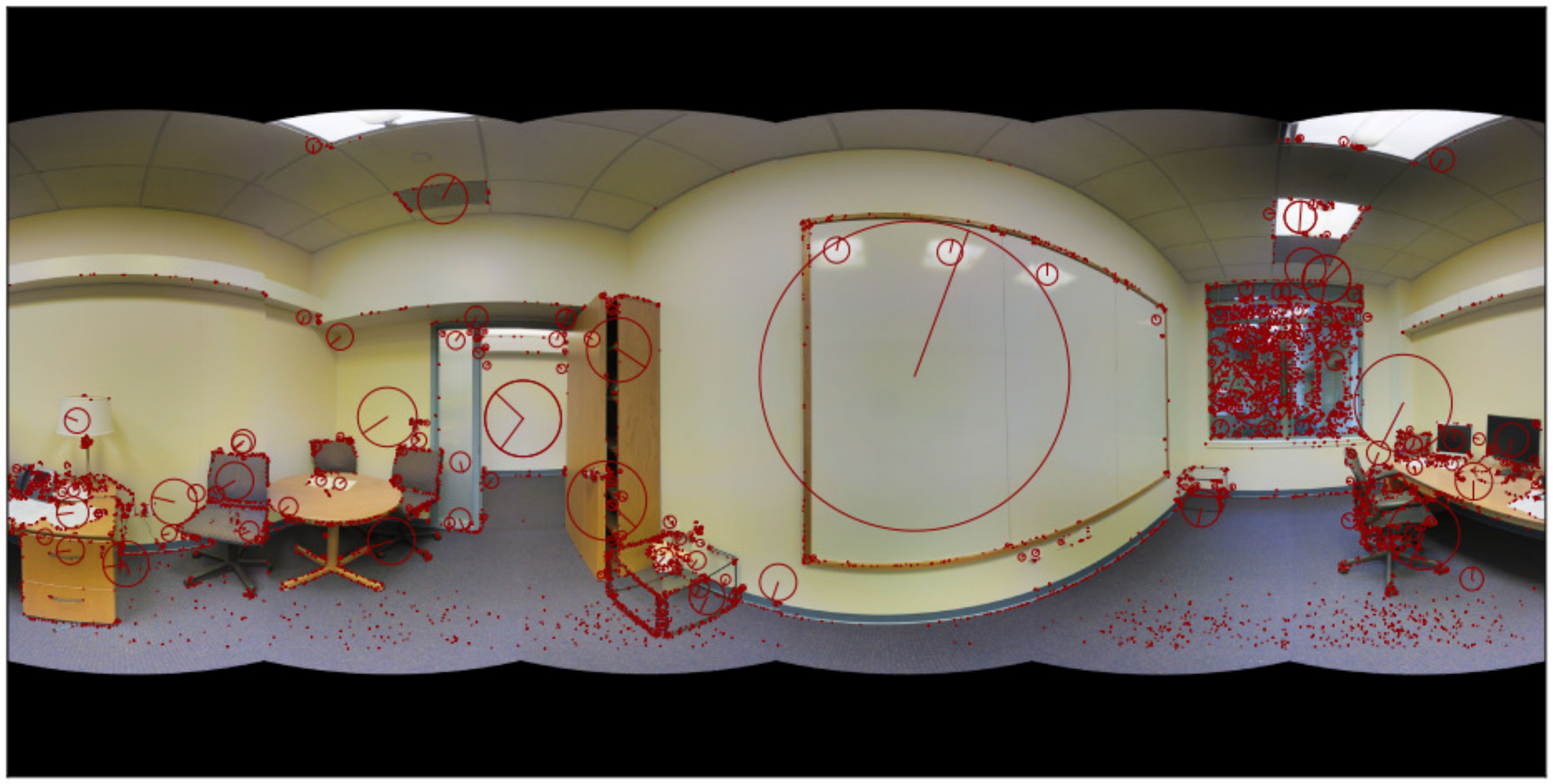}
        \end{subfigure}\\
        \begin{subfigure}[b]{\textwidth}
            \centering
            \includegraphics[width=1.0\linewidth, trim={0mm 0mm 0mm 0mm}, clip]{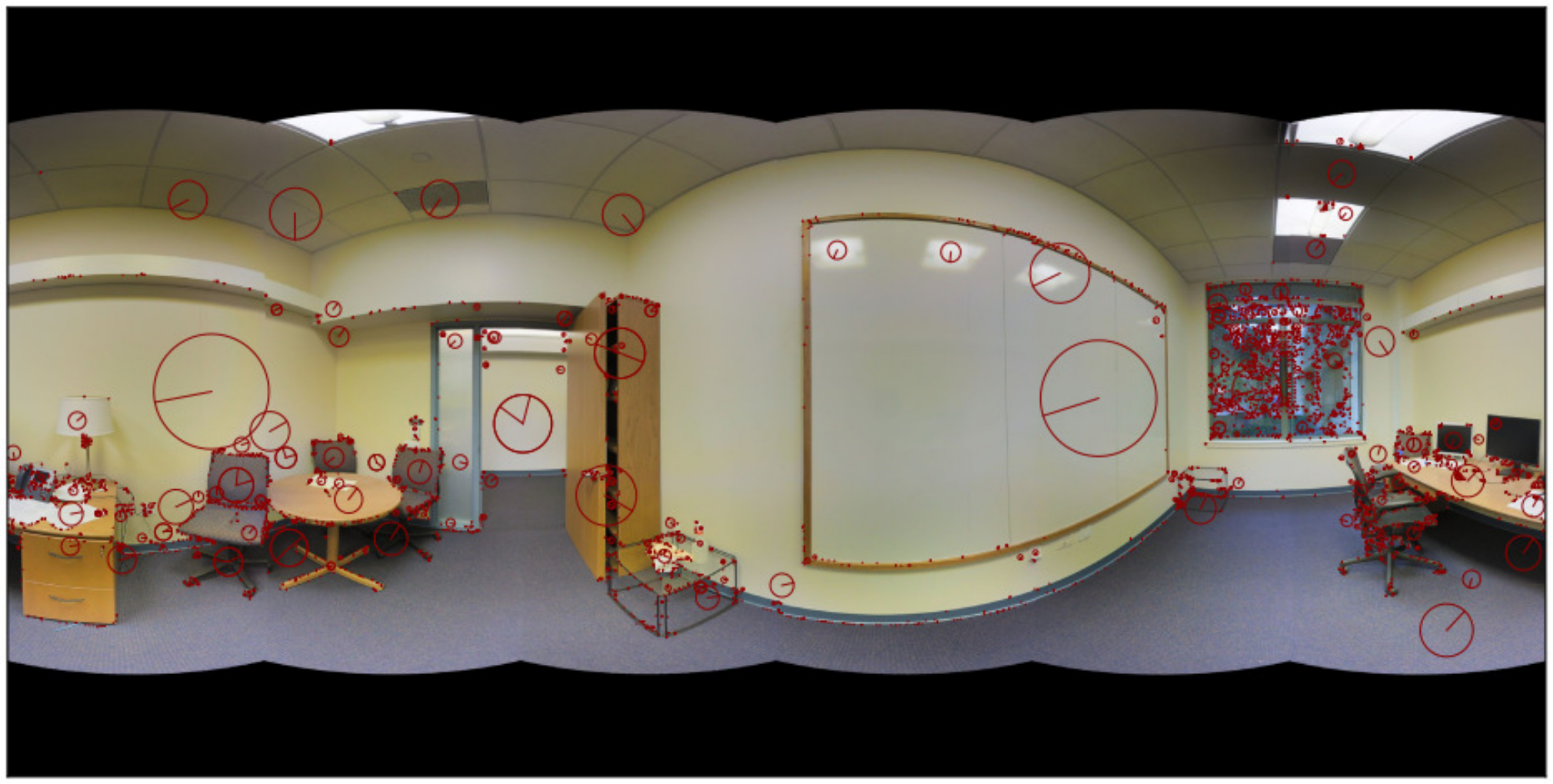}
        \end{subfigure}\\
        \begin{subfigure}[b]{\textwidth}
            \centering
            \includegraphics[width=1.0\linewidth, trim={0mm 0mm 0mm 0mm}, clip]{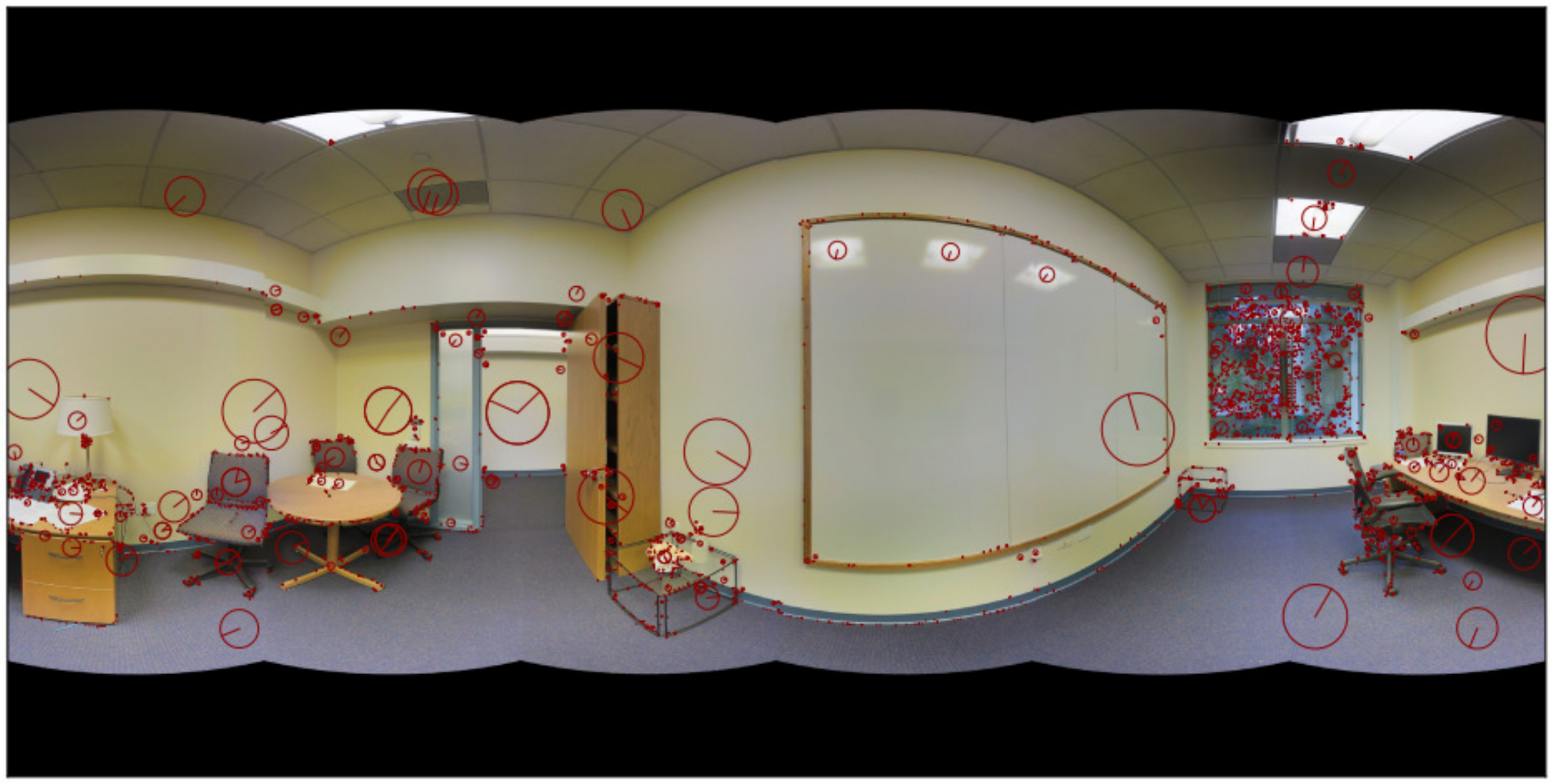}
        \end{subfigure}\\
        \begin{subfigure}[b]{\textwidth}
            \centering
            \includegraphics[width=1.0\linewidth, trim={0mm 0mm 0mm 0mm}, clip]{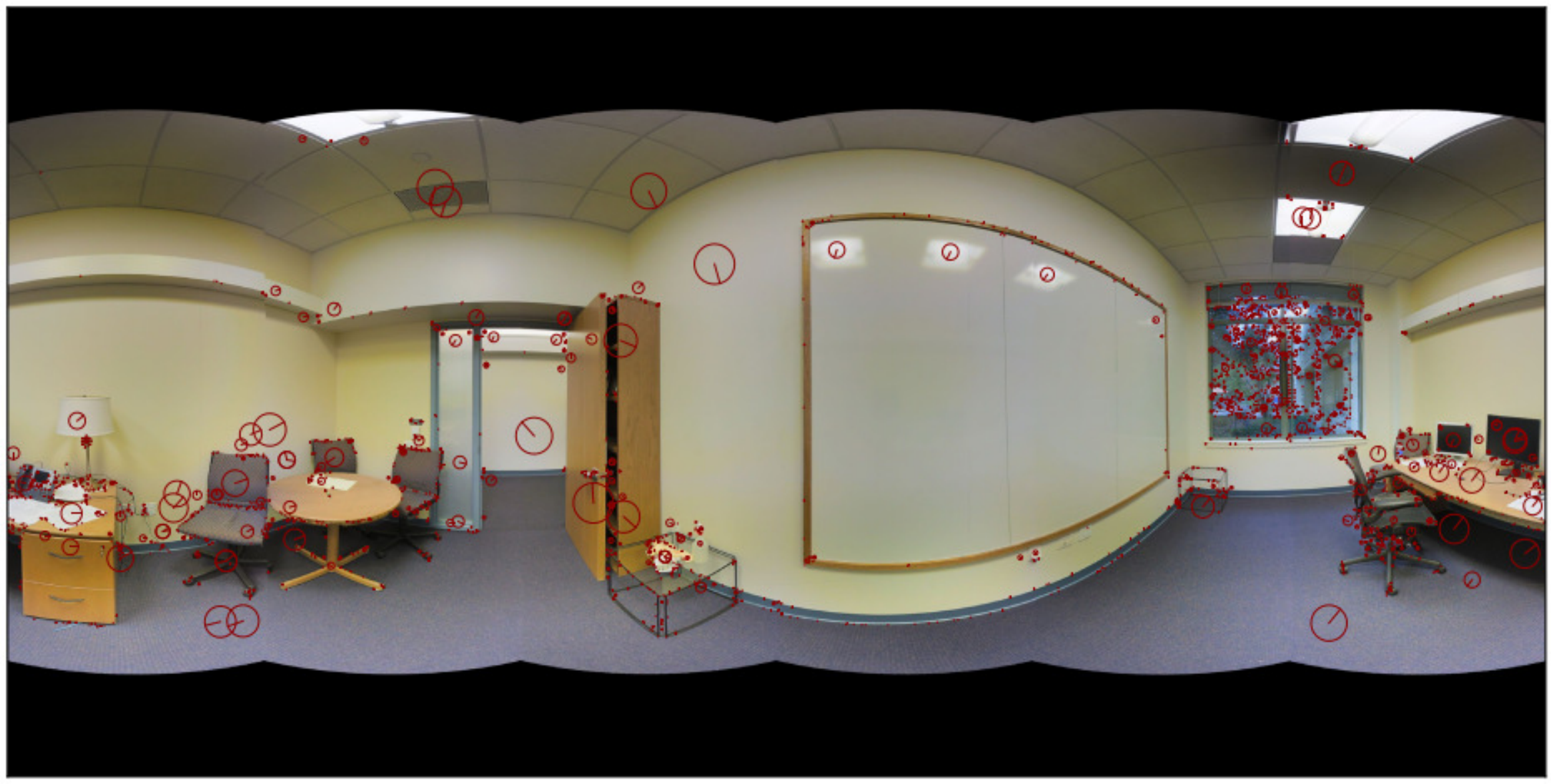}
        \end{subfigure}\\
        \caption{\small From image pair 58}
    \end{subfigure}
    ~
    \hspace{5mm}
    \begin{subfigure}[b]{0.45\textwidth}
        \begin{subfigure}[b]{\textwidth}
            \centering
            \includegraphics[width=1.0\linewidth, trim={0mm 0mm 0mm 0mm}, clip]{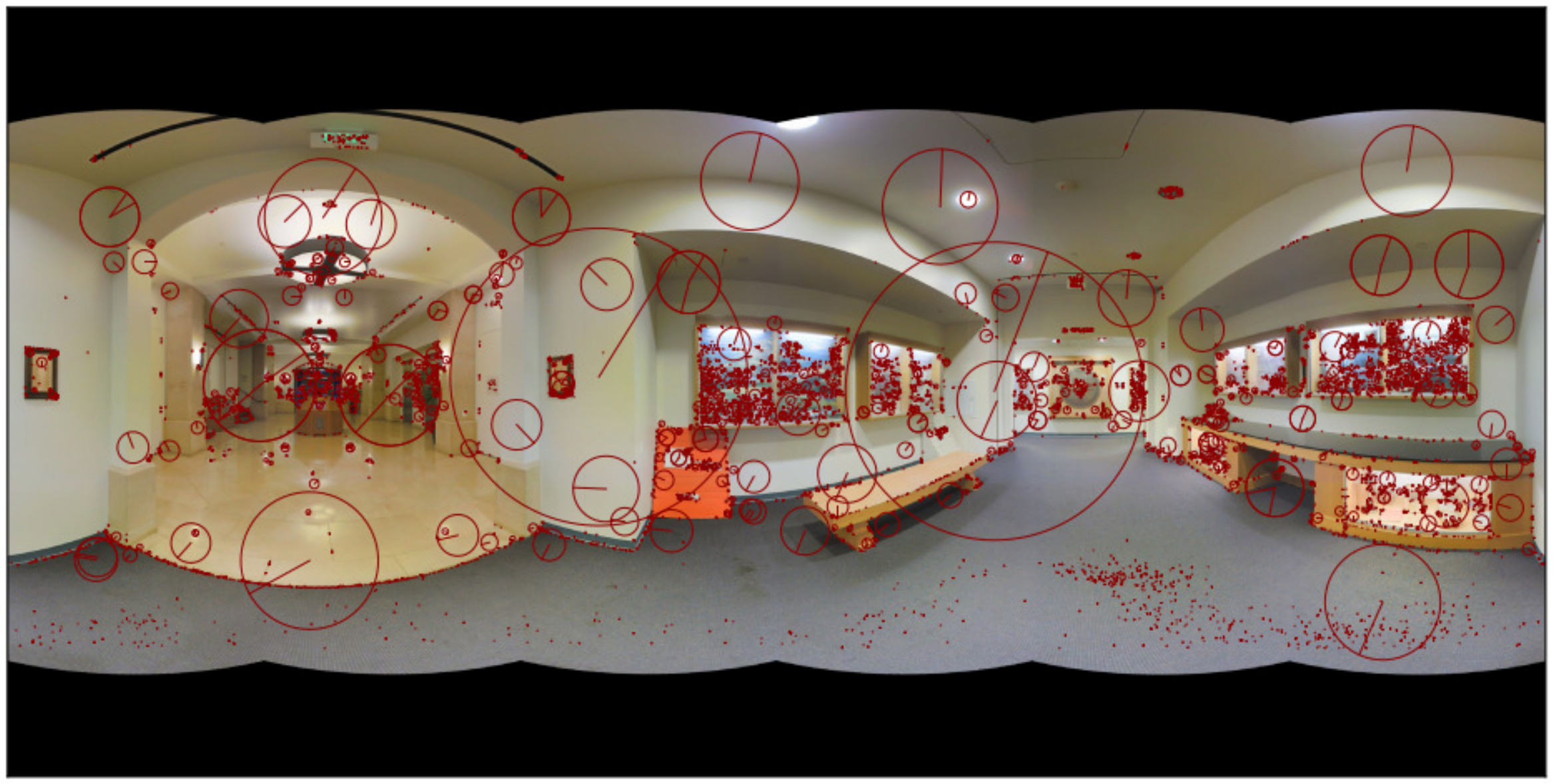}
        \end{subfigure}\\
        \begin{subfigure}[b]{\textwidth}
            \centering
            \includegraphics[width=1.0\linewidth, trim={0mm 0mm 0mm 0mm}, clip]{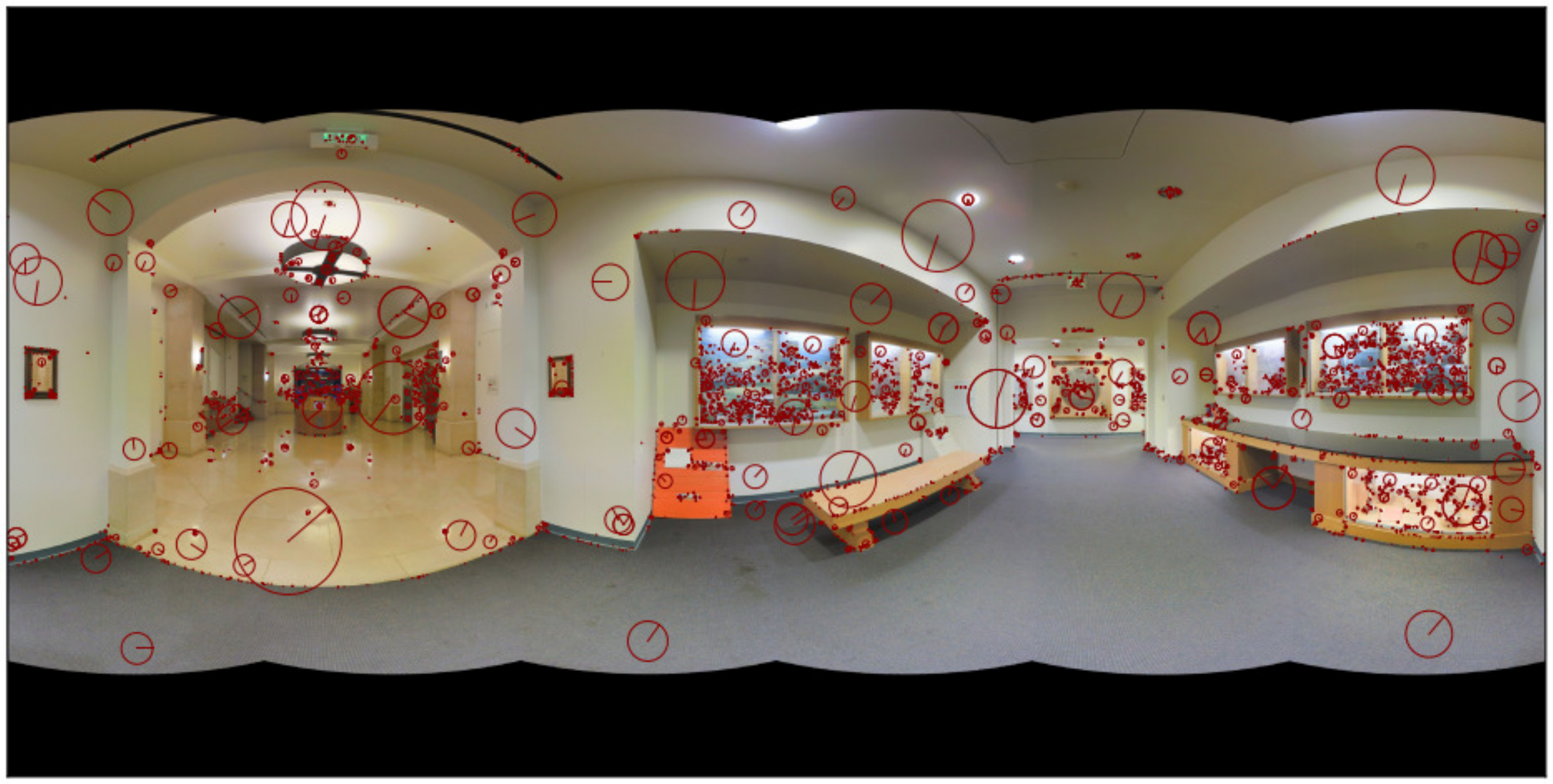}
        \end{subfigure}\\
        \begin{subfigure}[b]{\textwidth}
            \centering
            \includegraphics[width=1.0\linewidth, trim={0mm 0mm 0mm 0mm}, clip]{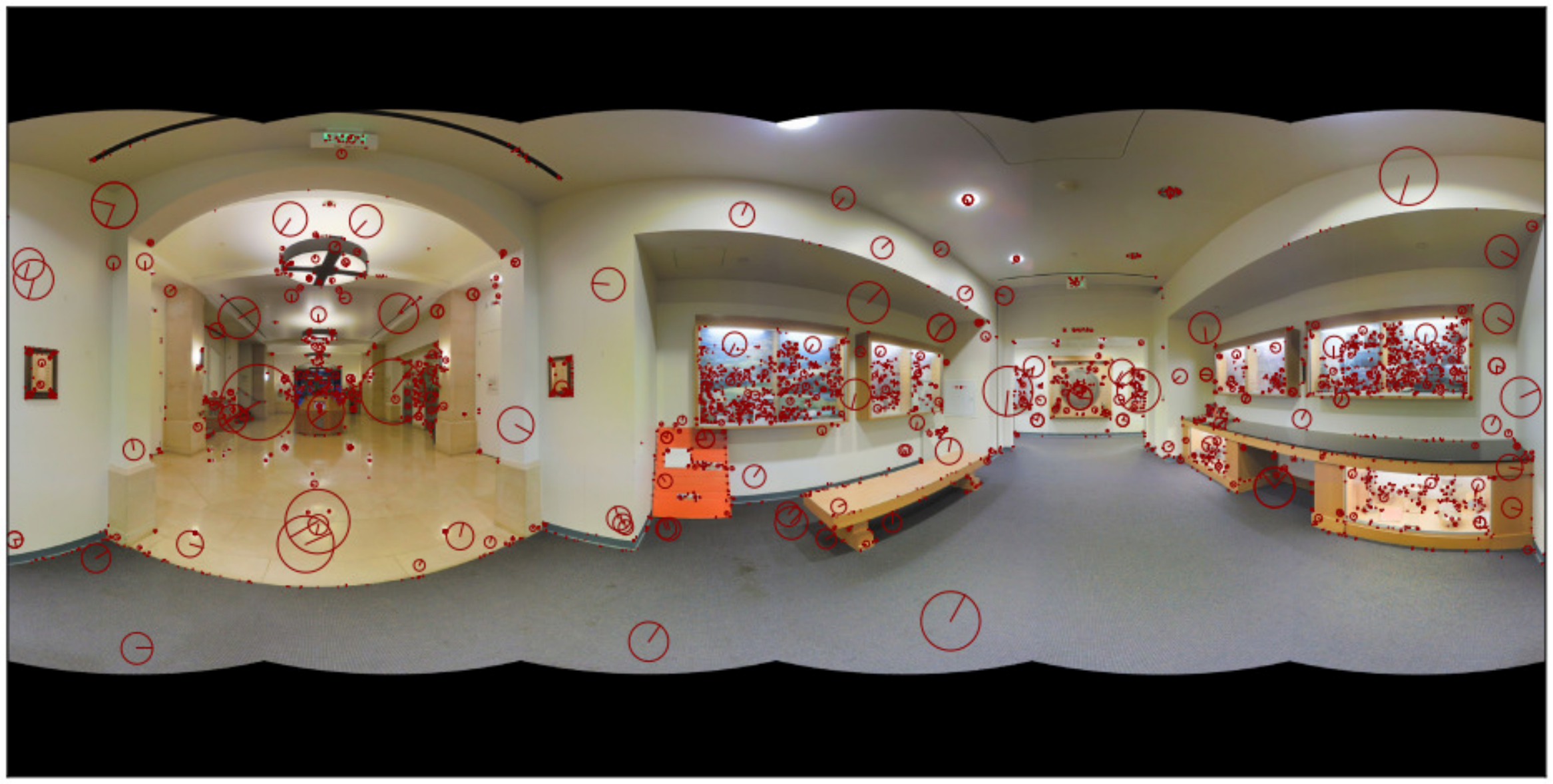}
            
        \end{subfigure}\\
        \begin{subfigure}[b]{\textwidth}
            \centering
            \includegraphics[width=1.0\linewidth, trim={0mm 0mm 0mm 0mm}, clip]{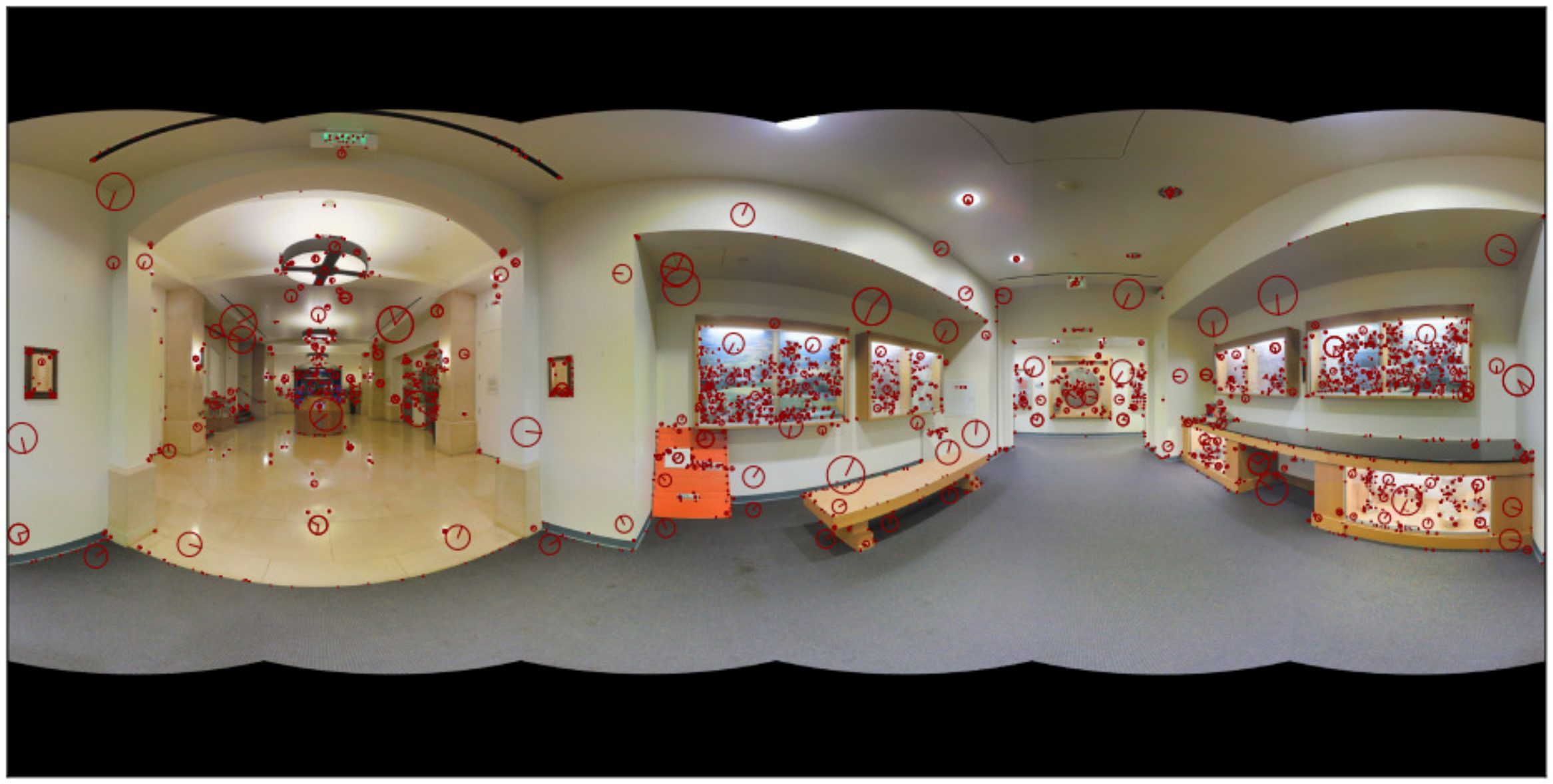}
            
        \end{subfigure}\\
        \caption{\small From image pair 33}
    \end{subfigure}
\caption{\small Comparison of SIFT keypoint detections. Each column, top to bottom: equirectangular, L0, L1, L2}
\label{fig:siftdetections}
\end{figure*}

%% file: supplementary/figures/sift-matches.tex
\begin{figure*}[ht]
    \centering
    \begin{subfigure}[b]{1.0\textwidth}
        \centering
        \includegraphics[width=1.0\linewidth, trim={0mm 0mm 0mm 0mm}, clip]{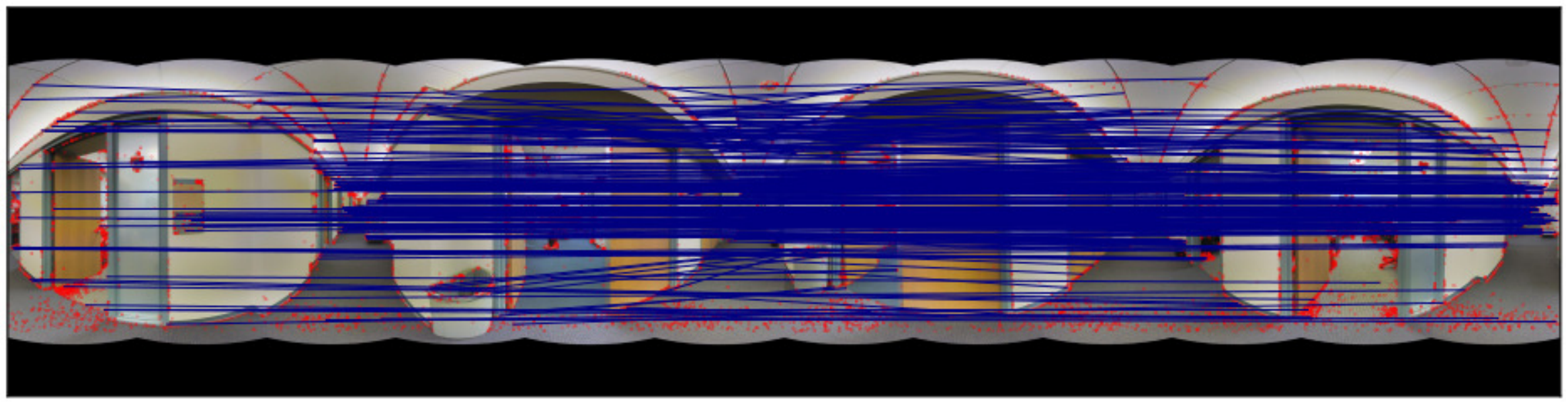}
    \end{subfigure}\\
    \begin{subfigure}[b]{1.0\textwidth}
        \centering
       \includegraphics[width=1.0\linewidth, trim={0mm 0mm 0mm 0mm}, clip]{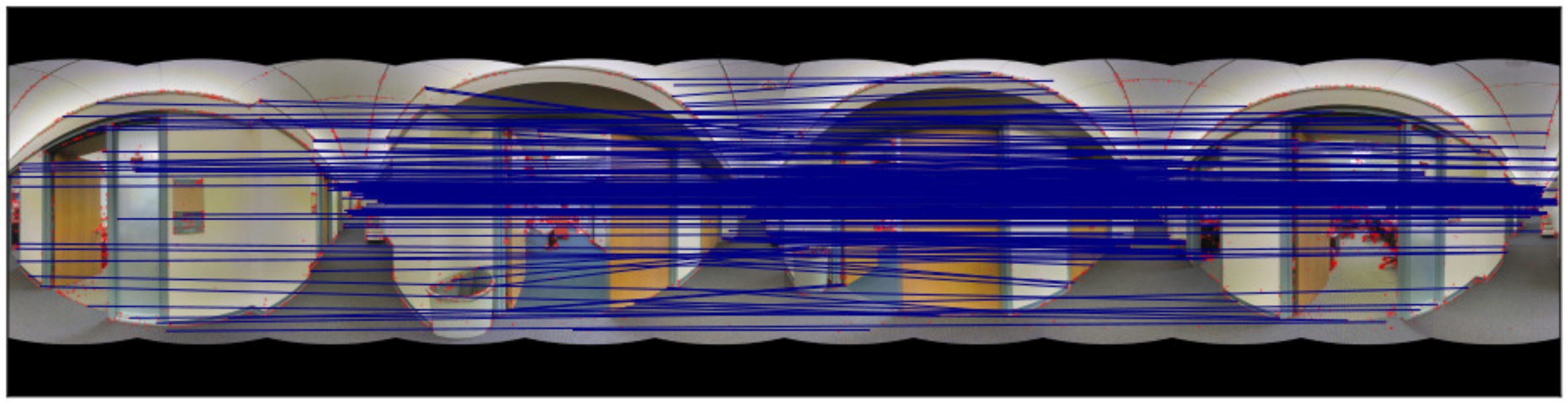}
    \end{subfigure}\\
    \begin{subfigure}[b]{1.0\textwidth}
        \centering
       \includegraphics[width=1.0\linewidth, trim={0mm 0mm 0mm 0mm}, clip]{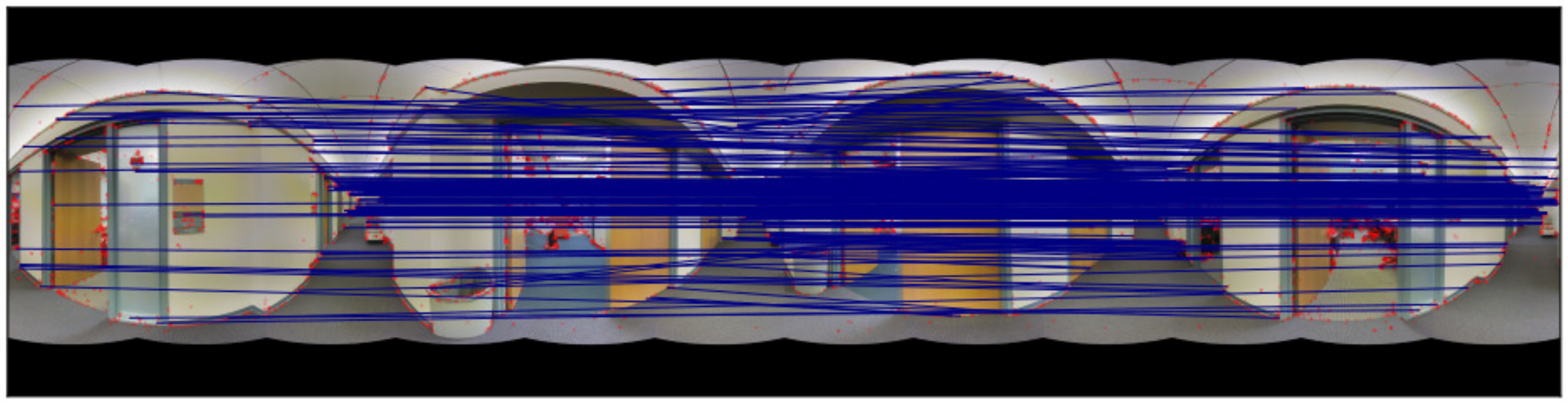}
    \end{subfigure}\\
    \begin{subfigure}[b]{1.0\textwidth}
        \centering
       \includegraphics[width=1.0\linewidth, trim={0mm 0mm 0mm 0mm}, clip]{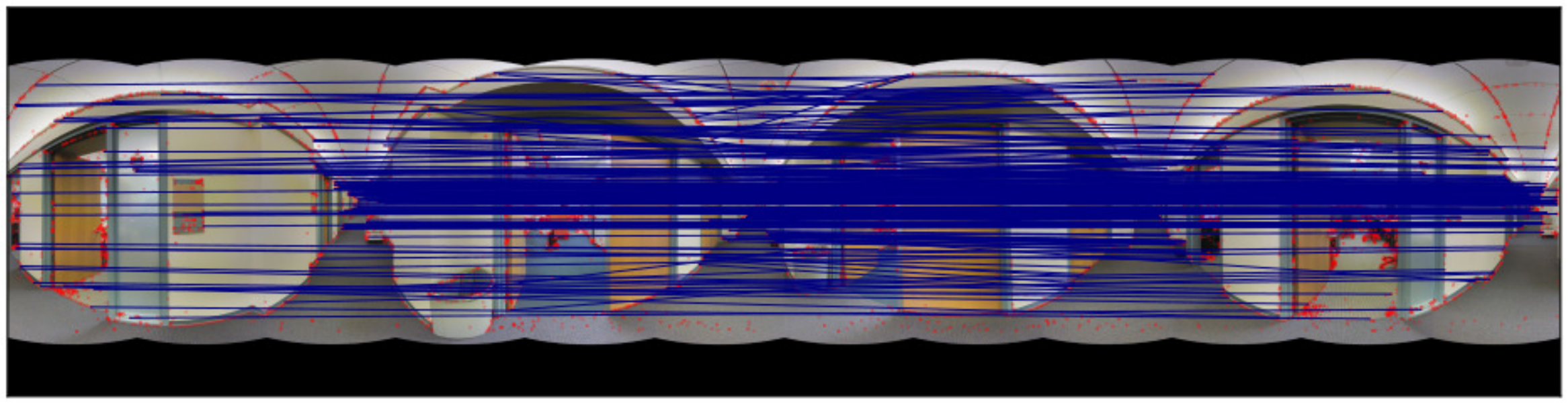}
    \end{subfigure}\\
\caption{\small Comparison of SIFT matches on image pair 15. From top to bottom: equirectangular, L0, L1, L2.}
\label{fig:siftcorr}
\end{figure*}